%% file: main.tex
\documentclass{package/fairmeta}

\title{
\vspace{-1pt}
\revision{%
  {\setstretch{1}\fontsize{13.8}{17.5}\selectfont
  LLM4Robot-Bench: How Far Can GPT-6 Astra Go in Robotic Manipulation?\par}%
  \vspace{4pt}%
  {\setstretch{1}\fontsize{12.5}{15}\selectfont\mdseries
  A Hierarchical, Paired Sim--Real Benchmark\par}%
}
}

\renewcommand\authorformat[2][]{\mbox{\sffamily\bfseries #2$^{#1}$}}
\renewcommand\author[2][]{\addtolist[#1]{#2}{\authorlist}{\authorformat}{\hspace{0.7em}}}
\renewcommand\affiliationformat[2][]{\mbox{\fontsize{8}{10}\selectfont $^{#1}$#2}}
\renewcommand\affiliation[2][]{\addtolist[#1]{#2}{\affiliationlist}{\affiliationformat}{\hspace{0.8em}}}

\author[1,2,*,\ddagger]{Haibao Yu}
\author[1,*]{Jiaqi Meng}
\author[1,3,*]{Zijian Zhang}
\author[1]{Bowen Jing}
\author[4]{Lei Yang}
\author[5]{Yue Hu}
\author[1,6]{Chenchen Ge}
\author[1,6]{Yiming Hou}
\author[1]{Chenbo Xin}
\author[1]{Dawen Ye}
\author[1]{Haifang Qin}
\author[7]{Jin Li}
\author[8]{Weitao Zhou}
\author[9]{Jianing Qiu}
\author[10]{Jiankai Sun}
\author[11]{Dandan Zhang}
\author[12]{Zaiqing Nie}
\author[2]{Ping Luo}
\author[13]{Ding Zhao}

\affiliation[1]{Tuojing Intelligence}
\affiliation[2]{The University of Hong Kong}
\affiliation[3]{Chinese Academy of Sciences}
\affiliation[4]{Nanyang Technological University}
\affiliation[5]{University of Michigan, Ann Arbor}
\affiliation[6]{Southeast University}
\affiliation[7]{Sapient Intelligence}
\affiliation[8]{SimpleAI}
\affiliation[9]{Mohamed bin Zayed University of Artificial Intelligence}
\affiliation[10]{Stanford University}
\affiliation[11]{Imperial College London}
\affiliation[12]{Tsinghua University}
\affiliation[13]{Carnegie Mellon University}

\contribution[*]{Equal contribution}
\contribution[\ddagger]{Corresponding author}

\input{package/preamble}

\newcommand{\institutionLogo}[2][]{%
  \begin{minipage}[c][1.14cm][c]{0.135\linewidth}
    \centering
    \makebox[\linewidth][c]{\includegraphics[keepaspectratio,#1]{#2}}%
  \end{minipage}%
}
\newcommand{\institutionLogos}{%
  \par\vspace{3pt}
  \begingroup
  \setlength{\parskip}{0pt}
  \noindent\institutionLogo[width=2.30cm,height=0.90cm]{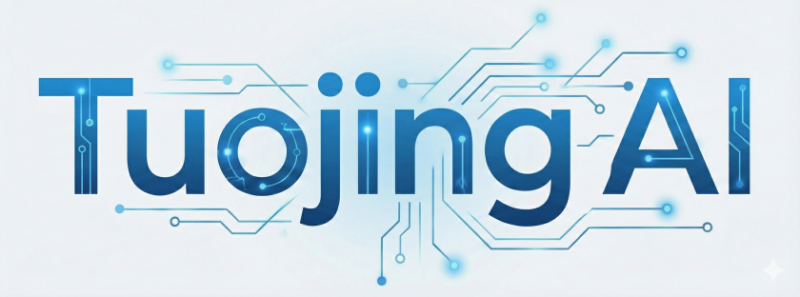}\hfill
  \institutionLogo[height=1.05cm]{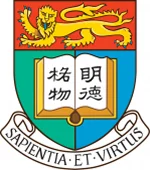}\hfill
  \institutionLogo[height=1.05cm]{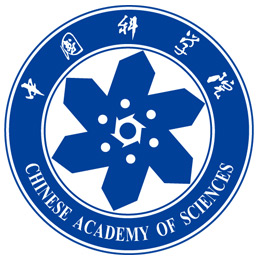}\hfill
  \institutionLogo[trim=0bp 0bp 517.44bp 0bp,clip,height=1.05cm]{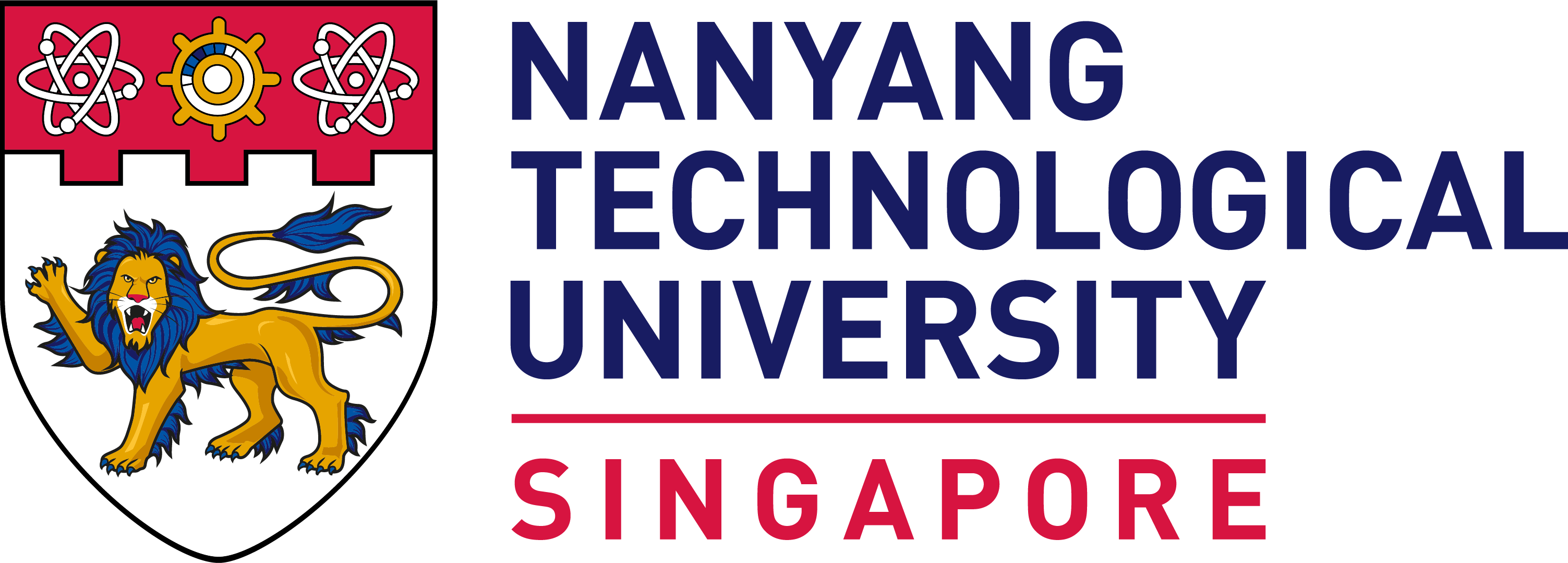}\hfill
  \institutionLogo[height=0.86cm]{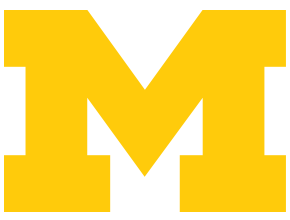}\hfill
  \institutionLogo[height=1.05cm]{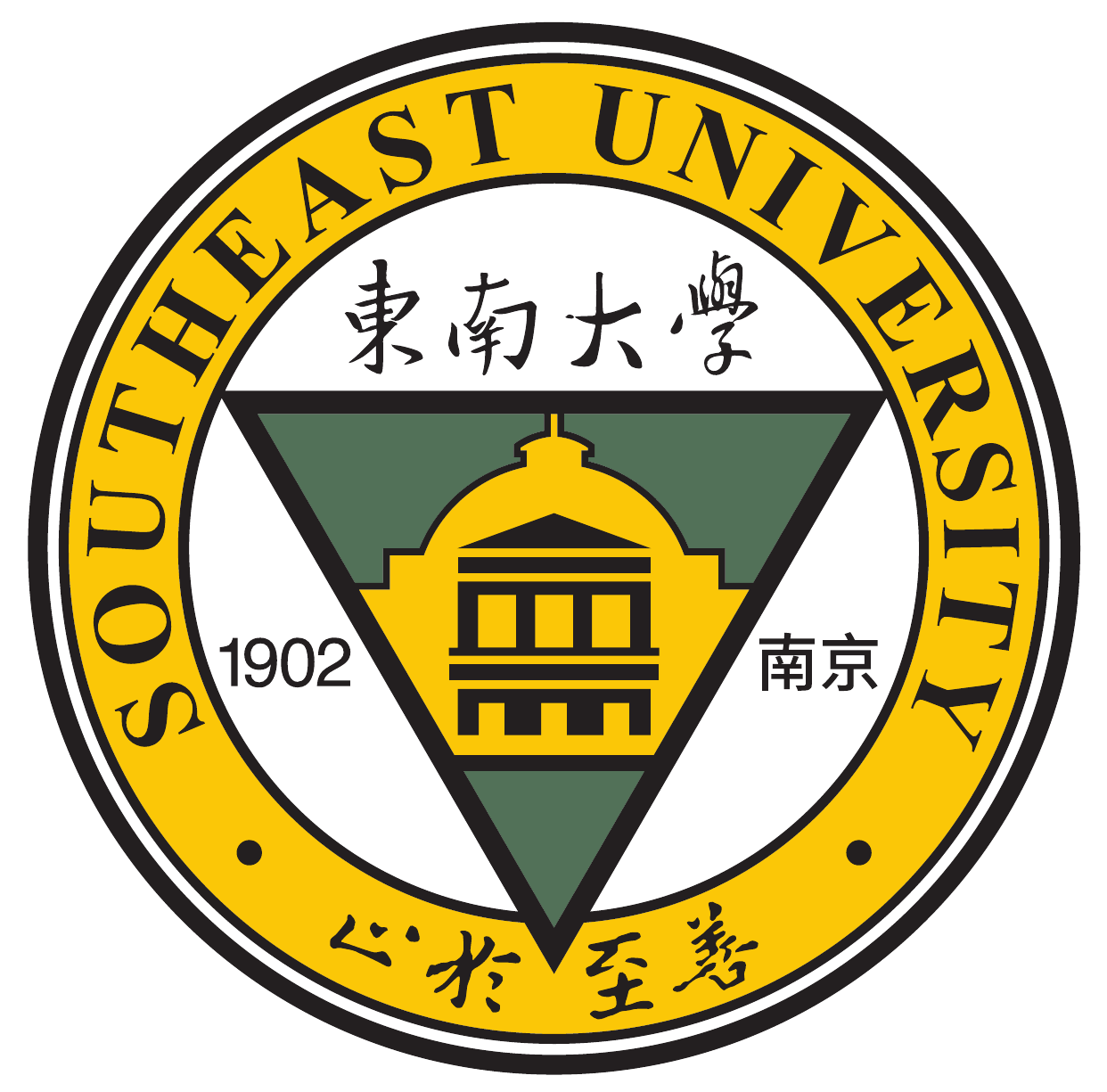}\par
  \vspace{3pt}
  \noindent\institutionLogo[height=1.00cm]{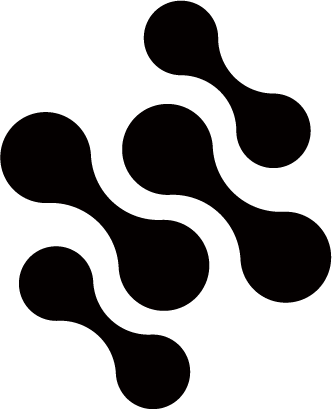}\hfill
  \institutionLogo[trim=88bp 89bp 89bp 88bp,clip,width=2.30cm]{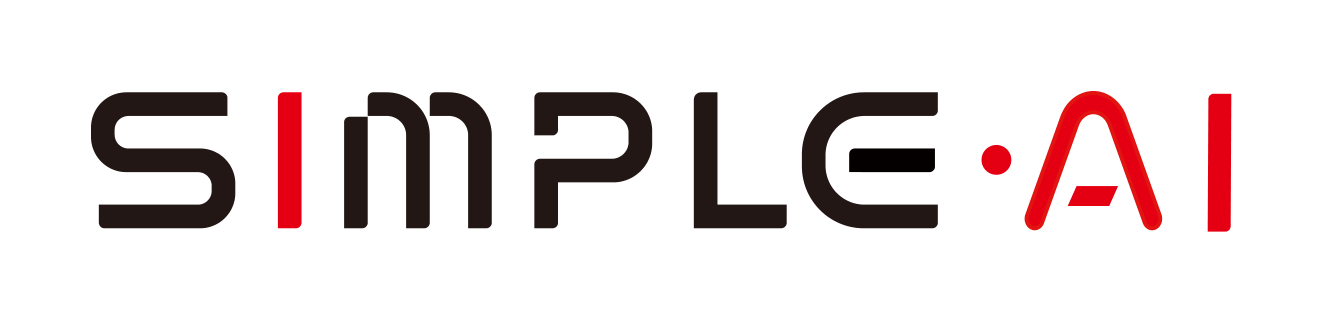}\hfill
  \institutionLogo[trim=45bp 69bp 46bp 70bp,clip,height=1.00cm]{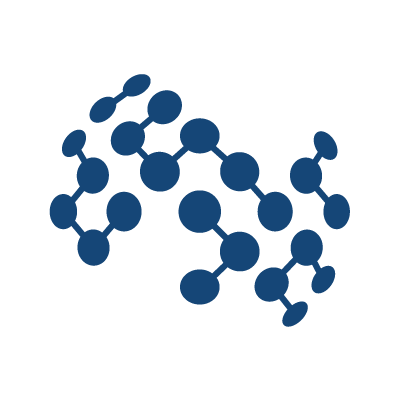}\hfill
  \institutionLogo[height=1.12cm]{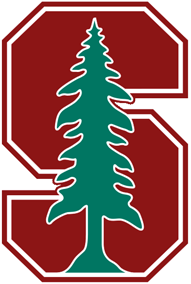}\hfill
  \institutionLogo[width=2.55cm]{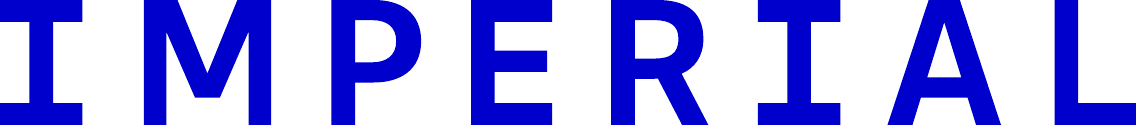}\hfill
  \institutionLogo[trim=450bp 65bp 478bp 65bp,clip,height=1.05cm]{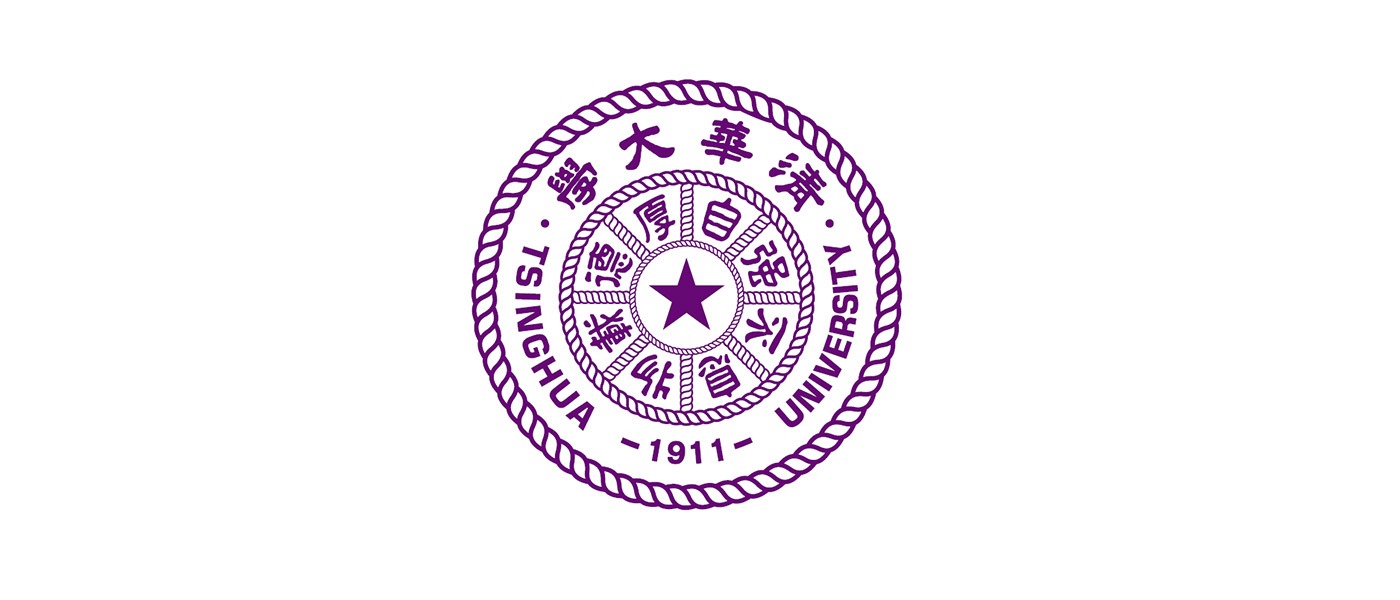}\hfill
  \institutionLogo[trim=0bp 0bp 0bp 20.16bp,clip,height=0.94cm]{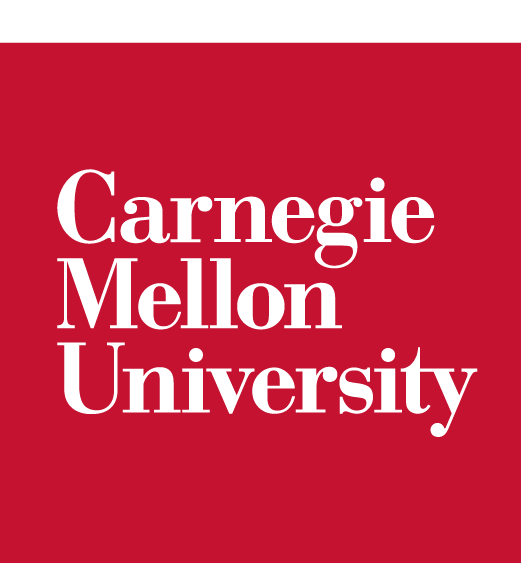}\par
  \endgroup
}
\patchcmd{\mymaketitle}{\tcbset{top=0.5cm}}{\tcbset{top=0.15cm}}{}{}
\patchcmd{\mymaketitle}{\tcbset{bottom=0.5cm}}{\tcbset{bottom=0.15cm}}{}{}
\patchcmd{\mymaketitle}{\setlength{\parskip}{0.5cm}}{\setlength{\parskip}{0.18cm}}{}{}
\patchcmd{\mymaketitle}{\vskip 0.5cm}{\vskip 0.1cm}{}{}
\patchcmd{\mymaketitle}{\authorlist\par}
  {{\fontsize{10}{12}\selectfont\authorlist\par}}{}
  {\PackageError{author-layout}{Author block patch failed}{Check fairmeta.cls}}
\patchcmd{\mymaketitle}{\affiliationlist\par}
  {{\fontsize{8}{10}\selectfont\affiliationlist\par}\vspace{3pt}}{}
  {\PackageError{author-layout}{Affiliation block patch failed}{Check fairmeta.cls}}
\patchcmd{\mymaketitle}{\contributionlist\par}
  {\contributionlist\par\institutionLogos}{}
  {\PackageError{author-layout}{Logo block patch failed}{Check fairmeta.cls}}
\patchcmd{\mymaketitle}{\tcbset{colback=metabg}}{\tcbset{colback=white}}{}{}
\patchcmd{\mymaketitle}{\includegraphics[width=2.0cm]{assets/tuojing_logo.png}}{}{}{}

\patchcmd{\mymaketitle}{\abstractlist\par}{}{}{}
\abstract{}%
\metadata[Website]{\url{https://llm4robot.github.io/bench/}}
\metadata[Code]{\url{https://github.com/LLM4Robot/LLM4Robot-Bench}}

\hypersetup{
  pdftitle={LLM4Robot-Bench: How Far Can GPT-6 Astra Go in Robotic Manipulation? A Hierarchical, Paired Sim--Real Benchmark},
  pdfsubject={A hierarchical, paired simulation and real-robot benchmark for physical interaction},
  pdfkeywords={LLM4Robot-Bench, robotic manipulation, contact-rich manipulation, sim-to-real, benchmark}
}

\begin{document}
\maketitle 
\input{figures/figure01_main_finding}

\begin{abstractbox}
\abstracttitle
\input{sec/0_abstract}

\end{abstractbox}

\Needspace{8\baselineskip}
\input{sec/1_introduction}

\FloatBarrier
\Needspace{10\baselineskip}
\input{sec/2_related_work}

\FloatBarrier

\input{sec/3_benchmark_design}
\input{sec/4_aligned_environments}
\input{sec/5_evaluating_gpt6_astra}

\input{sec/6_results_and_analysis}

\input{sec/7_discussion_and_limitations}
\input{sec/8_conclusion}

\bibliography{references/main}
\bibliographystyle{package/bibstyle}

\clearpage
\appendix
\input{sec/supplement_format}
\input{sec/A_task_specifications}

\input{sec/B_real_world_setup}

\input{sec/C_simulation_and_assets}

\input{sec/E_evaluation_specifications}

\end{document}

%% file: package/preamble.tex
\input{package/math_commands.tex}

\usepackage{graphicx}
\usepackage[table]{xcolor}
\usepackage{colortbl}

\definecolor{PLInk}{HTML}{1C2B33}
\definecolor{PLMuted}{HTML}{56677A}
\definecolor{PLLine}{HTML}{D5DFE6}
\definecolor{PLFill}{HTML}{F4F6F8}
\definecolor{PLNavy}{HTML}{172B49}
\definecolor{PLSim}{HTML}{5266A5}
\definecolor{PLReal}{HTML}{B86242}
\definecolor{PLLink}{HTML}{2B4C8C}
\definecolor{PVBlue}{HTML}{1B6EF2}
\definecolor{PVDeep}{HTML}{0E56AA}
\definecolor{PVNavy}{HTML}{0B1F44}
\definecolor{PVInk}{HTML}{172338}
\definecolor{PVMuted}{HTML}{5A6A82}
\definecolor{PVLine}{HTML}{C9D7EC}
\definecolor{PVSky}{HTML}{EDF4FF}
\definecolor{PVSky2}{HTML}{D9E7FE}
\definecolor{PVSim}{HTML}{6D55F5}
\definecolor{PVReal}{HTML}{F26B1D}
\definecolor{PVL1}{HTML}{2B88EA}
\definecolor{PVL2}{HTML}{1B5FDD}
\definecolor{PVL3}{HTML}{0B2A6E}
\definecolor{PVGood}{HTML}{14A05A}
\definecolor{PVBad}{HTML}{E5484D}
\AtBeginDocument{\hypersetup{linkcolor=PLLink, citecolor=PLLink, urlcolor=PLLink}}

\newif\ifshowrevisions
\showrevisionsfalse
\DeclareRobustCommand{\revision}[1]{%
  \ifshowrevisions
    {\color{magenta}\hypersetup{linkcolor=magenta}#1}%
  \else
    #1%
  \fi
}
\usepackage{booktabs}
\usepackage{tabularx}
\usepackage{array}
\usepackage{multirow}
\usepackage{diagbox}
\usepackage{hhline}
\usepackage{longtable}
\usepackage{makecell}
\usepackage{siunitx}
\usepackage{adjustbox}
\usepackage{wrapfig}
\usepackage{caption}   %
\newcolumntype{C}{>{\centering\arraybackslash}X}

\usepackage{amsmath,amsfonts,amssymb}
\usepackage{bm}
\usepackage{nicefrac}

\usepackage{enumitem}
\setlist[itemize]{leftmargin=*}

\usepackage{caption}

\crefname{figure}{Fig.}{Figs.}
\crefname{table}{Tab.}{Tabs.}

\usepackage{xspace}
\usepackage{calc}
\usepackage{etoolbox}
\usepackage{pifont}
\usepackage{fancyvrb}
\usepackage{needspace}

\definecolor{stepnavy}{HTML}{172B49}
\makeatletter
\newdimen\stepmark@r
\newdimen\stepmark@cy
\newdimen\stepmark@fs
\DeclareRobustCommand{\stepmark}[1]{%
  \begingroup
  \stepmark@r=0.875ex\relax                    %
  \stepmark@cy=0.5\fontcharht\font`H\relax  %
  \stepmark@fs=0.72\dimexpr\f@size pt\relax    %
  \kern0.04em
  \tikz[baseline=0pt]{%
    \draw[stepnavy, line width=0.45pt] (0pt,\stepmark@cy) circle[radius=\stepmark@r];
    \node[inner sep=0pt, text=stepnavy,
      font=\normalfont\sffamily\fontsize{\stepmark@fs}{\stepmark@fs}\selectfont]
      at (0pt,\stepmark@cy) {#1};}%
  \kern0.04em
  \endgroup}
\makeatother

\usepackage{titletoc}

\titlecontents{section}
[1.5em] %
{\addvspace{-0.5pt}} %
{\bfseries\contentslabel{2.3em}} %
{\hspace*{-2.3em}\bfseries} %
{\bfseries\titlerule*[.5pc]{.}\contentspage} %
\titlecontents{subsection}
[3.8em] %
{\addvspace{-2.2pt}} %
{\contentslabel{2.3em}}
{\hspace*{-2.3em}}
{\titlerule*[.5pc]{.}\contentspage}

\def\x{$\times$}

\definecolor{lightgray}{rgb}{0.95, 0.95, 0.95}

\definecolor{baselinecolor}{gray}{.9}

\usetikzlibrary{arrows.meta}
\usetikzlibrary{shadows.blur, fadings}
\usepackage{fontawesome5}

\colorlet{abstractframe}{PLNavy!38}
\newtcolorbox{abstractbox}{enhanced, colback=white, colframe=abstractframe, boxrule=0.6pt, arc=6pt, left=12pt, right=12pt, top=6pt, bottom=8pt, before skip=8pt, after skip=14pt}
\newcommand{\abstracttitle}{{\centering\large\bfseries\color{PLNavy}Abstract\par}\vspace{4pt}}

%% file: package/math_commands.tex
\usepackage{amsmath,amsfonts,bm}
\usepackage{xcolor}

\def\eqref#1{equation~\ref{#1}}

\def\1{\bm{1}}

\DeclareMathAlphabet{\mathsfit}{\encodingdefault}{\sfdefault}{m}{sl}
\SetMathAlphabet{\mathsfit}{bold}{\encodingdefault}{\sfdefault}{bx}{n}

%% file: figures/figure01_main_finding.tex
\input{tables/evaluation_result_cells}
\par\medskip\noindent
\begin{minipage}{\linewidth}
\captionsetup{type=figure}
\centering
\begingroup
\input{figures/comparison_figure_style}
\makeatletter
\def\PL@pct#1\%#2\@nil{#1}
\newcommand{\PLPct}[2]{\expandafter\expandafter\expandafter\PL@pct\csname AstraResult-#1-#2\endcsname\@nil}
\makeatother
\tikzset{
  fo brick/.style={rounded corners=2pt, minimum width=82pt, minimum height=11.5pt, inner sep=0pt,
    anchor=south west, font=\PVSans\fontseries{sb}\fontsize{7}{7}\selectfont},
  fo new/.style={fo brick, fill=white},
  fo kept/.style={fo brick, fill=white, fill opacity=0.24, text opacity=1, text=white},
  fo head/.style={inner sep=0pt, anchor=base west, text=white, font=\PVMono\fontseries{b}\fontsize{8.5}{9}\selectfont},
  fo count/.style={inner sep=0pt, anchor=base east, text=white, font=\PVMono\fontsize{7}{8}\selectfont},
}
\newcommand{\FoY}[2]{34+0.9*\PLPct{#1}{#2}}
\begin{tikzpicture}[x=1pt,y=1pt]
\path[pv card] (0,172) rectangle (470,266);
\PVPanelTitle{8}{252.5}{a}{Paired Sim\textendash Real Evaluation}
\node[pv muted, anchor=base east] at (462,252.5) {16 tasks\enspace\textperiodcentered\enspace 3 levels\enspace\textperiodcentered\enspace 3 trials per task and environment};
\PVFitPhoto{10}{176}{124}{70}{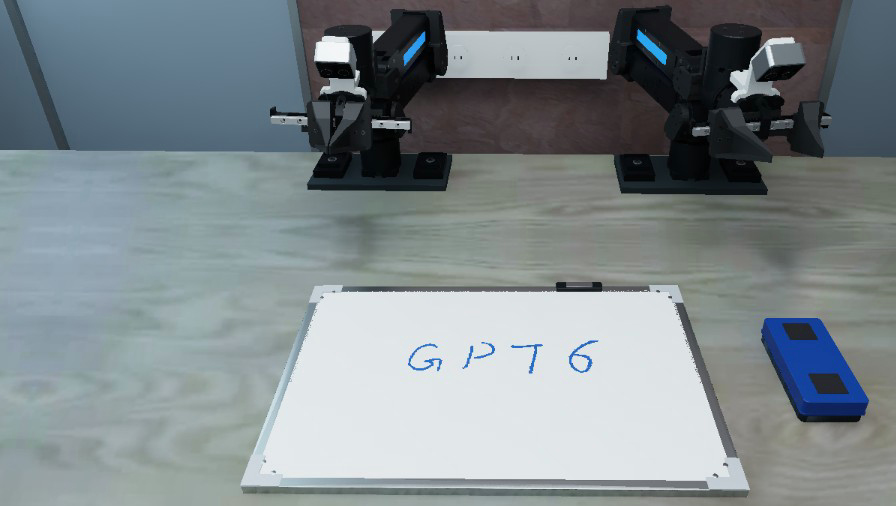}
\PVFitPhoto{336}{176}{124}{70}{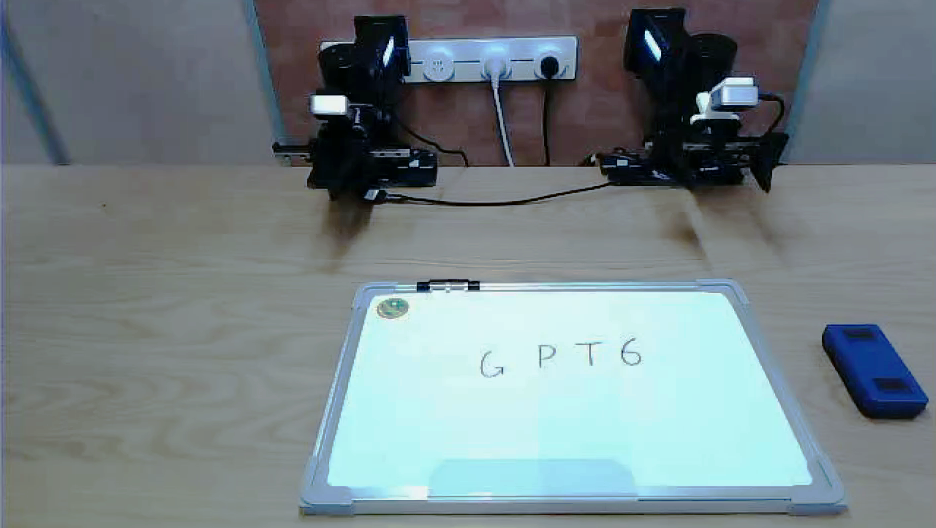}
\node[pv chip, fill=PVSim, anchor=north west] at (14,242) {SIMULATION};
\node[pv chip, fill=PVReal, anchor=north west] at (340,242) {REAL WORLD};
\path[rounded corners=6pt, top color=PVBlue!88!PVDeep, bottom color=PVNavy,
  blur shadow={shadow xshift=0pt, shadow yshift=-1.2pt, shadow blur radius=3pt, shadow opacity=35}]
  (176,200) rectangle (294,238);
\node[inner sep=0pt, text=white, font=\PVMono\fontseries{b}\fontsize{12.5}{14}\selectfont] at (235,224.5) {GPT-6 Astra};
\node[rounded corners=1.8pt, fill=white, fill opacity=0.2, text opacity=1, text=white,
  inner xsep=3pt, inner ysep=1.5pt, font=\PVMono\fontsize{7}{7}\selectfont] at (235,209) {zero-shot};
\draw[pv arrow, PVSim] (172,219) -- (140,219);
\draw[pv arrow, PVReal] (298,219) -- (330,219);
\node[pv muted, anchor=base] at (235,185) {Shared tasks and success criteria};
\path[pv card] (0,0) rectangle (296,164);
\PVPanelTitle{8}{150.5}{b}{Interaction Hierarchy}
\node[pv muted, anchor=base east] at (288,150.5) {Cumulative interaction requirements};
\path[rounded corners=3pt, top color=PVL1!80!white, bottom color=PVL1] (10,8) rectangle (100,40);
\path[rounded corners=3pt, top color=PVL2!82!white, bottom color=PVL2] (103,8) rectangle (193,54);
\path[rounded corners=3pt, top color=PVL3!80!white, bottom color=PVL3] (196,8) rectangle (286,68);
\node[fo new, text=PVL1] at (14,12) {Selection \& alignment};
\node[fo kept] at (107,12) {Selection \& alignment};
\node[fo new, text=PVL2] at (107,26) {+ Force regulation};
\node[fo kept] at (200,12) {Selection \& alignment};
\node[fo kept] at (200,26) {+ Force regulation};
\node[fo new, text=PVL3] at (200,40) {+ Reconfiguration};
\node[fo head] at (14,29.5) {L1};
\node[fo count] at (96,29.5) {5 tasks};
\node[fo head] at (107,43.5) {L2};
\node[fo count] at (189,43.5) {7 tasks};
\node[fo head] at (200,57.5) {L3};
\node[fo count] at (282,57.5) {4 tasks};
\PVPhoto{13}{44}{84}{63}{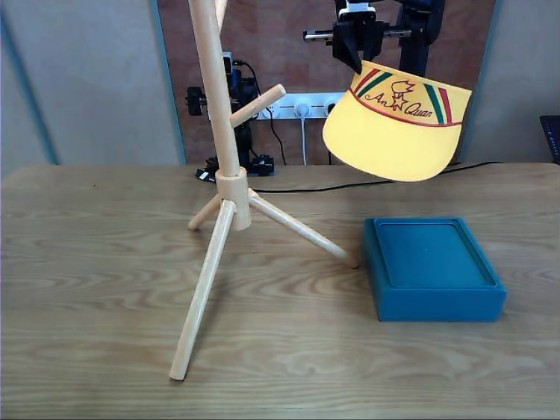}
\PVPhoto{106}{58}{84}{63}{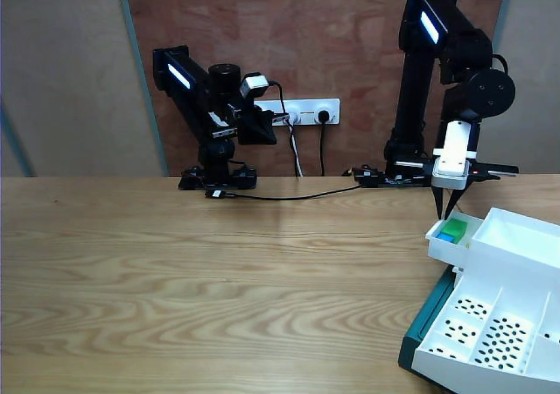}
\PVPhoto{199}{72}{84}{63}{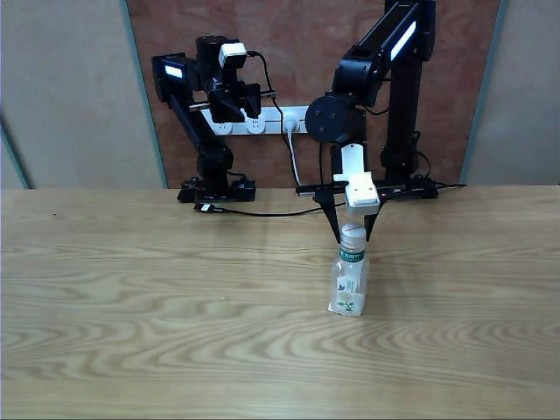}
\node[pv glass, anchor=south west] at (16,47) {Hat Hanging};
\node[pv glass, anchor=south west] at (109,61) {Drawer Pull \& Return};
\node[pv glass, anchor=south west] at (202,75) {Bottle Cap Unscrewing};
\draw[PVBlue!65, line width=0.9pt, dash pattern=on 2.4pt off 1.8pt, line cap=round, line join=round,
  -{Stealth[length=4.6pt,width=4.2pt,round]}] (38,114) -- (130,128) -- (222,142);
\path[pv card] (304,0) rectangle (470,164);
\PVPanelTitle{312}{150.5}{c}{Success by Level}
\draw[PVReal, line width=1.3pt] (316,138) -- (330,138);
\fill[PVReal] (323,138) circle (2.3pt);
\node[pv text, anchor=west] at (333,138) {Real world};
\draw[PVSim, line width=1.3pt, dash pattern=on 2.6pt off 1.6pt] (380,138) -- (394,138);
\draw[PVSim, fill=white, line width=1pt] (387,138) circle (2.1pt);
\node[pv text, anchor=west] at (397,138) {Simulation};
\fill[PVBad!7] (419,34) rectangle (462,124);
\foreach \v/\lab in {0/0\%, 50/50\%, 100/100\%} {
  \draw[PVLine, line width=0.45pt] (332,{34+0.9*\v}) -- (462,{34+0.9*\v});
  \node[pv muted, font=\PVMono\fontsize{7}{7}\selectfont, anchor=east] at (329,{34+0.9*\v}) {\lab};}
\draw[PVBad, line width=0.7pt, dash pattern=on 2.6pt off 2pt] (419,32) -- (419,126);
\node[pv text, text=PVBad, align=center, anchor=north, font=\PVSans\fontseries{sb}\fontsize{7}{8.2}\selectfont] at (440.5,121.5) {No L3 task\\completed};
\draw[PVSim, line width=1.3pt, dash pattern=on 2.8pt off 1.8pt, line join=round]
  (350,{\FoY{Main-L1}{sim-zs}}) -- (396,{\FoY{Main-L2}{sim-zs}}) -- (442,{\FoY{Main-L3}{sim-zs}});
\draw[PVReal, line width=1.3pt, line join=round]
  (350,{\FoY{Main-L1}{real-main}}) -- (396,{\FoY{Main-L2}{real-main}}) -- (442,{\FoY{Main-L3}{real-main}});
\foreach \lv/\x in {L1/350, L2/396, L3/442} {
  \fill[PVReal, draw=white, line width=0.6pt] (\x,{\FoY{Main-\lv}{real-main}}) circle (2.7pt);
  \draw[PVSim, fill=white, line width=1pt] (\x,{\FoY{Main-\lv}{sim-zs}}) circle (2.2pt);}
\node[pv chip, fill=PVL1] at (350,23) {L1};
\node[pv chip, fill=PVL2] at (396,23) {L2};
\node[pv chip, fill=PVL3] at (442,23) {L3};
\node[pv value, text=PVReal, anchor=south] at (350,{\FoY{Main-L1}{real-main}+4.5}) {\PLPct{Main-L1}{real-main}\%};
\node[pv value, text=PVSim, anchor=east] at (345.5,{\FoY{Main-L1}{sim-zs}}) {\PLPct{Main-L1}{sim-zs}\%};
\node[pv value, text=PVInk, anchor=north east] at (394,{\FoY{Main-L2}{real-main}-3}) {\PLPct{Main-L2}{real-main}\%};
\node[pv value, text=PVInk, anchor=south] at (442,{\FoY{Main-L3}{real-main}+4.5}) {0\%};
\node[rounded corners=5pt, fill=PVSky2, text=PVDeep, inner xsep=4pt, inner ysep=2pt,
  font=\PVSans\fontseries{sb}\fontsize{7}{8}\selectfont] at (387,8.5) {10 of 16 tasks: identical success counts};
\end{tikzpicture}
\arrayrulecolor{black}
\endgroup
\caption{\textbf{LLM4Robot-Bench: hierarchical, paired Sim--Real evaluation of GPT-6 Astra's contact-rich manipulation capabilities.}
(a) Sixteen tasks pair physically and visually matched simulated and real-world environments with shared task objectives and success criteria.
(b) Three cumulative levels progress from contact selection and geometric alignment (L1), to force regulation within sustained contact (L2), and coordinated contact reconfiguration and force regulation (L3). The photographs illustrate task requirements.
(c) \textbf{Success declines across levels in both environments, with no L3 task completed.} Rates average task-level success equally, using three trials per task in each environment. \textbf{Ten of the 16 tasks have identical Sim--Real success counts.} These shared trends support matched simulation as a repeatable testbed for studying Astra's limits in contact-rich manipulation.}
\label{fig:main-finding}
\end{minipage}
\par\medskip

%% file: sec/0_abstract.tex
\looseness=-1
We introduce LLM4Robot-Bench, a hierarchical, paired Sim--Real benchmark for evaluating physical interaction capabilities in contact-rich manipulation. Its 16 household tasks span three cumulative levels: contact selection and geometric alignment (L1), force regulation within sustained contact (L2), and coordinated contact reconfiguration and force regulation (L3). Each task is instantiated in physically and visually matched simulated and real-world environments, with shared task definitions and success criteria. We evaluate GPT-6 Astra zero-shot with three trials per task in each environment. Real-world success rates decline from 80.0\% at L1 to 47.6\% at L2 and 0.0\% at L3, with corresponding simulation rates of 66.7\%, 47.6\%, and 0.0\%. Ten of the 16 tasks have identical success counts across environments; the remaining six differ by one success out of three trials. Across both environments, recurring difficulties in spatial reasoning, contact selection, force regulation, and coordination often prevent local physical progress from accumulating into task completion. These findings characterize Astra's observed capability boundary across increasing interaction demands and, for this model, support matched simulation as a repeatable, scalable testbed for studying the sustained, coordinated physical interactions that limit reliable manipulation.

%% file: sec/1_introduction.tex
\section{Introduction}
\label{sec:introduction}

GPT-6 Astra and other general-purpose foundation models have shown promise for robotic manipulation, while reliable execution remains challenging~\citep{robodojo2026astra,yang2026robotworld,huang2026robocoliseum}. Recent work explores adaptation from demonstrations and interaction feedback without parameter updates~\citep{cheng2026gptpolicy,liu2026roboicl}, as well as iterative program development for long-horizon, contact-rich tasks~\citep{shen2026embodiedswe}. Practical use in homes and factories, however, requires reliable contact-rich execution: establishing suitable contacts, regulating forces through them, and coordinating contact changes as a task proceeds~\citep{suomalainen2022survey,forge2024}. Aggregate success rates alone do not identify which physical interaction requirements constrain performance. We therefore ask: which requirements can Astra satisfy as interactions become more demanding, and to what extent do the observed capabilities and limitations persist across simulation and reality?

To address these questions, we introduce LLM4Robot-Bench, a hierarchical, paired Sim--Real benchmark that organizes 16 household manipulation tasks by their physical interaction requirements (Figure~\ref{fig:task-suite}; Section~\ref{sec:benchmark-design}). The hierarchy is organized around two dimensions: contact configuration and force regulation. L1 emphasizes contact selection and geometric alignment. L2 adds task-specific force regulation within sustained contact, directly or through a tool or container, without requiring contact reconfiguration to continue the operation. L3 additionally requires coordinating contact reconfiguration with force regulation. Levels describe task versions rather than task names: blocks clustered at a dustpan entrance can be swept in one stroke (L2), whereas dispersed blocks require successive contacts (L3). Spatial reasoning and execution precision remain relevant throughout. By relating performance to explicit interaction requirements, the hierarchy supports analysis of capabilities and limitations and provides successive evaluation targets beyond grasping and placement.

\input{figures/figure02_task_suite}

LLM4Robot-Bench provides physically and visually matched real-world and simulated environments for each task (Figure~\ref{fig:main-finding}). Real-world trials provide a reference for physical execution, while simulation supports controlled, repeatable evaluation at scale. Using the physical setup as the reference, we match workspace layouts, object geometry and appearance, robot embodiment, and camera views. Physical measurements and observed contact responses guide simulation calibration (Section~\ref{sec:aligned-environments}). This correspondence supports comparisons of task completion, intermediate progress, and recurring interaction difficulties. Corresponding task definitions and documented success criteria guide these comparisons.

\looseness=-1
We evaluate GPT-6 Astra zero-shot with three trials per task in each environment. The evaluation yields three main findings. First, reported success rates decline across the interaction levels: from 80.0\% at L1 to 47.6\% at L2 and 0.0\% at L3 in the real world, with corresponding simulation rates of 66.7\%, 47.6\%, and 0.0\%. Second, local physical progress does not reliably accumulate into task completion. Astra may turn a cap or move a drawer without completing the required operation. Execution analysis identifies four recurring interaction bottlenecks: spatial reasoning and execution precision, contact configuration selection and establishment, force regulation, and coordination within and across interactions (Section~\ref{sec:failure-analysis}). Third, for GPT-6 Astra, the two environments show consistent performance trends and similar interaction difficulties. Ten of the 16 tasks have identical reported success counts, and the remaining six differ by one successful trial. Together, these findings characterize Astra's observed limitations under increasing interaction demands and support matched simulation as a testbed for studying the sustained, coordinated physical interactions that constrain reliable manipulation.

Our main contributions are:
\begin{itemize}
    \item \textbf{LLM4Robot-Bench, a hierarchical benchmark for physical interaction.} We define three cumulative levels based on contact configuration and force regulation, and organize 16 household tasks according to the requirements of each task version, providing successive evaluation targets beyond grasping and placement.
    \Needspace{7\baselineskip}
    \item \textbf{Physically and visually matched Sim--Real environments.} We construct simulated counterparts of physical tasks through scene reconstruction and calibration against observed contact responses, with corresponding task definitions and documented success criteria for comparison across environments.
    \item \textbf{A systematic zero-shot evaluation of GPT-6 Astra.} We characterize task completion, intermediate progress, and four recurring interaction bottlenecks across levels and environments, identifying the physical interaction requirements that challenge reliable execution.
\end{itemize}

%% file: figures/figure02_task_suite.tex
\begingroup
\input{figures/comparison_figure_style}
\tikzset{
  ft name/.style={inner sep=0pt, anchor=base west, text=PVInk,
    font=\PVSans\fontseries{sb}\fontsize{7.5}{9}\selectfont},
  ft legend/.style={inner sep=0pt, anchor=base west, text=PVInk, font=\PVSans\fontsize{7.5}{9}\selectfont},
}
\newcommand{\FtTask}[5]{%
  \pgfmathsetmacro\FtX{#1*(112+22/3)}%
  \pgfmathsetmacro\FtT{389-#2*57.5}%
  \edef\FtCC{\PVLevelColor{#3}}%
  \node[pv chip, fill=\FtCC, anchor=base west] at (\FtX,\FtT-7.4) {#3};
  \ifnum#2=3
    \def\FtDrop{9}%
    \node[ft name, anchor=north west, align=left, text width=82.5pt] at (\FtX+29.5,\FtT-1.4) {#4};
  \else
    \def\FtDrop{0}%
    \node[ft name] at (\FtX+29.5,\FtT-7.4) {#4};
  \fi
  \PVFitPhoto{\FtX}{\FtT-51.6-\FtDrop}{54.5}{40.9}{assets/figure02_vivid/#5_sim.jpg}
  \PVFitPhoto{\FtX+57.5}{\FtT-51.6-\FtDrop}{54.5}{40.9}{assets/figure02_vivid/#5_real.jpg}
  \fill[PVSim, rounded corners=0.8pt] (\FtX,\FtT-54.4-\FtDrop) rectangle ++(54.5,1.7);
  \fill[PVReal, rounded corners=0.8pt] (\FtX+57.5,\FtT-54.4-\FtDrop) rectangle ++(54.5,1.7);}
\newcommand{\FtGroup}[5]{%
  \pgfmathsetmacro\FtX{#1*245}%
  \pgfmathsetmacro\FtT{131.5-#2*70.25}%
  \edef\FtCC{\PVLevelColor{#3}}%
  \node[pv chip, fill=\FtCC, anchor=base west] at (\FtX,\FtT-7.4) {#3};
  \node[ft name] at (\FtX+29.5,\FtT-7.4) {#4};
  \foreach \k/\dx in {1/0, 2/76, 3/152} {
    \PVFitPhoto{\FtX+\dx}{\FtT-65.25}{73}{54.75}{assets/figure02_vivid/#5_configuration_0\k.jpg}
    \node[pv glass, font=\PVMono\fontseries{b}\fontsize{7}{7}\selectfont, inner xsep=2.2pt, inner ysep=1.4pt,
      anchor=north west] at (\FtX+\dx+2.5,\FtT-13) {\k};}}
\begin{figure*}[!t]
\centering
\begin{tikzpicture}[x=1pt,y=1pt]
\path[use as bounding box] (0,-9) rectangle (470,407);
\PVPanelTitle{0}{396.5}{a}{Contact-Rich Task Suite}
\node[pv chip, fill=PVL1, anchor=base west] (lv1) at (196,396.5) {L1};
\node[ft legend] at ([xshift=2.5pt]lv1.base east) {5 tasks};
\node[pv chip, fill=PVL2, anchor=base west] (lv2) at (243,396.5) {L2};
\node[ft legend] at ([xshift=2.5pt]lv2.base east) {7 tasks};
\node[pv chip, fill=PVL3, anchor=base west] (lv3) at (290,396.5) {L3};
\node[ft legend] at ([xshift=2.5pt]lv3.base east) {4 tasks};
\fill[PVSim, rounded corners=0.8pt] (352,398.2) rectangle ++(12,1.9);
\node[ft legend] at (367.5,396.5) {Simulation};
\fill[PVReal, rounded corners=0.8pt] (414,398.2) rectangle ++(12,1.9);
\node[ft legend] at (429.5,396.5) {Real world};
\FtTask{0}{0}{L1-01}{Marker Placement}{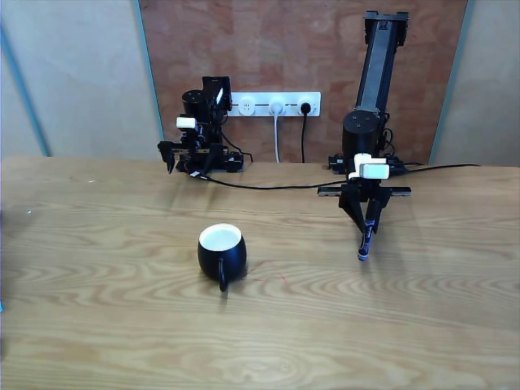}
\FtTask{1}{0}{L1-02}{Charger Alignment}{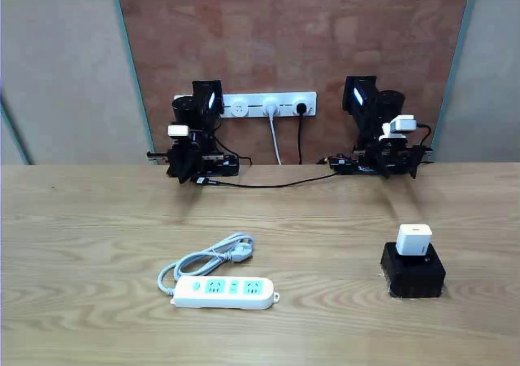}
\FtTask{2}{0}{L2-01}{Marker Cap Removal}{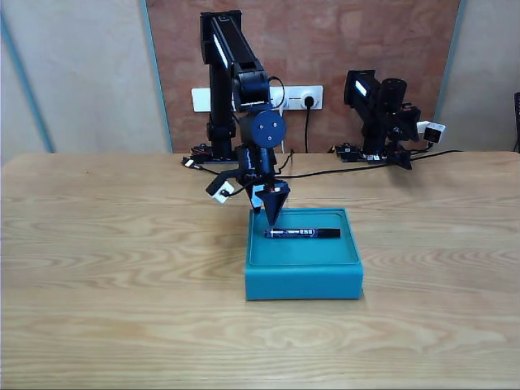}
\FtTask{3}{0}{L2-02}{Book Pulling}{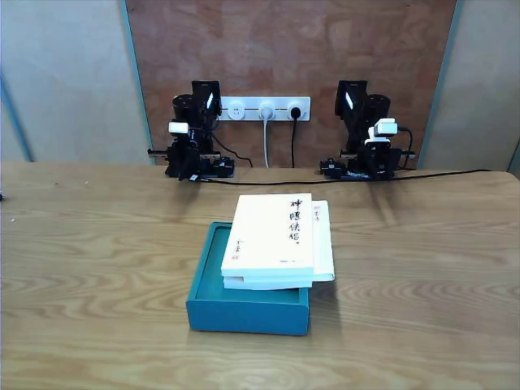}
\FtTask{0}{1}{L1-03}{Hat Hanging}{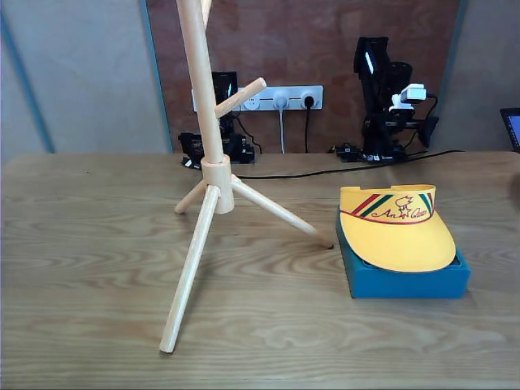}
\FtTask{1}{1}{L1-04}{Cup Nesting}{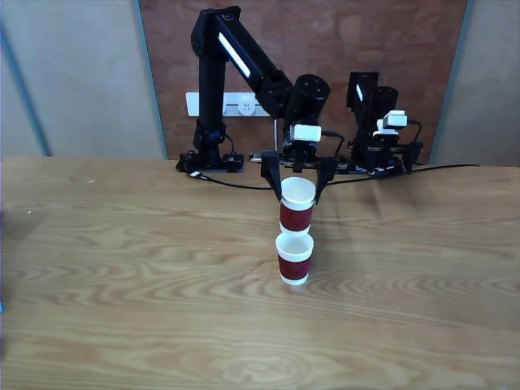}
\FtTask{2}{1}{L2-03}{Marble Pouring}{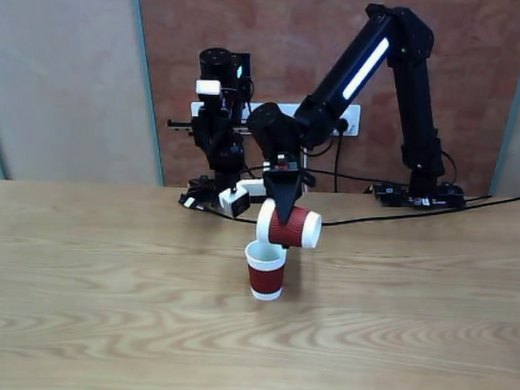}
\FtTask{3}{1}{L2-04}{Orange Handling}{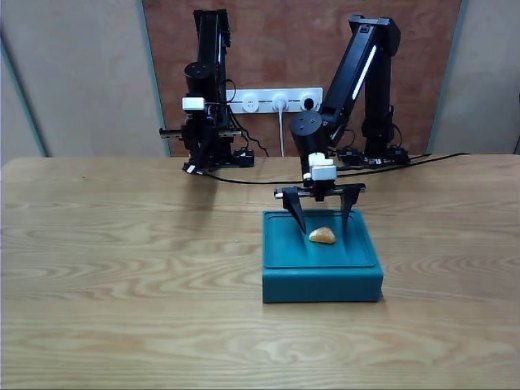}
\FtTask{0}{2}{L1-05}{Flower Transfer}{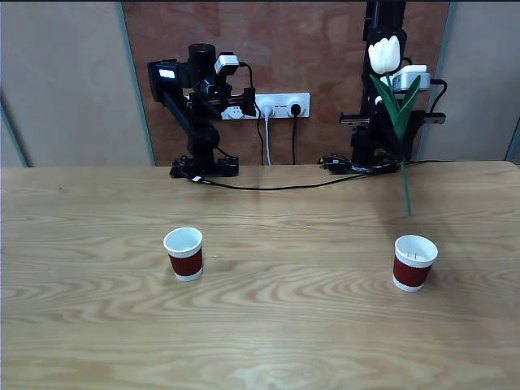}
\FtTask{1}{2}{L2-05}{Drawer Pull \& Return}{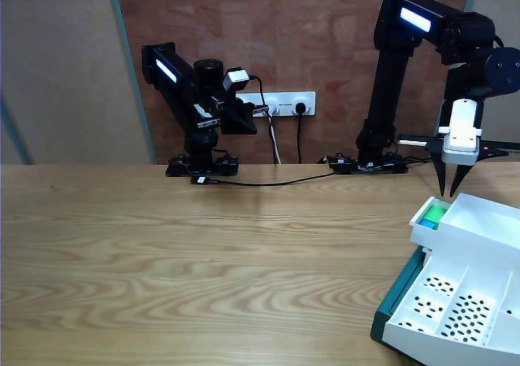}
\FtTask{2}{2}{L2-06}{Jelly Handling}{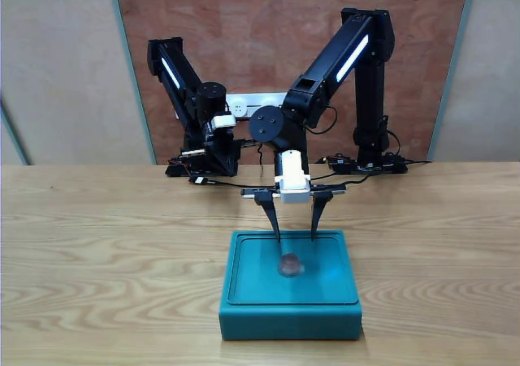}
\FtTask{3}{2}{L2-07}{Whiteboard Wiping}{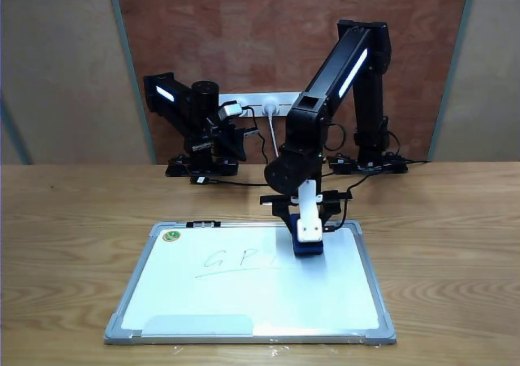}
\FtTask{0}{3}{L3-01}{Bottle Cap Unscrewing}{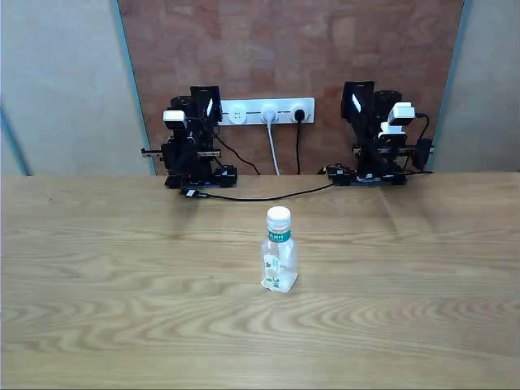}
\FtTask{1}{3}{L3-02}{Nut Unscrewing}{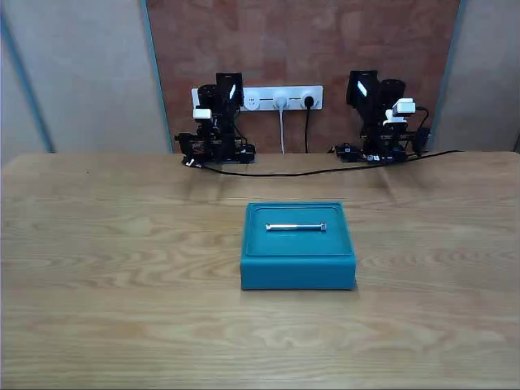}
\FtTask{2}{3}{L3-03}{Stacked-Book\\Pushing \& Rotation}{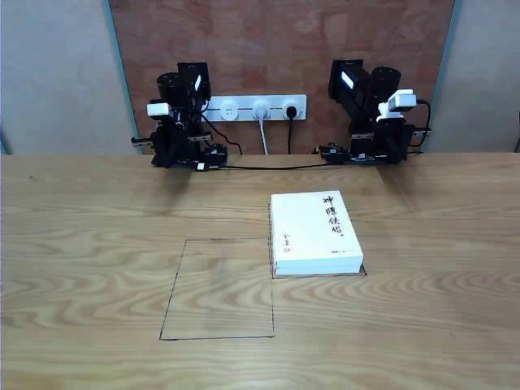}
\FtTask{3}{3}{L3-04}{Block Sweeping}{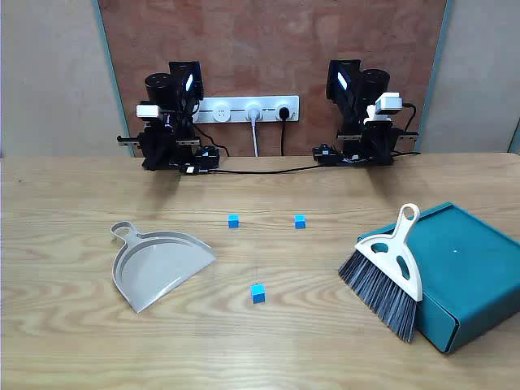}
\PVPanelTitle{0}{139.5}{b}{Initial-State Variations}
\node[pv chip, fill=PVSim, anchor=base east] at (470,139.5) {SIMULATION};
\FtGroup{0}{0}{L1-01}{Marker Placement}{marker}
\FtGroup{1}{0}{L2-03}{Marble Pouring}{marbles}
\FtGroup{0}{1}{L2-01}{Marker Cap Removal}{uncapping}
\FtGroup{1}{1}{L3-04}{Block Sweeping}{sweeping}
\end{tikzpicture}
\caption{\textbf{Contact-rich manipulation tasks and representative initial-state variations.}
(a) The 16 tasks span three interaction levels, identified by task tags colored by level:
L1 (5 tasks) emphasizes contact selection and geometric alignment;
L2 (7 tasks) adds force regulation within sustained contact;
L3 (4 tasks) requires coordinated contact reconfiguration and force regulation.
Each task unit pairs a simulated view (left, violet bar) with a real-world view (right, orange bar).
(b) Four representative simulation tasks are each shown in three initial configurations, illustrating variations in object positions, relative placement, and orientation. Marker Cap Removal is shown with horizontal, vertical, and diagonal marker placements. Task tags follow the same L1--L3 grouping and task IDs in both panels.
Task specifications are detailed in Appendix~\ref{app:task-specifications}.}
\label{fig:task-suite}
\end{figure*}
\arrayrulecolor{black}
\endgroup

%% file: figures/comparison_figure_style.tex
\colorlet{PLInk}{PVInk}
\colorlet{PLMuted}{PVMuted}
\colorlet{PLLine}{PVLine}
\colorlet{PLFill}{PVSky}
\colorlet{PLNavy}{PVDeep}
\colorlet{PLSim}{PVSim}
\colorlet{PLReal}{PVReal}
\colorlet{ComparisonInk}{PVInk}
\colorlet{ComparisonMuted}{PVMuted}
\colorlet{ComparisonLine}{PVLine}
\colorlet{ComparisonFill}{PVSky}
\colorlet{ComparisonSim}{PVSim}
\colorlet{ComparisonReal}{PVReal}
\input{figures/physladder_vivid_style}
\newcommand{\ComparisonFont}{\PVSans}
\makeatletter
\newcommand{\PLTaskChip}[1]{\begingroup\edef\PV@cc{\PVLevelColor{#1}}%
  \tikz[baseline=(PLchip.base)]\node[pv chip, fill=\PV@cc] (PLchip) {#1};\endgroup}
\def\PL@count#1(#2)#3\@nil{#2\PL@dagger#3\PL@enddagger}
\def\PL@dagger#1\PL@enddagger{\in@{^}{#1}\ifin@\textsuperscript{\dag}\fi}
\newcommand{\PLCount}[2]{\expandafter\expandafter\expandafter\PL@count
  \csname AstraResult-#1-#2\endcsname(--)\@nil}
\makeatother
\newcommand{\PLPill}[2]{\tikz[baseline=(PLpill.base)]\node[rounded corners=3.6pt, fill=#1!13,
  text=#1!88!black, inner xsep=2.6pt, inner ysep=1.7pt,
  font=\PVMono\fontseries{b}\fontsize{7}{7}\selectfont] (PLpill) {#2};}
\newcommand{\PLTally}[1]{\PLPill{PVSim}{SIM\,\PLCount{#1}{sim-zs}}\hspace{2.5pt}%
  \PLPill{PVReal}{REAL\,\PLCount{#1}{real-skills}}}
\newcommand{\PLLetterTile}[1]{\tikz[baseline=(PLlt.base)]\node[pv letter, text height=6.6pt,
  text depth=0pt, minimum width=10pt, minimum height=10pt, inner sep=0pt] (PLlt) {#1};}
\newcommand{\PLPanelHead}[5]{%
  \par\noindent\parbox{\linewidth}{%
    {\PVMono\fontseries{b}\fontsize{9}{11}\selectfont\color{PVBlue}%
      \PLLetterTile{#1}\hspace{0.45em}%
      \if\relax\detokenize{#2}\relax\else\PLTaskChip{#2}\hspace{0.45em}\fi
      #3\hfill{\normalfont\PVSans\fontsize{7.5}{9}\selectfont#5}\par}%
    \if\relax\detokenize{#4}\relax\else
      \vspace{1.5pt}{\PVSans\fontsize{7.5}{9}\selectfont\color{PVMuted}#4\par}%
    \fi}\par}
\makeatletter
\newcommand{\PLEnvLabel}[1]{%
  \ifstrequal{#1}{sim}{\def\PL@envcolor{PVSim}\def\PL@envtext{SIMULATION}}%
    {\def\PL@envcolor{PVReal}\def\PL@envtext{REAL WORLD}}%
  \parbox{\linewidth}{\leavevmode\color{\PL@envcolor}%
    \leaders\hrule height 3.1pt depth -2.4pt\hfill\hspace{4pt}%
    \tikz[baseline=(PLenv.base)]\node[pv chip, fill=\PL@envcolor] (PLenv) {\PL@envtext};%
    \hspace{4pt}\leaders\hrule height 3.1pt depth -2.4pt\hfill\null}}
\makeatother
\newcommand{\PLEnvNote}[1]{\ifstrequal{#1}{sim}%
  {\tikz[baseline=(PLen.base)]\node[pv chip, fill=PVSim] (PLen) {SIMULATION};}%
  {\tikz[baseline=(PLen.base)]\node[pv chip, fill=PVReal] (PLen) {REAL WORLD};}}
\newcommand{\PLArrow}[2]{\raisebox{#2}[0pt][0pt]{\makebox[0pt][c]{%
  \tikz[baseline=-0.5ex]\draw[pv arrow, draw=#1, line width=1.1pt] (0,0) -- (9pt,0);}}}
\newcommand{\PLGutter}{\hspace{\dimexpr3mm-.25pt\relax}{\color{PVLine}\vrule width .5pt}%
  \hspace{\dimexpr3mm-.25pt\relax}}
\newcommand{\PLIcon}[1]{\textcolor{PVBlue}{\faIcon{#1}}\hspace{0.4em}}
\newcommand{\PLBottleneckSpatial}{\PLIcon{crosshairs}Spatial reasoning and execution precision}
\newcommand{\PLBottleneckContact}{\PLIcon{hand-paper}Contact configuration selection and establishment}
\newcommand{\PLBottleneckForce}{\PLIcon{compress-arrows-alt}Force regulation}
\newcommand{\PLBottleneckCoordination}{\PLIcon{sync-alt}Coordination within and across interactions}
\setlength{\parskip}{0pt}
\setlength{\tabcolsep}{0pt}
\renewcommand{\arraystretch}{1}
\arrayrulecolor{ComparisonLine}
\newcommand{\PVRoundImage}[2]{%
  \sbox\PVImgBox{\includegraphics[#1]{#2}}%
  \begin{tikzpicture}[baseline=0pt]
    \path[use as bounding box] (0,0) rectangle (\wd\PVImgBox,\ht\PVImgBox);
    \path[pv photo shadow, rounded corners=2.5pt] (0,0) rectangle (\wd\PVImgBox,\ht\PVImgBox);
    \begin{scope}\clip[rounded corners=2.5pt] (0,0) rectangle (\wd\PVImgBox,\ht\PVImgBox);
      \node[inner sep=0pt, anchor=south west] at (0,0) {\usebox\PVImgBox};\end{scope}
  \end{tikzpicture}}
\newcommand{\ComparisonHeading}[2]{%
  \par\noindent\parbox{\linewidth}{%
    {\PVMono\color{PVBlue}\fontsize{9.5}{10.5}\selectfont\bfseries #1\par}%
    \if\relax\detokenize{#2}\relax\else
      \vspace{1.5pt}{\PVSans\color{PVMuted}\fontsize{7.5}{9}\selectfont #2\par}%
    \fi}\par}
\newcommand{\ComparisonHalfHeading}[2]{%
  \par\noindent\parbox[t]{\linewidth}{%
    {\PVMono\color{PVBlue}\fontsize{9.5}{10.5}\selectfont\bfseries #1\par}%
    \vspace{1.5pt}{\PVSans\color{PVMuted}\fontsize{7.5}{9}\selectfont #2\par}}\par}
\newcommand{\ComparisonEnvironment}[1]{%
  {\PVSans\color{PVInk}\fontsize{8}{9.5}\selectfont\bfseries #1}}
\newcommand{\ComparisonState}[1]{%
  \par\vspace{2.5pt}{\PVSans\fontsize{7.5}{9}\selectfont\color{PVInk}#1\par}}
\newcommand{\ComparisonImage}[3]{%
  \begin{minipage}[t]{\linewidth}\centering
  \PVRoundImage{width=\linewidth,height=#1,keepaspectratio}{#2}%
  \ComparisonState{#3}\end{minipage}}
\newcommand{\ComparisonRule}{%
  \par\vspace{6pt}\noindent\textcolor{ComparisonLine}{\rule{\linewidth}{0.5pt}}\par\vspace{6pt}}

%% file: figures/physladder_vivid_style.tex
\newcommand{\PVMono}{\fontencoding{T1}\fontfamily{FiraMono-TLF}\selectfont}
\newcommand{\PVSans}{\fontencoding{T1}\fontfamily{FiraSans-TLF}\selectfont}
\tikzset{
  pv card/.style={draw=PVDeep, line width=0.75pt, rounded corners=5pt,
    top color=white, bottom color=PVSky,
    blur shadow={shadow xshift=0pt, shadow yshift=-1pt, shadow blur radius=2.5pt, shadow opacity=18}},
  pv title/.style={inner sep=0pt, anchor=base west, text=PVBlue,
    font=\PVMono\fontseries{b}\fontsize{10.5}{12}\selectfont},
  pv letter/.style={fill=PVBlue, text=white, rounded corners=2pt, inner sep=0pt,
    minimum width=10.5pt, minimum height=10.5pt,
    font=\PVSans\fontseries{b}\fontsize{8.5}{8.5}\selectfont},
  pv text/.style={inner sep=0pt, text=PVInk, font=\PVSans\fontsize{7}{8.4}\selectfont},
  pv muted/.style={inner sep=0pt, text=PVMuted, font=\PVSans\fontsize{7}{8.4}\selectfont},
  pv mono/.style={inner sep=0pt, text=PVInk, font=\PVMono\fontsize{7}{8.4}\selectfont},
  pv value/.style={inner sep=0pt, font=\PVMono\fontseries{b}\fontsize{7.5}{8}\selectfont},
  pv chip/.style={rounded corners=1.8pt, inner xsep=2.8pt, inner ysep=1.6pt, text=white,
    font=\PVMono\fontseries{b}\fontsize{7}{7}\selectfont},
  pv glass/.style={rounded corners=1.8pt, inner xsep=3pt, inner ysep=1.7pt, text=white,
    fill=PVNavy, fill opacity=0.66, text opacity=1,
    font=\PVSans\fontseries{sb}\fontsize{7}{7}\selectfont},
  pv arrow/.style={-{Stealth[length=5pt,width=4.6pt,round]}, line width=1.25pt, line cap=round},
  pv photo shadow/.style={fill=white, blur shadow={shadow xshift=0pt, shadow yshift=-0.8pt,
    shadow blur radius=2pt, shadow opacity=30}},
}
\ifdefined\PVImgBox\else\newsavebox\PVImgBox\fi %
\newcommand{\PVCoverImage}[3]{%
  \sbox\PVImgBox{\includegraphics{#3}}%
  \ifdim\dimexpr\wd\PVImgBox*\dimexpr#2pt\relax/\ht\PVImgBox\relax<#1pt
    \includegraphics[width=#1pt]{#3}%
  \else
    \includegraphics[height=#2pt]{#3}%
  \fi}
\newcommand{\PVPhoto}[5]{%
  \path[pv photo shadow, rounded corners=3pt] (#1,#2) rectangle ++(#3,#4);
  \begin{scope}
    \clip[rounded corners=3pt] (#1,#2) rectangle ++(#3,#4);
    \node[inner sep=0pt] at ({#1+#3/2},{#2+#4/2}) {\PVCoverImage{#3}{#4}{#5}};
  \end{scope}
  \draw[white, line width=0.7pt, rounded corners=3pt] (#1,#2) rectangle ++(#3,#4);}
\newcommand{\PVFitPhoto}[5]{%
  \sbox\PVImgBox{\includegraphics[width=#3pt,height=#4pt,keepaspectratio]{#5}}%
  \pgfmathsetmacro\PVfw{\wd\PVImgBox}\pgfmathsetmacro\PVfh{\ht\PVImgBox}%
  \pgfmathsetmacro\PVfx{#1+(#3-\PVfw)/2}\pgfmathsetmacro\PVfy{#2+(#4-\PVfh)/2}%
  \path[pv photo shadow, rounded corners=2.5pt] (\PVfx,\PVfy) rectangle ++(\PVfw,\PVfh);
  \begin{scope}\clip[rounded corners=2.5pt] (\PVfx,\PVfy) rectangle ++(\PVfw,\PVfh);
    \node[inner sep=0pt, anchor=south west] at (\PVfx,\PVfy) {\usebox\PVImgBox};
  \end{scope}}
\newcommand{\PVPanelTitle}[4]{%
  \node[pv letter, anchor=south west] at (#1,{#2-2.2}) {#3};
  \node[pv title] at ({#1+14.5},#2) {#4};}
\makeatletter
\def\PV@lv#1#2#3\@nil{PV#1#2}
\newcommand{\PVLevelColor}[1]{\PV@lv#1\@nil}
\makeatother

%% file: sec/2_related_work.tex
\section{Related Work}
\label{sec:related-work}
\subsection{\revision{Embodied Evaluation of General-Purpose Foundation Models}}
Embodied evaluation of general-purpose foundation models~\citep[e.g.,][]{openai2026gpt6,anthropic2026fable} examines whether models can act on environmental observations and task instructions and use execution feedback to complete tasks.
Existing work follows two main directions: broad interactive benchmarks and agents for challenging robotic manipulation.
EmbodiedEval~\citep{cheng2025embodiedeval} and EmbodiedBench~\citep{yang2025embodiedbench} assess visual understanding, spatial reasoning, and sequential decision-making through navigation and manipulation tasks, while RobotWorld extends evaluation across diverse simulated tasks and robot embodiments~\citep{yang2026robotworld}.
Robotic manipulation requires integrating these capabilities with feedback-driven control~\citep{diao2026liberoagent}.
GPT-6 Astra~\citep{openai2026gpt6} demonstrates task understanding and local error correction, but struggles with precision manipulation, dynamic control, and bimanual coordination~\citep{robodojo2026astra,galbot2026astra,robocurve2026stationerybench}.
RoboColiseum compares frontier models through shared robot-control interfaces and likewise finds a gap between task understanding and reliable manipulation~\citep{huang2026robocoliseum}.
Harness refinements and demonstration-based in-context learning can improve execution~\citep{pantograph2026frontierrobots,internrobotics2026ebenchastra}; RoboICL combines demonstration context and interaction memory without parameter updates~\citep{liu2026roboicl}, while EmbodiedSWE studies iterative program development for long-horizon, contact-rich simulation tasks~\citep{shen2026embodiedswe}.
To characterize the remaining gap, we evaluate GPT-6 Astra across a hierarchy of contact demands in physically and visually matched real and simulated environments, examining whether task understanding and local corrections translate into sustained physical progress.

\subsection{\revision{Contact-Rich Manipulation Benchmarks}}
Contact-rich manipulation benchmarks provide an experimental foundation for studying robotic execution under physical constraints, covering individual operations and compositions of multiple operations.
At the level of individual operations, standardized assembly tests and configurable insertion benchmarks vary part geometry, mating clearances, and assembly constraints to assess alignment, insertion, and part mating~\citep{nist2020assembly,delgado2026peginbench}.
Articulated-object benchmarks further examine manipulation under constrained motion and generalization across objects~\citep{bao2023dexart}.
At the level of skill composition, benchmarks connect grasping, reorientation, and insertion into complete manipulation sequences, or use furniture assembly to study skill transitions and execution reliability over longer tasks~\citep{luo2024fmb,heo2023furniturebench}.
These benchmarks support evaluations of individual skills, skill composition, and task completion, but operation categories and task length do not directly determine contact difficulty.
Different tasks may share similar contact requirements, while the same operation may impose different control demands depending on motion constraints, resistance between mating parts, or changes in contact.
To relate these shared requirements across tasks, we organize a hierarchical task suite around contact selection and geometric alignment, force regulation within sustained contact, and coordinated contact reconfiguration and force regulation.
This hierarchy supports analysis of which physical interaction demands GPT-6 Astra~\citep{openai2026gpt6} can handle and how execution difficulties arise as these demands change, linking individual task results to a systematic characterization of its contact-rich manipulation capabilities and limits.

\subsection{\revision{Physical and Visual Matching for Sim--Real Evaluation}}
Using simulation to assess contact capabilities requires both corresponding experimental conditions and validation against real-world execution.
To establish this correspondence, existing studies use visual and control matching to reduce the effects of observation and action-execution differences on policy evaluation~\citep{simpler2024,realm2025}.
Real-scene reconstruction broadens the range of environments available for evaluation and, together with appropriate adaptation methods, improves the correlation between simulated and real-world performance~\citep{polaris2025}.
For contact-rich manipulation, object motion, deformation, and contact responses also determine task outcomes, making physical correspondence an important aspect of evaluation environment construction.
Soft-body interaction evaluations and controlled comparisons of reconstruction pipelines investigate how interaction dynamics, geometry, and physical properties affect agreement in behavior and outcomes between simulation and reality~\citep{softbodysim2025,reconstructioneffects2026}.
To validate conclusions drawn from simulation, related studies compare policy rankings, performance correlations, and failure patterns under perturbations~\citep{wang2026practicalrecipe}, and assess execution quality beyond binary task success~\citep{r2seval2026}.
These efforts establish a foundation for paired Sim--Real evaluation and show that the value of environment matching must be assessed through real-world execution.
We apply this perspective to the contact-rich capabilities of general-purpose foundation models, evaluating GPT-6 Astra on corresponding simulated and real-world tasks with physical and visual matching.
By comparing performance across the three interaction levels and examining contact processes, we test whether the model's strengths and limitations observed in simulation recur in real-world execution and identify discrepancies that depend on specific environmental conditions.

%% file: sec/3_benchmark_design.tex
\section{\revision{Hierarchical Benchmark Design for Contact-Rich Manipulation}}
\label{sec:benchmark-design}

Robotic manipulation involves rich interactions with objects, requiring an understanding of how contact transmits forces and constrains motion. To understand GPT-6 Astra's capabilities in these practical contact-rich settings, we design a hierarchical benchmark that characterizes manipulation performance across increasing contact demands. Specifically, we introduce (i) the principles underlying the three-level contact hierarchy in Section~\ref{sec:contact-hierarchy}, (ii) a suite of 16 household manipulation tasks in Section~\ref{sec:task-suite-overview}, and (iii) task specifications and observable success criteria in Section~\ref{sec:task-success-criteria}.

\subsection{Principles of Hierarchical Task Design}
\label{sec:contact-hierarchy}

\begingroup
\setlength{\parskip}{0.4\baselineskip plus 1pt}
\titleformat{\paragraph}[runin]
  {\normalfont\normalsize\bfseries}{}{0pt}{}
\titlespacing*{\paragraph}{0pt}{0.4\baselineskip}{0.6em}

Contact-rich manipulation uses contact to transmit forces and constrain motion. The intended physical effect depends on where and how contacts are established and on how the transmitted forces or torques affect objects. We organize the hierarchy around these two coupled dimensions:

\Needspace{4\baselineskip}

\begin{itemize}[leftmargin=1.4em,topsep=0.15\baselineskip,itemsep=0.3\baselineskip,parsep=0pt,partopsep=0pt]
\item Contact configuration. This dimension describes where and how the robot contacts task objects, directly or through a tool or container~\citep{mason2001mechanics}. It includes end-effector--object, tool--object, container--contents, and object--support contacts used for grasping, pushing, sliding, or support. The selected contacts must permit the intended operation: a secure grasp is insufficient if it obstructs that operation. Contact reconfiguration is required when the robot must release and re-establish contacts to continue the main operation. Initial grasping, final release, and contact changes arising naturally from sliding, rolling, or the release of contents do not by themselves constitute contact reconfiguration.

\item Force regulation. In this benchmark, force regulation refers to controlling how contact forces or torques change or constrain an object's motion or deformation, beyond ordinary support, transport, and geometric placement. These requirements include overcoming resistance, maintaining pressure, limiting deformation, and regulating an object's speed or direction of motion through contact. They may be met directly or through a tool or container, using motion or gripper commands~\citep{raibert1981hybrid,hogan1985impedance}. Regulation may involve maintaining suitable forces or adjusting them as needed. Direct force measurements or commands are not required; forces during motion alone do not establish a force-regulation requirement.
\end{itemize}

\Needspace{3\baselineskip}
\paragraph{Level assignment.}
These two dimensions define three cumulative levels (Table~\ref{tab:level-criteria}). L1 requires contact selection and geometric alignment, without force regulation beyond ordinary support, transport, and geometric placement. L2 adds force regulation as defined above, without requiring contact reconfiguration to continue the main operation. L3 additionally requires coordinating contact reconfiguration with force regulation. Contact changes used only for positioning, reorientation, or recovery from a failed grasp do not by themselves raise a task above L1. The table illustrates observable failures; Section~\ref{sec:failure-analysis} examines related interaction bottlenecks.

\begin{table}[t]
\centering
\caption{\textbf{Interaction-level criteria.} Requirements accumulate across levels. The last column gives examples of observable failure, which do not uniquely identify the failed requirement. Table~\ref{tab:level-assignment} applies the criteria to all 16 tasks.}
\label{tab:level-criteria}
\small
\setlength{\tabcolsep}{4pt}
\renewcommand{\arraystretch}{1.12}
\begin{tabularx}{\linewidth}{@{}l>{\raggedright\arraybackslash}X>{\raggedright\arraybackslash}X>{\raggedright\arraybackslash}X@{}}
\toprule
\textbf{Level} & \textbf{Force regulation} & \textbf{Contact configuration} & \textbf{Examples of observable failure} \\
\midrule
L1 & No additional regulation beyond ordinary support, transport, and geometric placement & Suitable contacts must be selected; sequential grasps for positioning are allowed & Misalignment; a grasp location that blocks placement or alignment \\
L2 & Required, in the task-level sense defined above & Reconfiguration is not required to continue the main operation & Slip, stalled motion, crushing or damage, displacement, spilling \\
L3 & Required and coordinated with contact reconfiguration & Contacts must be re-established to continue the main operation & Lost torque transmission or support after a contact change; progress that does not accumulate \\
\bottomrule
\end{tabularx}
\end{table}

\Needspace{3\baselineskip}
\paragraph{L1: Contact selection and geometric alignment.}
The five L1 tasks place or align a held object, so success depends on where the object is grasped and how it is aligned. In Hat Hanging, the grasp must hold the hat securely without blocking its opening, allowing the opening to be aligned with the designated rack branch. Charger Alignment requires aligning the prongs with the socket holes without insertion. Cup Nesting uses two sequential grasps for geometric placement, without additional force-regulation requirements.

\Needspace{3\baselineskip}
\paragraph{L2: Force regulation within sustained contact.}
In the seven L2 tasks, force-regulation requirements arise from resistance in Marker Cap Removal, Book Pulling, and Drawer Pull \& Return; from damage limits in Orange Handling and Jelly Handling; from the pressing force needed in Whiteboard Wiping; and, in Marble Pouring, from using cup orientation and motion to regulate the marbles' movement and release into the receiving cup. In Drawer Pull \& Return, for example, one handle grasp is kept while the pulling and pushing forces overcome friction along the guide without moving the cabinet.

\Needspace{3\baselineskip}
\paragraph{L3: Coordinated contact reconfiguration and force regulation.}
In the four L3 tasks, a maintained contact configuration cannot complete the main operation. In Bottle Cap Unscrewing, the rotation needed to disengage the thread exceeds the end effector's turning range within one grasp; regrasping must restore torque transmission for further turning. Nut Unscrewing additionally requires the other arm to hold the unfixed bolt steady. In Stacked-Book Pushing \& Rotation, pushing contacts must change to turn the stack through a full rotation. In Block Sweeping, three separated blocks require successive brush--block contacts. These contact changes must be coordinated with forces that continue the operation while preserving earlier progress.

\Needspace{3\baselineskip}
\paragraph{\normalfont\itshape Scope and interpretation.}
Levels describe a \emph{task version}: its initial configuration, goal, and process constraints (Section~\ref{sec:task-success-criteria}), assessed for the evaluated robot setup. They do not follow from task names, model performance, control interfaces, or contact changes observed in a particular execution. Blocks clustered at a dustpan entrance, for example, can be collected in one sweep and impose only L2 demands (Section~\ref{sec:initial-condition-sensitivity}); a partially unscrewed cap needing only a short turn is likewise L2. External support or an end effector that turns a cap without regrasping can change the requirements relative to our two-finger grippers (Section~\ref{sec:real-world-setup}). The hierarchy describes increasing physical interaction demands without imposing an absolute difficulty ordering, as perception, spatial reasoning, and geometric precision also affect performance. Comparisons across levels characterize performance under combined demands without isolating any factor's causal contribution.

\par
\endgroup

\subsection{Task Suite Overview}
\label{sec:task-suite-overview}

Our benchmark comprises 16 household manipulation tasks: five at L1, seven at L2, and four at L3. Figure~\ref{fig:task-suite}(a) shows simulated and real-world views of the complete suite, expanding the representative overview in Figure~\ref{fig:main-finding}(b). \textcolor{black}{L1 covers placement and alignment in Marker Placement, Charger Alignment, Hat Hanging, Cup Nesting, and Flower Transfer. L2 covers constrained motion and separation in Marker Cap Removal, Drawer Pull \& Return, and Book Pulling, deformable-object manipulation in Orange Handling and Jelly Handling, pouring loose marbles in Marble Pouring, and sustained tool--surface contact in Whiteboard Wiping. L3 includes Bottle Cap Unscrewing, Nut Unscrewing, Stacked-Book Pushing \& Rotation, and Block Sweeping.}

The suite spans rigid objects, deformable food objects, and constrained mechanisms, through direct manipulation and tool or container use. This coverage allows us to examine whether strengths and failures recur across different objects and operations with related interaction demands. \textcolor{black}{Section~\ref{sec:aligned-environments} explains how the corresponding environments and assets are constructed and physically and visually matched.} Figure~\ref{fig:task-suite}(b) shows three initial configurations for each of four representative simulation tasks, illustrating variations in object positions, relative placement, and orientation. Task-specific goals and constraints are defined in Section~\ref{sec:task-success-criteria} and Appendix~\ref{app:task-specifications}, while Section~\ref{sec:evaluation-setup} specifies the model interface and evaluation protocol.

\subsection{Task Specification and Success Criteria}
\label{sec:task-success-criteria}

Each task version is specified by its initial conditions, goal, and process constraints. Initial conditions describe object positions and orientations, any external fixation or support, whether the robot already holds an object, and whether a connection has been partially loosened. The task goal specifies the operation to be completed and its required outcome, such as pulling a drawer outward and returning it to its initial opening. Process constraints define permitted actions and tools, together with requirements such as avoiding damage or maintaining object support during execution. These requirements define the manipulation problem without prescribing a unique action sequence. Complete task specifications are provided in Appendix~\ref{app:task-specifications}.

Task success is assessed using the explicit task-specific criteria in Appendix~\ref{app:task-specifications}, including the required outcomes and process constraints. Completion is assessed through observable outcomes, such as separation, prescribed displacement, or stable support after release. For Jelly Handling, success requires the prescribed lift, return, and release, with no damage throughout execution. Sim--Real comparisons use corresponding success conditions, verified through observable criteria in each environment.

For diagnostic analysis, we define \emph{effective contact} as contact that transmits the force or torque needed for the main operation, and distinguish it from complete success. In Bottle Cap Unscrewing, local cap rotation relative to the fixed bottle indicates that effective contact has been established, whereas complete success requires full thread disengagement and controlled lifting of the cap to the specified separation. Effective contact may be lost and re-established during execution. It helps interpret local physical progress but does not replace the task-level success criterion.

%% file: sec/4_aligned_environments.tex
\section{Physically and Visually Matched Sim--Real Environments}
\label{sec:aligned-environments}

Our contact-rich manipulation tasks span the matched simulated and real-world environments illustrated in Figure~\ref{fig:main-finding}(a), enabling us to assess Astra's capabilities and cross-environment differences across the three interaction levels. To establish comparable environments, we match four key components: workspace layout and task initialization; task assets and contact responses; robot embodiment and configuration; and camera configuration and visual observations. Specifically, we introduce (i) the real-world setup in Section~\ref{sec:real-world-setup}, (ii) the simulated environment and task assets in Section~\ref{sec:simulation-reconstruction}, and (iii) the physical and visual matching procedures and the resulting correspondence in Section~\ref{sec:physical-visual-alignment}.

\subsection{Real-World Setup}
\label{sec:real-world-setup}

As shown in Figure~\ref{fig:real-world-setup}(a), the real-world platform comprises two ARX R5 robotic arms, each equipped with an R5-series two-finger gripper. Both arms are secured at the table edge and share a tabletop workspace. This configuration supports single-arm manipulation and dual-arm coordination, including tasks in which one arm stabilizes an object while the other manipulates it. Task objects and supporting fixtures are placed within this workspace; their initial configurations and task-specific constraints are defined in Appendix~\ref{app:task-specifications}.

The visual sensing setup consists of one external Orbbec Gemini Pro camera on a separate stand and two wrist-mounted Orbbec Gemini 305 cameras, one on each arm. The external camera provides an overall view of the workspace and object layout, while the wrist cameras provide local views of the grippers and nearby objects. Figure~\ref{fig:real-world-setup}(b) and (c) show how each wrist camera is mounted relative to its gripper on the left and right arms. These three views provide complementary observations of the scene and local manipulation. The arms and grippers have no dedicated contact-force sensors, so no direct contact-force measurements are supplied to the model. Hardware and sensing details are provided in Appendix~\ref{app:real-world-setup}.

\input{figures/figure03_real_world_setup}

\Needspace{6\baselineskip}
\subsection{\revision{Simulation Environment and Task Assets}}
\label{sec:simulation-reconstruction}
\input{figures/figure04_simulation_assets}

We build the simulated environments on NVIDIA Isaac Sim\footnote{\url{https://github.com/isaac-sim/IsaacSim}}. The simulated setup includes the open-source ARX R5 model in a dual-arm configuration, together with external and wrist-mounted cameras. Workspace reconstruction begins with multiple video captures of the real scene. Automated segmentation identifies individual scene components, while measurements of key dimensions establish their scale. Reconstructed geometry and textures represent task-contact elements, such as the tabletop and supports, as well as background elements, such as walls, that primarily affect visual observations. Contact surfaces additionally receive collision geometry and friction settings.

\textcolor{black}{Task assets use their physical counterparts as references for appearance, task-relevant geometry, and interaction features. Figure~\ref{fig:simulation-assets}(a) presents the object, tool, and support assets used across the benchmark tasks, including irregular shapes with distinct contact geometries and access constraints. Figure~\ref{fig:simulation-assets}(b) illustrates relative motion and separation between components, deformation under contact, inter-object sliding, and changes in visible surface state during wiping. These features can coexist within an asset and combine to support the contact requirements of tasks across L1--L3. Each object type is represented by a single selected variant, with repeated copies where required by the task.}

\textcolor{black}{These features are implemented through contact geometry, component constraints, and physical properties. Rigid-body assets retain collision geometry and task-specific constraints, including drawer sliding joints. Mass and friction affect object motion and slip, while finite-element representations of orange and jelly use elastic parameters to model deformation under contact. Wiping updates the visible surface state from contact and sliding.} Asset representations, implementation settings, and parameter sources are provided in Appendix~\ref{app:simulation-assets}.
\subsection{Physical and Visual Matching}
\label{sec:physical-visual-alignment}
\label{sec:alignment-evidence} %
\input{figures/figure03_alignment_pipeline}

\textcolor{black}{Physical and visual matching aims to preserve the interaction conditions and visual observations that govern task execution across the real and simulated environments. Figure~\ref{fig:alignment-pipeline} illustrates the resulting correspondence.}

\noindent\textit{Workspace Layout and Task Initialization.}
Workspace geometry and task initialization determine the contacts available at the start of a trial and the remaining motion required to complete the task. Measured table dimensions and spatial layout establish the workspace geometry. Task-specific initial configurations follow Appendix~\ref{app:task-specifications}, covering object positions and orientations, support and fixation conditions, and initial mechanical engagement, including the screw--nut engagement that determines the rotation remaining before separation. Figure~\ref{fig:alignment-pipeline}(a) illustrates the workspace layout and initial object placement for Block Sweeping and, on the right, compares the initial screw--nut engagement in the real and simulated setups.

\noindent\textit{Task Assets and Contact Responses.}
We reconstruct asset geometry and appearance from multi-view images of real objects and use measured dimensions to set their physical scale, matching object shape, color, and surface texture across environments (Figure~\ref{fig:alignment-pipeline}(b1)).

Physical parameters are set through direct measurement and refinement against observed interaction responses. Object masses are set from weighing measurements, as illustrated by the jelly example in Figure~\ref{fig:alignment-pipeline}(b2). Contact friction parameters are adjusted based on relative sliding observed during manipulation. For deformable assets, material parameters are refined by comparing simulated deformation with that observed in the physical objects. The refinement targets contact responses relevant to the evaluated tasks. Figure~\ref{fig:alignment-pipeline}(b3) shows recorded jelly states before and during gripper contact. A shared global color adjustment is applied to simulated views; these qualitative frames show observed states without quantifying agreement in deformation or contact force.

\noindent\textit{Robot Embodiment and Configuration.}
We use the open-source ARX R5 model\footnote{Official ARX R5 resources: \url{https://github.com/ARXroboticsX/R5}.} to match the geometry and kinematic structure of the physical robot, including its joint layout and degrees of freedom. We further configure the mounting positions and orientations of both arms, the two-finger grippers, and their spatial relationship to the tabletop to match the real-world setup. Figure~\ref{fig:alignment-pipeline}(c) illustrates the resulting correspondence in arm placement and gripper geometry.

\noindent\textit{Camera Configuration and Visual Observations.}
We match camera projection parameters and mounting configurations to reproduce the observation geometry of the real-world setup. This includes the external camera pose and the mounting poses of both wrist cameras relative to the end effectors. Reference images from real-scene videos guide parameter estimation and refinement. We compare real and rendered observations at corresponding viewpoints and comparable robot and object states, examining image scale, spatial relationships, and local visibility. Lighting and background appearance are further adjusted against the real images. Figure~\ref{fig:alignment-pipeline}(d) presents paired external, left-wrist, and right-wrist views, illustrating correspondence in both workspace context and local observations around the grippers. For display in Figure~\ref{fig:main-finding}(a), images are cropped for framing and the real-world image receives a global color adjustment.

\FloatBarrier

%% file: figures/figure03_real_world_setup.tex
\begingroup
\input{figures/comparison_figure_style}
\tikzset{
  ftlabel/.style={fill=white, fill opacity=0.95, text opacity=1, draw=PVBlue!50, line width=0.55pt,
    rounded corners=2pt, inner xsep=3pt, inner ysep=1.9pt, anchor=south west,
    font=\PVSans\fontseries{sb}\fontsize{7.5}{9}\selectfont, text=PVInk},
  ftmodel/.style={ftlabel, font=\PVMono\fontseries{b}\fontsize{7.5}{9}\selectfont, text=PVBlue},
  fttitle/.style={pv title, font=\PVMono\fontseries{b}\fontsize{10}{12}\selectfont},
}
\newcommand{\FigThreeTitle}[4]{%
  \node[pv letter, anchor=south west, yshift=-2.2pt] at (#1,#2) {#3};
  \node[fttitle, xshift=14.5pt] at (#1,#2) {#4};}
\newcommand{\FigThreeShadow}[2]{\path[pv photo shadow, rounded corners=3pt] #1 rectangle #2;}
\newcommand{\FigThreeRim}[2]{\draw[white, line width=0.7pt, rounded corners=3pt] #1 rectangle #2;}
\newcommand{\FigThreeLeader}[2]{%
  \draw[white, line width=1.6pt, line cap=round] #1 -- #2;
  \draw[PVBlue, line width=0.65pt, line cap=round] #1 -- #2;
  \fill[PVBlue, draw=white, line width=0.5pt] #2 circle (1.25pt);}
\begin{figure*}[!t]
\centering
\begin{tikzpicture}[x=0.2611111pt, y=0.2611111pt]
\FigThreeTitle{10}{510}{a}{Dual-Arm Tabletop Platform}
\FigThreeTitle{866.4}{510}{b}{Left Wrist Camera}
\FigThreeTitle{1340.4}{510}{c}{Right Wrist Camera}
\FigThreeShadow{(10,23.6)}{(842.4,491)}
\FigThreeShadow{(866.4,23.6)}{(1316.4,491)}
\FigThreeShadow{(1340.4,23.6)}{(1790.4,491)}
\begin{scope}\clip[rounded corners=3pt] (10,23.6) rectangle (842.4,491);
  \node[anchor=south west, inner sep=0pt] at (10,-133.3) {\includegraphics[width=217.35pt,height=163.01pt]{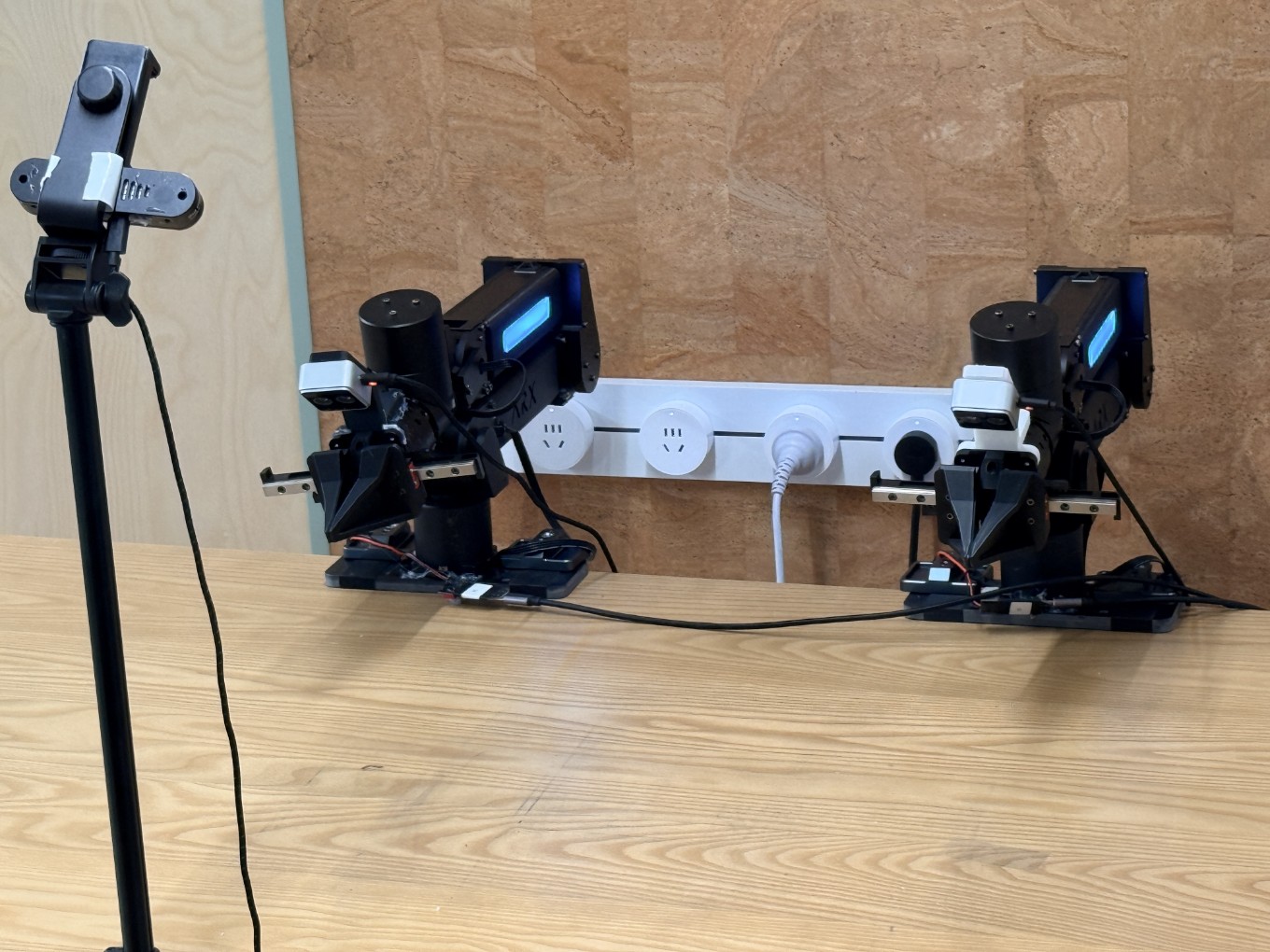}};\end{scope}
\begin{scope}\clip[rounded corners=3pt] (866.4,23.6) rectangle (1316.4,491);
  \node[anchor=south west, inner sep=0pt] at (847.805,-72.16411) {\includegraphics[width=133.18pt,height=177.57pt]{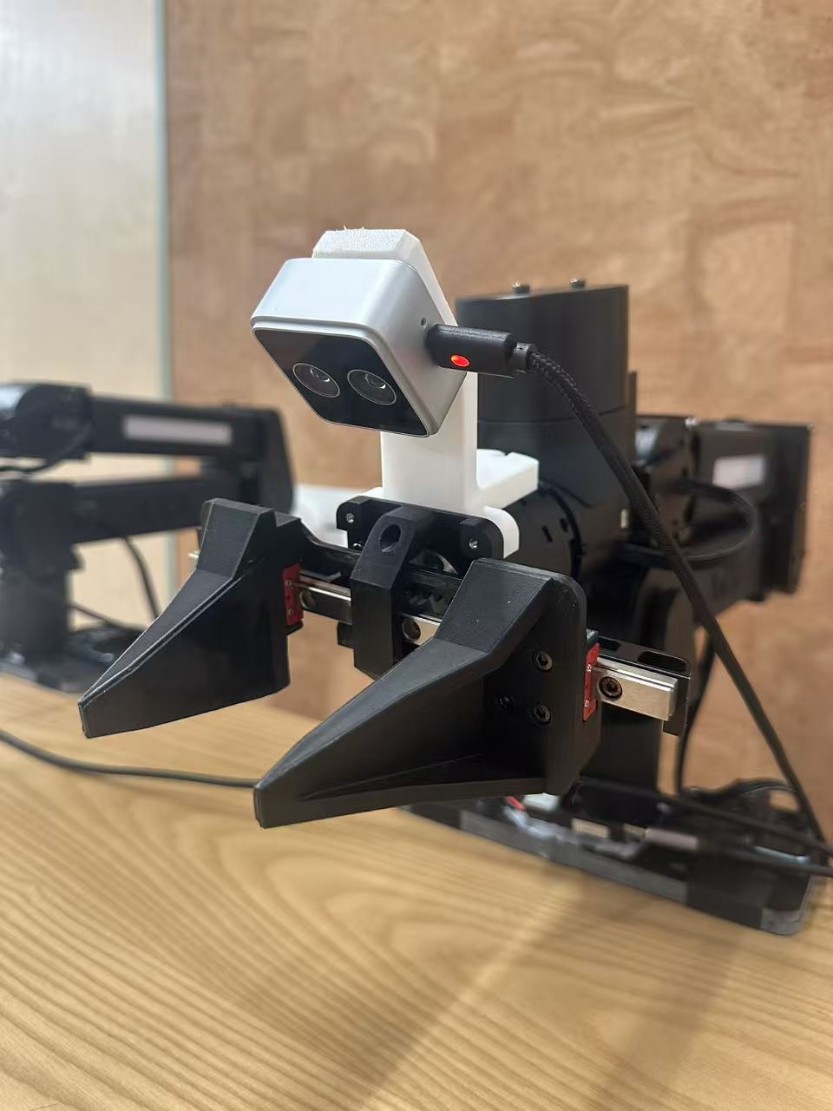}};\end{scope}
\begin{scope}\clip[rounded corners=3pt] (1340.4,23.6) rectangle (1790.4,491);
  \node[anchor=south west, inner sep=0pt] at (1290.307,-88.74026) {\includegraphics[width=131.84pt,height=175.79pt]{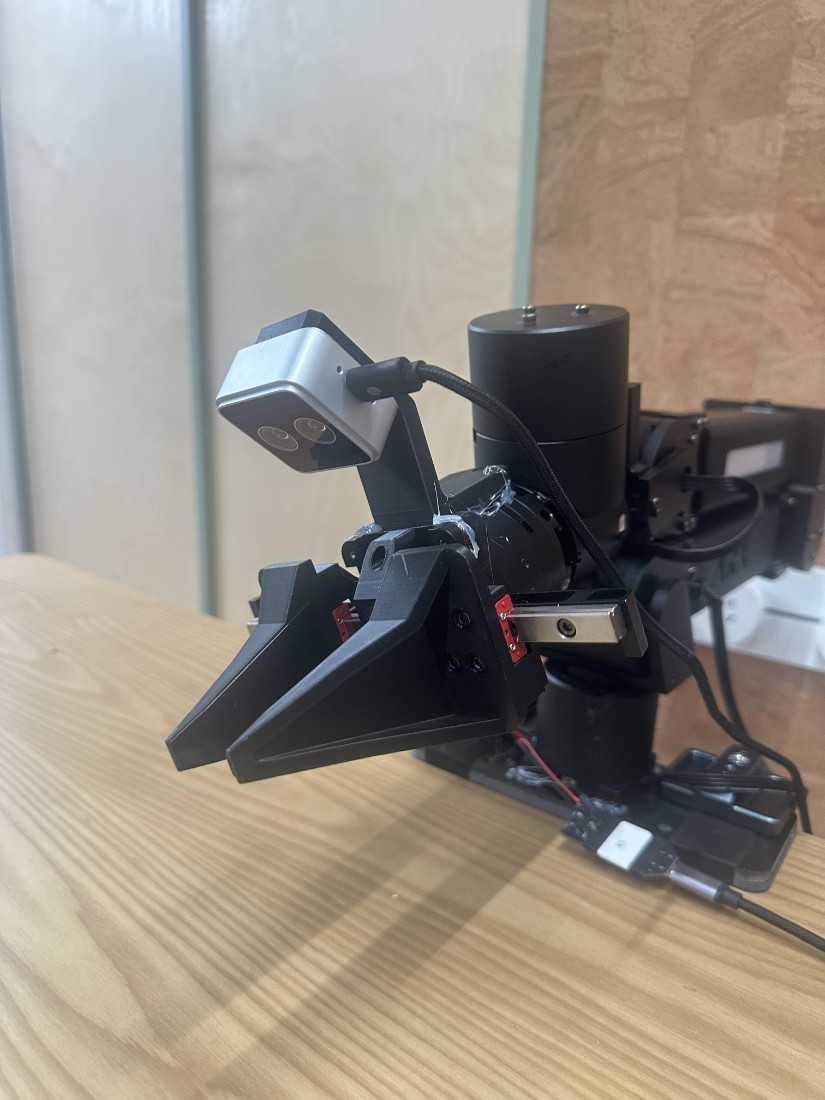}};\end{scope}
\FigThreeRim{(10,23.6)}{(842.4,491)}
\FigThreeRim{(866.4,23.6)}{(1316.4,491)}
\FigThreeRim{(1340.4,23.6)}{(1790.4,491)}
\FigThreeLeader{(182,433)}{(78.128,356.809)}
\node[ftlabel] at (182,415) {External camera mount};
\FigThreeLeader{(334,330)}{(228.835,239.133)}
\FigThreeLeader{(443,330)}{(664.442,230.875)}
\node[ftlabel] at (276,330) {Two wrist cameras};
\FigThreeLeader{(477,103)}{(311.415,125.586)}
\FigThreeLeader{(580,103)}{(678.893,113.199)}
\node[ftmodel] at (420,67) {ARX R5 dual arms};
\draw[fill=white, fill opacity=0.95, draw=PVBlue, line width=0.7pt, dash pattern=on 2.4pt off 1.6pt,
  rounded corners=3pt] (556,298) rectangle (836,482);
\node[inner sep=0pt] at (696,428) {\includegraphics[width=40pt]{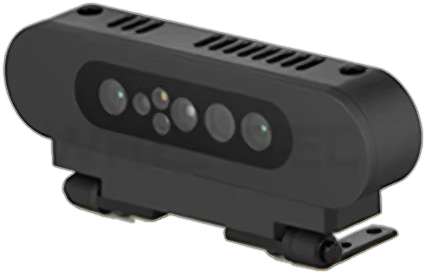}};
\node[inner sep=0pt, font=\PVMono\fontseries{b}\fontsize{7}{8.4}\selectfont, text=PVBlue] at (696,357) {Orbbec Gemini Pro};
\node[inner sep=0pt, font=\PVSans\fontsize{7}{8.4}\selectfont, text=PVMuted] at (696,323) {External camera};
\FigThreeLeader{(1041,441)}{(1070.414,392.712)}
\node[ftmodel] at (882,441) {Orbbec Gemini 305};
\FigThreeLeader{(1161,74)}{(1177.203,193.479)}
\node[ftlabel] at (1004,38) {R5 two-finger gripper};
\FigThreeLeader{(1470,449)}{(1464.296,324.024)}
\node[ftmodel] at (1352,449) {Orbbec Gemini 305};
\FigThreeLeader{(1480,83)}{(1547.784,183.931)}
\node[ftlabel] at (1352,47) {R5 two-finger gripper};
\end{tikzpicture}
\caption{\textbf{Real-world experimental setup.} (a) Two ARX R5 arms with two-finger grippers share a tabletop workspace observed by one external Orbbec Gemini Pro camera and two wrist-mounted Orbbec Gemini 305 cameras. The external-camera mount is visible at left; the inset identifies the camera used. (b) Left and (c) right wrist cameras with their two-finger grippers. Photographs show the hardware before task objects are placed.}
\label{fig:real-world-setup}
\end{figure*}
\arrayrulecolor{black}
\endgroup

%% file: figures/figure04_simulation_assets.tex
\begingroup
\input{figures/comparison_figure_style}
\tikzset{
  ff name/.style={inner sep=0pt, anchor=base, text=PVInk, font=\PVSans\fontseries{sb}\fontsize{7}{8.4}\selectfont},
  ff state/.style={inner sep=0pt, anchor=base, text=PVInk, font=\PVSans\fontsize{7.2}{8.6}\selectfont},
  ff group/.style={draw=PVBlue!45, line width=0.7pt, dash pattern=on 3pt off 2pt, rounded corners=5pt},
  ff head/.style={inner sep=0pt, anchor=base west, text=PVBlue, font=\PVSans\fontseries{sb}\fontsize{7.8}{9}\selectfont},
  ff arrow/.style={-{Stealth[length=4.4pt,width=4pt,round]}, line width=1.1pt, line cap=round, draw=PVBlue},
}
\newcommand{\FfShadow}[3]{\fill[PVInk, fill opacity=0.22, path fading=circle with fuzzy edge 20 percent]
  (#1,#2) ellipse ({#3} and 9);}
\newcommand{\FfBadge}[2]{\node[circle, fill=white, draw=PVBlue!35, line width=0.5pt, inner sep=0pt,
  minimum size=10pt, text=PVBlue, font=\fontsize{6.5}{6.5}\selectfont] at (#1,#2) {\faArrowRight};}
\begin{figure*}[!t]
\centering
\begin{tikzpicture}[x=0.29375pt, y=0.29375pt]
\path[use as bounding box] (0,0) rectangle (1600,1400);
\node[pv letter, anchor=south west, yshift=-2.2pt] at (8,1366) {a};
\node[pv title, xshift=14.5pt] at (8,1366) {Simulation Task Assets};
\node[pv muted, anchor=base east] at (1592,1366) {21 representative asset types\enspace\textperiodcentered\enspace views not to scale};
\FfShadow{128}{1232.21}{83.2}
\FfShadow{352}{1191}{54.81}
\FfShadow{576}{1197.96}{83.2}
\FfShadow{800}{1191}{44.29}
\FfShadow{1024}{1192.45}{83.2}
\FfShadow{1248}{1193.58}{83.2}
\FfShadow{1472}{1191}{33.09}
\FfShadow{128}{981}{50.09}
\FfShadow{352}{998.22}{83.2}
\FfShadow{576}{981}{30}
\FfShadow{800}{981}{81.13}
\FfShadow{1024}{981}{76.8}
\FfShadow{1247.99}{981}{59.91}
\FfShadow{1472}{983}{83.2}
\FfShadow{128}{771}{30}
\FfShadow{352}{832.37}{83.2}
\FfShadow{576}{801.83}{83.2}
\FfShadow{800}{771}{65.68}
\FfShadow{1024}{771}{47.52}
\FfShadow{1248}{771}{74.82}
\FfShadow{1472}{771}{50.15}
\node[anchor=south west, inner sep=0pt] at (24,1229.21) {\includegraphics[width=61.1pt,height=22.79pt]{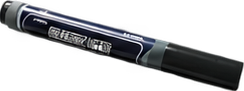}};
\node[anchor=south west, inner sep=0pt] at (283.49,1188) {\includegraphics[width=40.25pt,height=47pt]{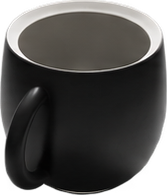}};
\node[anchor=south west, inner sep=0pt] at (472,1194.96) {\includegraphics[width=61.1pt,height=42.91pt]{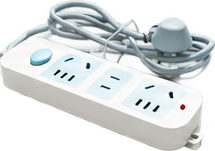}};
\node[anchor=south west, inner sep=0pt] at (744.64,1188) {\includegraphics[width=32.52pt,height=47pt]{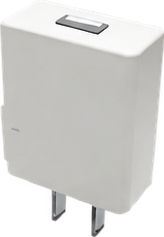}};
\node[anchor=south west, inner sep=0pt] at (920,1189.45) {\includegraphics[width=61.1pt,height=46.15pt]{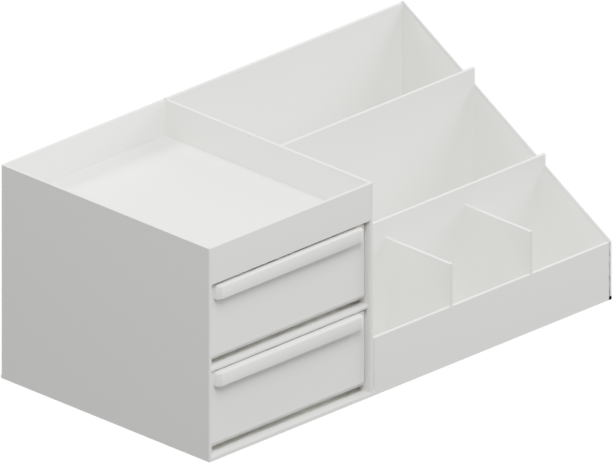}};
\node[anchor=south west, inner sep=0pt] at (1144,1190.58) {\includegraphics[width=61.1pt,height=45.48pt]{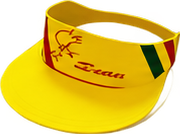}};
\node[anchor=south west, inner sep=0pt] at (1430.63,1188) {\includegraphics[width=24.3pt,height=47pt]{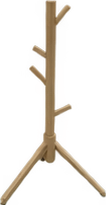}};
\node[anchor=south west, inner sep=0pt] at (65.39,978) {\includegraphics[width=36.78pt,height=47pt]{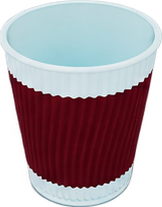}};
\node[anchor=south west, inner sep=0pt] at (248,995.22) {\includegraphics[width=61.1pt,height=36.88pt]{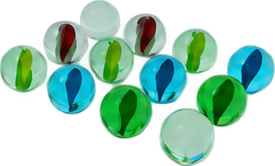}};
\node[anchor=south west, inner sep=0pt] at (553.84,978) {\includegraphics[width=13.02pt,height=47pt]{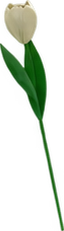}};
\node[anchor=south west, inner sep=0pt] at (698.59,978) {\includegraphics[width=59.58pt,height=47pt]{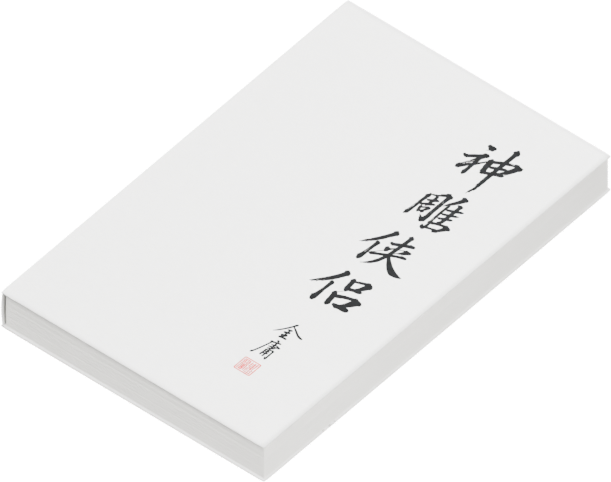}};
\node[anchor=south west, inner sep=0pt] at (928,978) {\includegraphics[width=56.4pt,height=47pt]{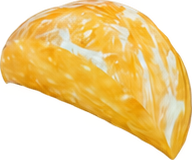}};
\node[anchor=south west, inner sep=0pt] at (1173.11,978) {\includegraphics[width=43.99pt,height=47pt]{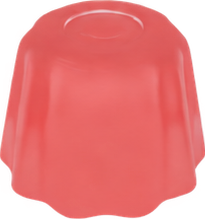}};
\node[anchor=south west, inner sep=0pt] at (1368,980) {\includegraphics[width=61.1pt,height=45.83pt]{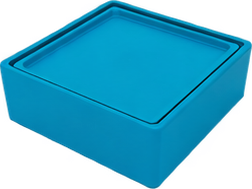}};
\node[anchor=south west, inner sep=0pt] at (99.18,768) {\includegraphics[width=16.93pt,height=47pt]{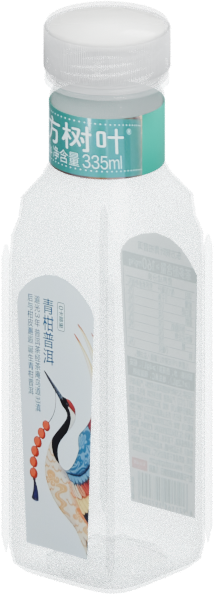}};
\node[anchor=south west, inner sep=0pt] at (248,829.37) {\includegraphics[width=61.1pt,height=10.95pt]{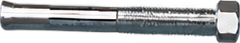}};
\node[anchor=south west, inner sep=0pt] at (472,798.83) {\includegraphics[width=61.1pt,height=28.89pt]{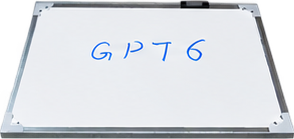}};
\node[anchor=south west, inner sep=0pt] at (717.91,768) {\includegraphics[width=48.23pt,height=47pt]{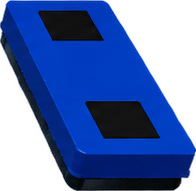}};
\node[anchor=south west, inner sep=0pt] at (964.6,768) {\includegraphics[width=34.9pt,height=47pt]{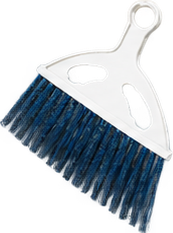}};
\node[anchor=south west, inner sep=0pt] at (1154.47,768) {\includegraphics[width=54.95pt,height=47pt]{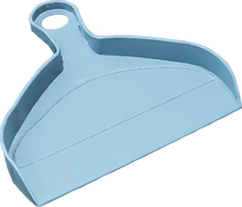}};
\node[anchor=south west, inner sep=0pt] at (1409.31,768) {\includegraphics[width=36.83pt,height=47pt]{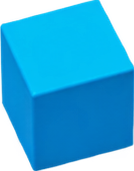}};
\node[ff name] at (128,1160) {Marker \& cap};
\node[ff name] at (352,1160) {Ceramic mug};
\node[ff name] at (576,1160) {Power strip};
\node[ff name] at (800,1160) {Charger};
\node[ff name] at (1024,1160) {Drawer organizer};
\node[ff name] at (1248,1160) {Hat};
\node[ff name] at (1472,1160) {Hat rack};
\node[ff name] at (128,950) {Paper cup};
\node[ff name] at (352,950) {Marbles};
\node[ff name] at (576,950) {Artificial flower};
\node[ff name] at (800,950) {Book};
\node[ff name] at (1024,950) {Orange segment};
\node[ff name] at (1248,950) {Jelly};
\node[ff name] at (1472,950) {Support box};
\node[ff name] at (128,740) {Bottle \& cap};
\node[ff name] at (352,740) {Bolt \& nut};
\node[ff name] at (576,740) {Whiteboard};
\node[ff name] at (800,740) {Board eraser};
\node[ff name] at (1024,740) {Hand brush};
\node[ff name] at (1248,740) {Dustpan};
\node[ff name] at (1472,740) {Block};
\node[pv letter, anchor=south west, yshift=-2.2pt] at (8,688) {b};
\node[pv title, xshift=14.5pt] at (8,688) {Interaction-Relevant Features};
\path[ff group] (8,342) rectangle (1592,668);
\path[ff group] (8,4) rectangle (700,328);
\path[ff group] (716,4) rectangle (1150,328);
\path[ff group] (1166,4) rectangle (1592,328);
\node[ff head] at (26,637) {\faArrowsAltH\,\,Relative motion and separation};
\node[ff head] at (26,300) {\faIcon{compress-arrows-alt}\,\,Deformation under contact};
\node[ff head] at (734,300) {\faLayerGroup\,\,Inter-object sliding};
\node[ff head] at (1184,300) {\faBroom\,\,Surface wiping};
\node[anchor=south west, inner sep=0pt] at (85.69,391) {\includegraphics[width=8.11pt,height=66.09pt]{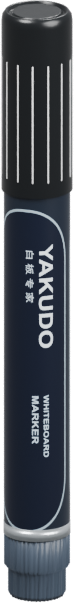}};
\node[anchor=south west, inner sep=0pt] at (278.49,391) {\includegraphics[width=5.88pt,height=66.09pt]{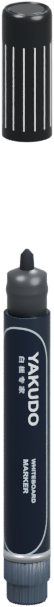}};
\path[pv photo shadow, rounded corners=2.5pt] (403.87,391) rectangle ++(432.26,225);
\begin{scope}\clip[rounded corners=2.5pt] (403.87,391) rectangle ++(432.26,225);
  \node[anchor=south west, inner sep=0pt] at (403.87,391) {\includegraphics[width=126.98pt,height=66.09pt]{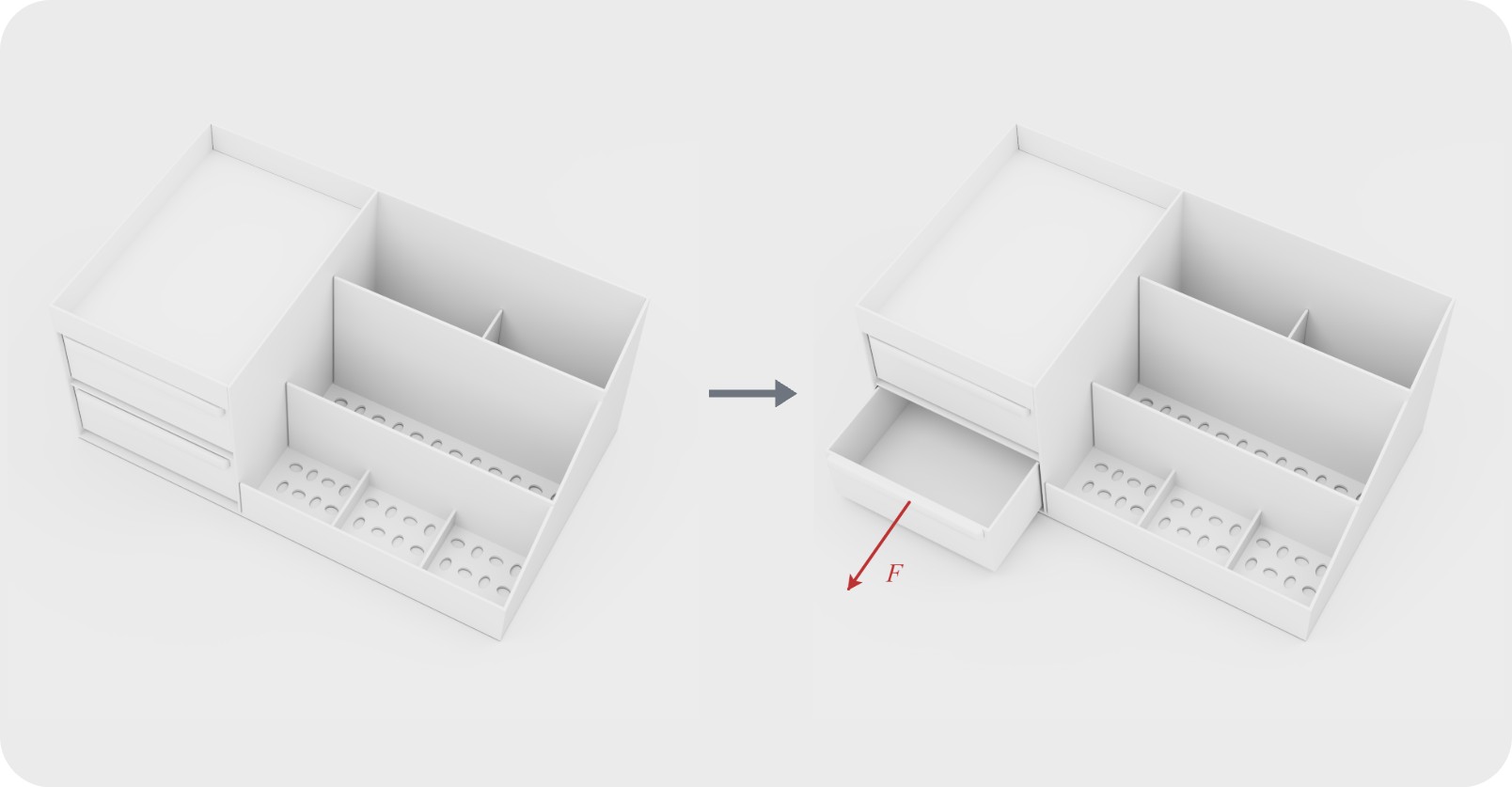}};\end{scope}
\path[pv photo shadow, rounded corners=2.5pt] (893.77,391) rectangle ++(316.47,225);
\begin{scope}\clip[rounded corners=2.5pt] (893.77,391) rectangle ++(316.47,225);
  \node[anchor=south west, inner sep=0pt] at (893.77,391) {\includegraphics[width=92.96pt,height=66.09pt]{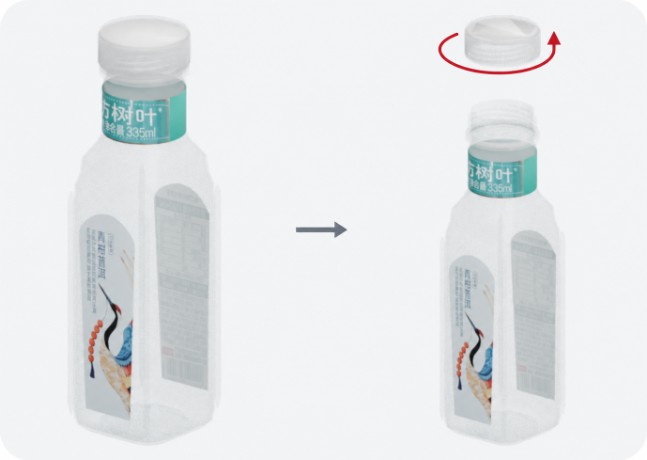}};\end{scope}
\node[anchor=south west, inner sep=0pt] at (1294.92,391) {\includegraphics[width=13.26pt,height=66.09pt]{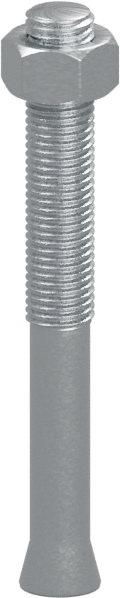}};
\node[anchor=south west, inner sep=0pt] at (1484.16,391) {\includegraphics[width=10.78pt,height=66.09pt]{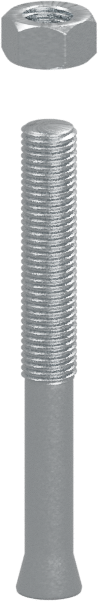}};
\path[pv photo shadow, rounded corners=2.5pt] (24,66.11) rectangle ++(206.67,201.77);
\begin{scope}\clip[rounded corners=2.5pt] (24,66.11) rectangle ++(206.67,201.77);
  \node[anchor=south west, inner sep=0pt] at (24,66.11) {\includegraphics[width=60.71pt,height=59.27pt]{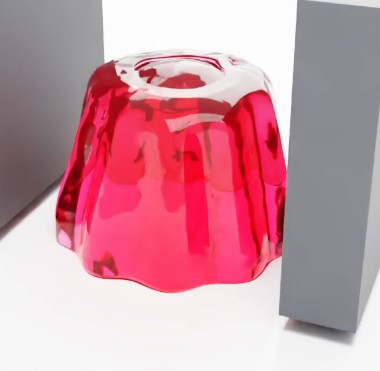}};\end{scope}
\path[pv photo shadow, rounded corners=2.5pt] (246.67,66.11) rectangle ++(206.67,201.77);
\begin{scope}\clip[rounded corners=2.5pt] (246.67,66.11) rectangle ++(206.67,201.77);
  \node[anchor=south west, inner sep=0pt] at (246.67,66.11) {\includegraphics[width=60.71pt,height=59.27pt]{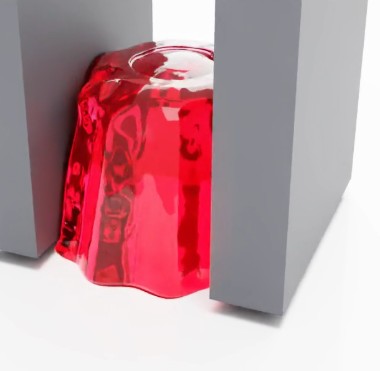}};\end{scope}
\path[pv photo shadow, rounded corners=2.5pt] (469.33,66.11) rectangle ++(206.67,201.77);
\begin{scope}\clip[rounded corners=2.5pt] (469.33,66.11) rectangle ++(206.67,201.77);
  \node[anchor=south west, inner sep=0pt] at (469.33,66.11) {\includegraphics[width=60.71pt,height=59.27pt]{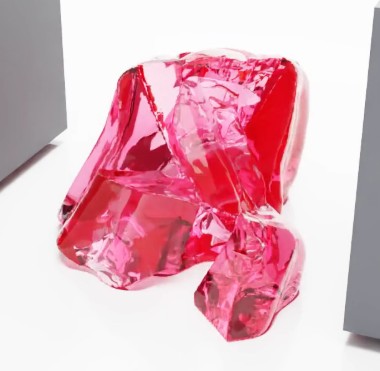}};\end{scope}
\node[anchor=south west, inner sep=0pt] at (740,107.96) {\includegraphics[width=54.34pt,height=34.69pt]{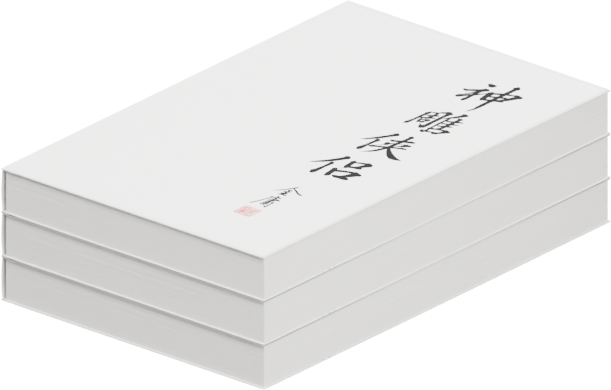}};
\node[anchor=south west, inner sep=0pt] at (941,115.83) {\includegraphics[width=54.34pt,height=30.06pt]{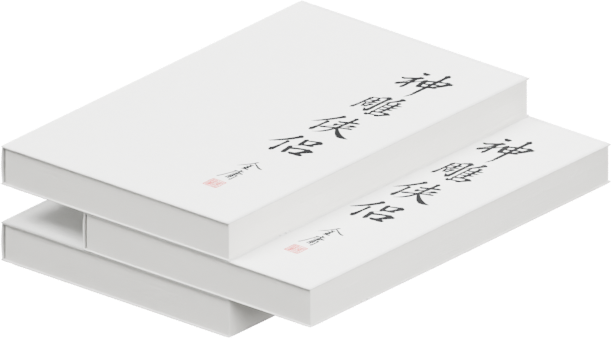}};
\path[pv photo shadow, rounded corners=2.5pt] (1190,85.65) rectangle ++(185,162.69);
\begin{scope}\clip[rounded corners=2.5pt] (1190,85.65) rectangle ++(185,162.69);
  \node[anchor=south west, inner sep=0pt] at (1190,85.65) {\includegraphics[width=54.34pt,height=47.79pt]{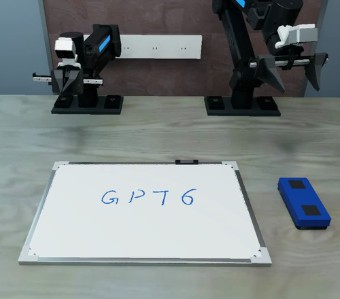}};\end{scope}
\path[pv photo shadow, rounded corners=2.5pt] (1391,85.65) rectangle ++(185,162.69);
\begin{scope}\clip[rounded corners=2.5pt] (1391,85.65) rectangle ++(185,162.69);
  \node[anchor=south west, inner sep=0pt] at (1391,85.65) {\includegraphics[width=54.34pt,height=47.79pt]{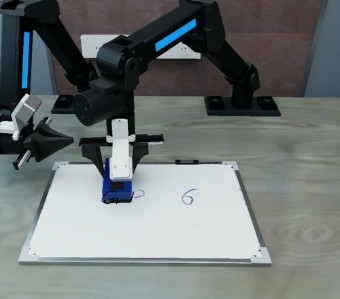}};\end{scope}
\draw[ff arrow] (160,503.5) -- (238,503.5);
\draw[ff arrow] (1378,503.5) -- (1456,503.5);
\FfBadge{238.67}{167}
\FfBadge{461.33}{167}
\FfBadge{933}{167}
\FfBadge{1383}{167}
\node[ff state] at (194,363) {Marker \& cap};
\node[ff state] at (620,363) {Drawer};
\node[ff state] at (1052,363) {Bottle \& cap};
\node[ff state] at (1409.5,363) {Bolt \& nut};
\node[ff state] at (127.3,24) {Initial};
\node[ff state] at (350,24) {Compressed};
\node[ff state] at (572.7,24) {After compression};
\node[ff state] at (832.5,24) {Stacked};
\node[ff state] at (1033.5,24) {Displaced};
\node[ff state] at (1282.5,24) {Before wiping};
\node[ff state] at (1483.5,24) {After wiping};
\end{tikzpicture}
\caption{\textbf{Simulation task assets and interaction-relevant features.}
(a) Illustrative views of 21 representative task asset types.
(b) Representative examples illustrate relative motion and separation between components, deformation under contact, inter-object sliding, and changes in visible surface state during wiping.
Jelly deformation and wiping are shown using recorded simulation frames; component and book arrangements illustrate representative configurations.
Asset views are not to scale. Simulation settings and contact and material models are provided in Appendix~\ref{app:simulation-assets}.}
\label{fig:simulation-assets}
\end{figure*}
\arrayrulecolor{black}
\endgroup

%% file: figures/figure03_alignment_pipeline.tex
\begingroup
\input{figures/comparison_figure_style}
\tikzset{
  fv title/.style={pv title, font=\PVMono\fontseries{b}\fontsize{10}{12}\selectfont},
  fv sub/.style={inner sep=0pt, anchor=base, text=PVInk, font=\PVSans\fontseries{sb}\fontsize{7.5}{9}\selectfont},
  fv legend/.style={inner sep=0pt, anchor=base west, text=PVInk, font=\PVSans\fontsize{7.5}{9}\selectfont},
  fv arrow/.style={-{Stealth[length=4pt,width=3.8pt,round]}, line width=1pt, line cap=round},
  fv pill/.style={rounded corners=3.6pt, inner xsep=3pt, inner ysep=1.7pt, anchor=center,
    font=\PVMono\fontseries{b}\fontsize{7}{7}\selectfont},
  fv divider/.style={draw=PVLine, line width=0.5pt},
  fv rowchip/.style={pv chip, rotate=90, anchor=center},
}
\newcommand{\FvPhoto}[5]{%
  \path[pv photo shadow, rounded corners=2.5pt] (#1,#2) rectangle ++(#3,#4);
  \begin{scope}\clip[rounded corners=2.5pt] (#1,#2) rectangle ++(#3,#4);
    \node[anchor=south west, inner sep=0pt] at (#1,#2) {\includegraphics[width=#3pt,height=#4pt]{assets/figure05_vivid/#5.jpg}};
  \end{scope}}
\newcommand{\FvBar}[4]{\fill[#4, rounded corners=0.8pt] (#1,#2-1.2) rectangle ++(#3,-1.6);}
\begin{figure*}[!t]
\centering
\begin{tikzpicture}[x=1pt,y=1pt]
\path[use as bounding box] (0,0) rectangle (470,337);
\node[pv letter, anchor=south west, yshift=-2.2pt] at (0,329) {a};
\node[fv title, xshift=14.5pt] at (0,329) {Workspace and Initial State};
\node[pv letter, anchor=south west, yshift=-2.2pt, minimum width=13pt] at (0,240.2) {b1};
\node[fv title, xshift=17pt] at (0,240.2) {Geometry and Texture};
\node[pv letter, anchor=south west, yshift=-2.2pt, minimum width=13pt] at (232.57,240.2) {b2};
\node[fv title, xshift=17pt] at (232.57,240.2) {Jelly Mass};
\node[pv letter, anchor=south west, yshift=-2.2pt, minimum width=13pt] at (352.65,240.2) {b3};
\node[fv title, xshift=17pt] at (352.65,240.2) {Jelly Contact};
\node[pv letter, anchor=south west, yshift=-2.2pt] at (0,135.2) {c};
\node[fv title, xshift=14.5pt] at (0,135.2) {Robot Embodiment};
\node[pv letter, anchor=south west, yshift=-2.2pt] at (211.77,135.2) {d};
\node[fv title, xshift=14.5pt] at (211.77,135.2) {Camera Observations};
\node[fv legend, anchor=base east] (lgR) at (470,329) {Real world};
\fill[PVReal, rounded corners=0.8pt] ([xshift=-15pt,yshift=2.1pt]lgR.base west) rectangle ++(12,1.8);
\node[fv legend, anchor=base east] (lgS) at ([xshift=-24pt]lgR.base west) {Simulation};
\fill[PVSim, rounded corners=0.8pt] ([xshift=-15pt,yshift=2.1pt]lgS.base west) rectangle ++(12,1.8);
\draw[fv divider] (276.82,322.5) -- (276.82,253.2);
\draw[fv divider] (223.77,247.2) -- (223.77,148.2);
\draw[fv divider] (343.84,247.2) -- (343.84,148.2);
\draw[fv divider] (201.88,142.2) -- (201.88,3.9);
\FvPhoto{0.16}{256}{125.62}{55}{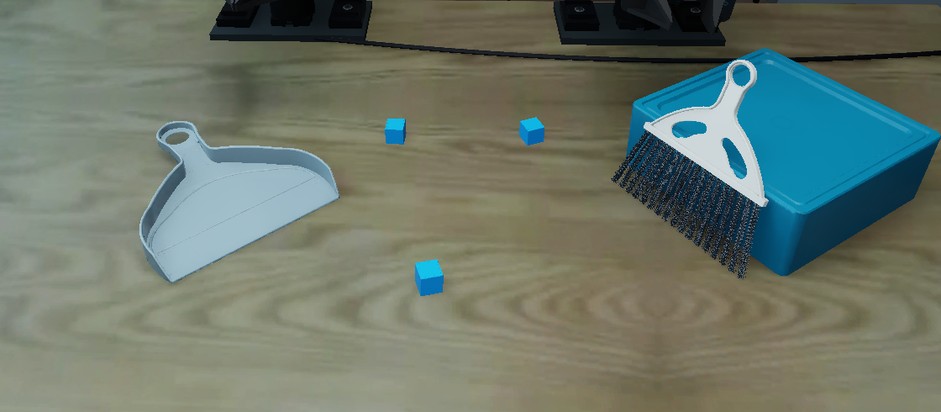}
\FvPhoto{130.78}{256}{138.04}{55}{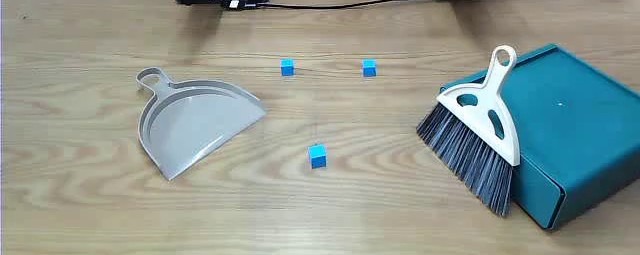}
\FvPhoto{284.82}{256}{90.32}{55}{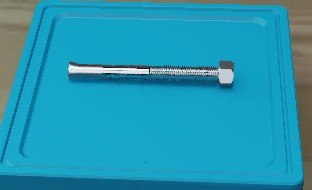}
\FvPhoto{380.13}{256}{89.71}{55}{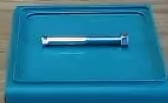}
\FvPhoto{0}{164.2}{77.41}{58}{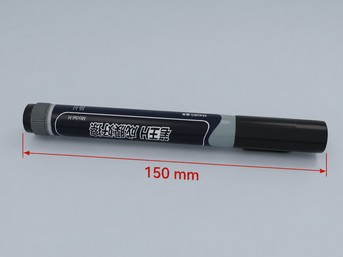}
\FvPhoto{81.41}{164.2}{43.55}{58}{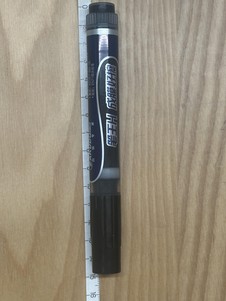}
\FvPhoto{134.96}{198.64}{80}{23.56}{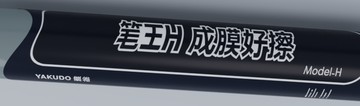}
\FvPhoto{134.96}{164.2}{80}{19.78}{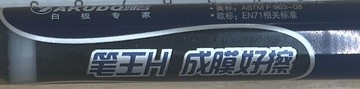}
\FvPhoto{232.57}{164.2}{54.92}{58}{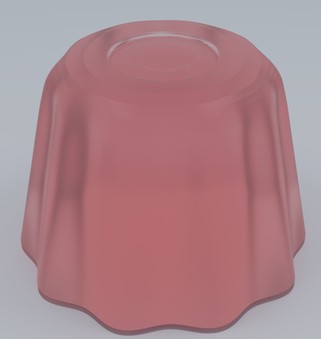}
\FvPhoto{291.49}{164.2}{43.54}{58}{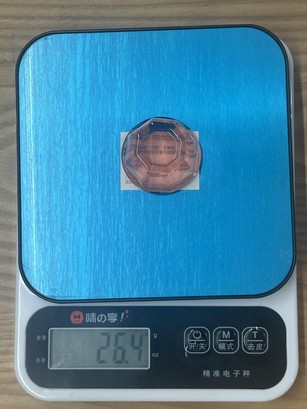}
\FvPhoto{369}{189.7}{41.98}{32.5}{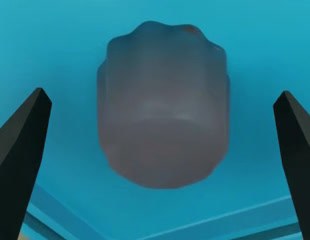}
\FvPhoto{427.67}{189.7}{41.98}{32.5}{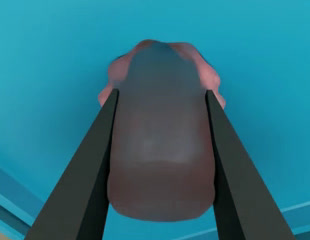}
\FvPhoto{368.65}{151}{42.68}{32.5}{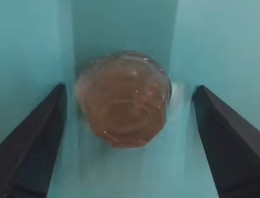}
\FvPhoto{427.32}{151}{42.68}{32.5}{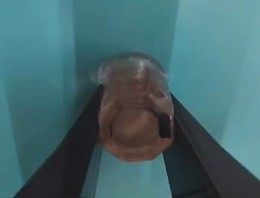}
\FvPhoto{0}{70.2}{94}{47}{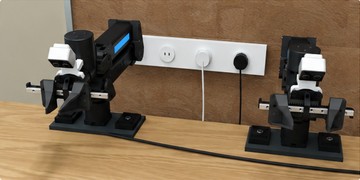}
\FvPhoto{98}{70.2}{94}{47}{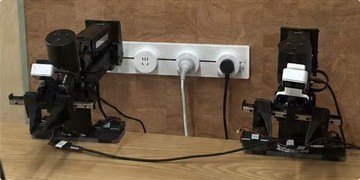}
\FvPhoto{15.67}{6.7}{78.33}{47}{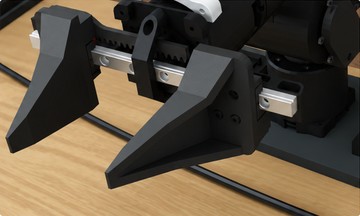}
\FvPhoto{98}{6.7}{78.33}{47}{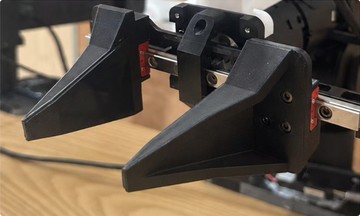}
\FvPhoto{227.77}{70.2}{62.99}{47}{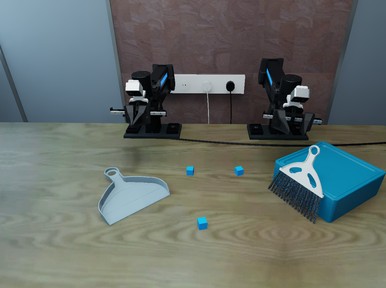}
\FvPhoto{227.93}{6.7}{62.67}{47}{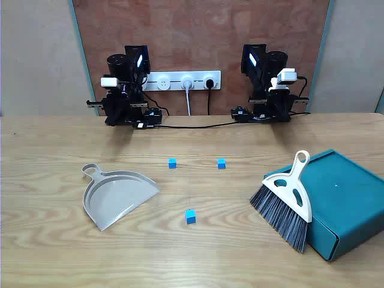}
\FvPhoto{296.76}{70.2}{83.56}{47}{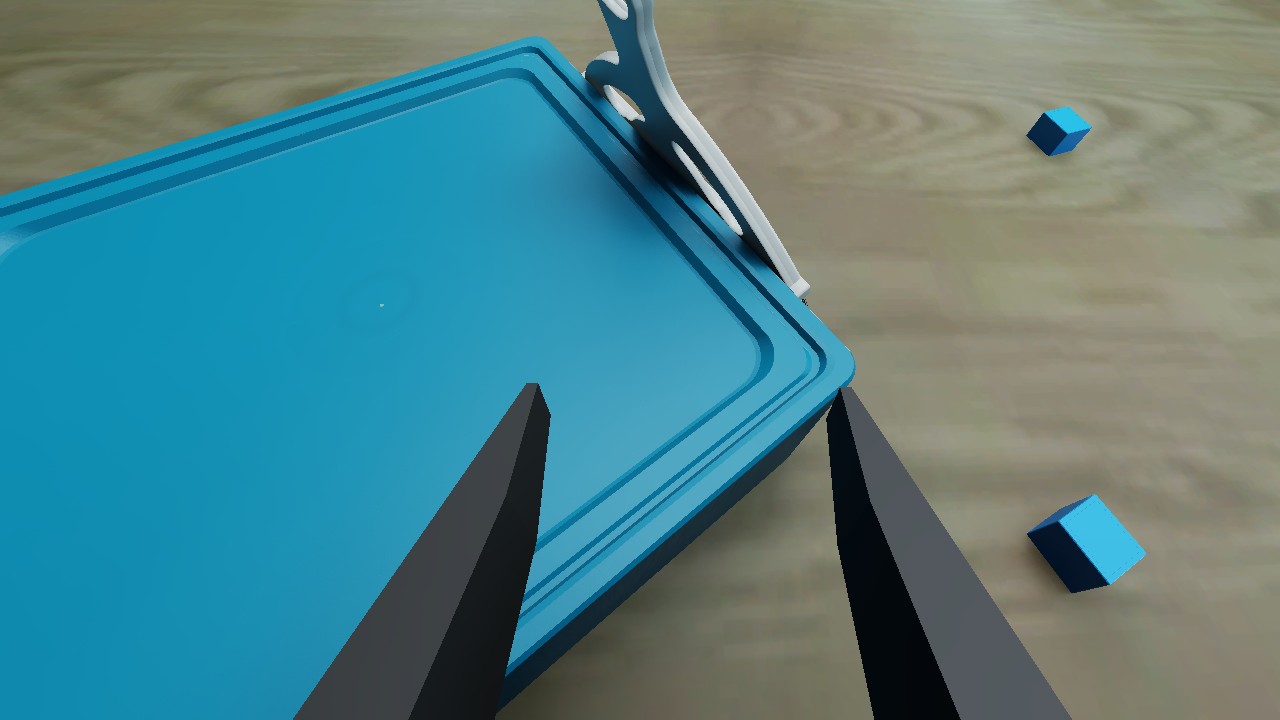}
\FvPhoto{296.76}{6.7}{83.56}{47}{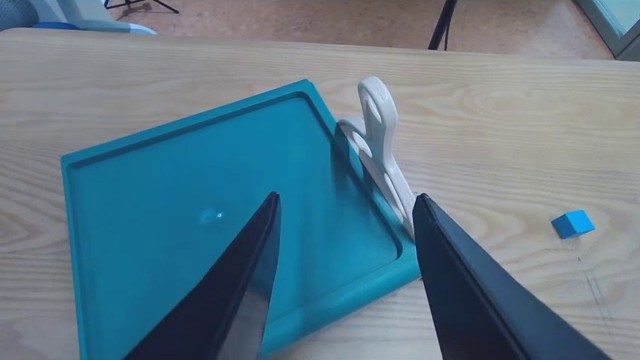}
\FvPhoto{386.32}{70.2}{83.68}{47}{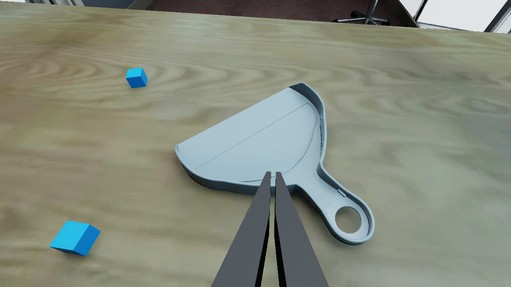}
\FvPhoto{386.32}{6.7}{83.68}{47}{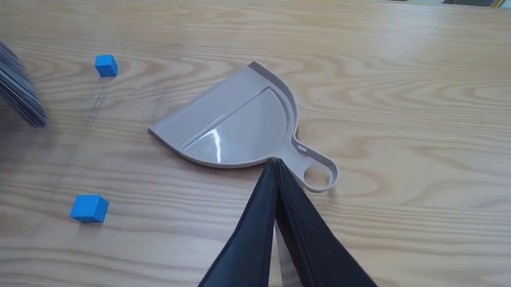}
\FvBar{0.16}{256}{125.62}{PVSim}
\FvBar{130.78}{256}{138.04}{PVReal}
\FvBar{284.82}{256}{90.32}{PVSim}
\FvBar{380.13}{256}{89.71}{PVReal}
\FvBar{0}{164.2}{77.41}{PVSim}
\FvBar{81.41}{164.2}{43.55}{PVReal}
\FvBar{134.96}{198.64}{80}{PVSim}
\FvBar{134.96}{164.2}{80}{PVReal}
\FvBar{232.57}{164.2}{54.92}{PVSim}
\FvBar{291.49}{164.2}{43.54}{PVReal}
\FvBar{369}{189.7}{41.98}{PVSim}
\FvBar{427.67}{189.7}{41.98}{PVSim}
\FvBar{368.65}{151}{42.68}{PVReal}
\FvBar{427.32}{151}{42.68}{PVReal}
\FvBar{0}{70.2}{94}{PVSim}
\FvBar{98}{70.2}{94}{PVReal}
\FvBar{15.67}{6.7}{78.33}{PVSim}
\FvBar{98}{6.7}{78.33}{PVReal}
\FvBar{227.77}{70.2}{62.99}{PVSim}
\FvBar{227.93}{6.7}{62.67}{PVReal}
\FvBar{296.76}{70.2}{83.56}{PVSim}
\FvBar{296.76}{6.7}{83.56}{PVReal}
\FvBar{386.32}{70.2}{83.68}{PVSim}
\FvBar{386.32}{6.7}{83.68}{PVReal}
\node[fv sub] at (134.49,315.5) {\PLTaskChip{L3-04}\hspace{3pt}Block Sweeping};
\node[fv sub] at (377.33,315.5) {\PLTaskChip{L3-02}\hspace{3pt}Nut Unscrewing};
\node[fv sub] at (62.48,226.7) {Capped marker length};
\node[fv sub] at (174.96,226.7) {Surface texture};
\node[fv sub] at (389.98,226.7) {Before};
\node[fv sub] at (448.66,226.7) {Contact};
\node[fv pill, fill=PVSim!13, text=PVSim!88!black, anchor=east] at (60.48,153.4) {SIM\,150\,mm};
\node[fv pill, fill=PVReal!13, text=PVReal!88!black, anchor=west] at (64.48,153.4) {REAL\,150\,mm};
\node[fv pill, fill=PVSim!13, text=PVSim!88!black, anchor=east] at (281.8,153.4) {SIM\,26.4\,g};
\node[fv pill, fill=PVReal!13, text=PVReal!88!black, anchor=west] at (285.8,153.4) {REAL\,26.4\,g};
\node[fv rowchip, fill=PVSim] at (358.65,205.95) {SIM};
\node[fv rowchip, fill=PVReal] at (358.65,167.25) {REAL};
\draw[fv arrow, draw=PVSim] (414.82,205.95) -- (423.82,205.95);
\draw[fv arrow, draw=PVReal] (414.82,167.25) -- (423.82,167.25);
\node[fv sub] at (96,121.7) {Arm bases and mounting};
\node[fv sub] at (96,58.2) {Empty two-finger grippers};
\node[fv sub] at (259.26,121.7) {External view};
\node[fv sub] at (338.54,121.7) {Left wrist view};
\node[fv sub] at (428.16,121.7) {Right wrist view};
\node[fv rowchip, fill=PVSim] at (217.77,93.7) {SIM};
\node[fv rowchip, fill=PVReal] at (217.77,30.2) {REAL};
\end{tikzpicture}
\caption{\textbf{Physical and visual matching of real and simulated environments.}
Violet and orange bars denote simulated and real views, respectively; paired views place simulation to the left of or above the corresponding real view.
(a) Workspace appearance and initial object placement for Block Sweeping (left) and the initial screw--nut engagement for Nut Unscrewing (right).
(b1) Capped marker geometry with simulated and real lengths, and printed surface texture; the real length is documented by a ruler photograph.
(b2) The configured jelly mass in simulation (26.4 g) and the documented scale reading (26.4 g).
(b3) Recorded jelly states before and during gripper contact: simulation at 00:30/00:40 and real world at 01:16/01:22 of their source clips. The same global color-balance adjustment is applied to both simulated frames; object geometry and deformation are unmodified. These qualitative views are not synchronized or a quantitative validation of matched physical response.
(c) Arm-base placement and empty two-finger gripper geometry.
(d) External, left-wrist, and right-wrist views for Block Sweeping. The external views and simulated left-wrist view show the initial state. The real left-wrist frame at 16:40.5 and right-wrist frame at 28:18 are selected during execution to match the gripper--box and gripper--dustpan geometry, respectively. The views are unsynchronized.
}
\label{fig:alignment-pipeline}
\end{figure*}
\arrayrulecolor{black}
\endgroup

%% file: sec/5_evaluating_gpt6_astra.tex
\Needspace{6\baselineskip}
\section{\revision{Evaluation Setup and Protocol}}
\label{sec:evaluation-setup}

We evaluate GPT-6 Astra on physically and visually matched real and simulated tasks to characterize its physical interaction capabilities and their correspondence across environments. This section describes the task and model configuration, model inputs and closed-loop execution, and evaluation protocol and metrics.

\subsection{\revision{Task Setup and Model Configuration}}
\label{sec:workflow}

Each evaluation task is specified by its initial conditions, physical goal, and process constraints, with detailed task definitions provided in Appendix~\ref{app:task-specifications}. Starting from initial configurations matched to the real-world setup, we construct simulated variants by varying target-object positions and orientations (Figure~\ref{fig:task-suite}(b)). These variants reflect the placement variability introduced by manual resets in real-world trials and probe generalization across initial layouts. The task objective and success criteria remain fixed across these variants.

We evaluate GPT-6 Astra with reasoning effort set to \texttt{xhigh}. Codex\footnote{\url{https://github.com/openai/codex}} provides the runtime for invoking the model and connecting it to the environment interfaces.

Zero-shot evaluation is defined with respect to the benchmark tasks: the model is not supplied with prior executions, demonstrations, or task-specific strategies for these tasks. It may use observations, actions, and feedback from the current episode to guide subsequent decisions, without carrying evaluation-task experience into later trials. Section~\ref{sec:observation-action-interface} describes the fixed basic Skills for real-world embodiment and interface adaptation.

\subsection{\revision{Model Inputs and Closed-Loop Execution}}
\label{sec:observation-action-interface}
\label{sec:model-context-configuration}

\noindent\textbf{Inputs.} Each episode provides the model with two fixed inputs (Figure~\ref{fig:execution-contact-rules}(a), \stepmark{1}): the natural-language task instruction, which states the objects, objective, arm roles where required, and constraints (Appendix~\ref{app:task-specifications}), and a description of the execution interface, covering the action space, coordinate conventions, and execution feedback. In simulation, the policy adapter supplies this description as model instructions; in real-world execution, it is provided by the fixed basic Skills prepared before evaluation. Both inputs remain unchanged throughout the episode, and no further task information, hints, or corrections are provided. Image observations, proprioceptive states, execution feedback, and the interaction history are updated as the episode proceeds.

\noindent\textbf{Basic Skills.} Basic Skills follow the Agent Skills mechanism of Codex\footnote{\url{https://developers.openai.com/codex/skills}}: each Skill is a directory with a \texttt{SKILL.md} instruction file and optional reference documents, which the agent reads when relevant. The real-world basic Skills describe the embodiment and interface, including arm identities, arm--camera correspondences, command frames and units, and gripper semantics, and give general guidance on active observation, grasp and contact verification, dual-arm coordination, and local recovery (Figure~\ref{fig:execution-contact-rules}(b); Appendix~\ref{app:basic-skills}). They were prepared before evaluation through general robot interactions and interface testing, contain no strategies or demonstrations from the evaluation tasks, and remain fixed across trials. Simulation uses no Skills: its conventions are fully specified by the simulator configuration, so the policy adapter provides the corresponding description and guidance, including observation guidance adapted from the Skills, as fixed model instructions. Both environments receive the same categories of task-agnostic information by different mechanisms.

\noindent\textbf{Observation.} At each decision, the model observes RGB images from one external camera, which shows the workspace and arm clearance, and two wrist cameras, which show grasps and local contact, together with proprioceptive states and execution feedback (Figure~\ref{fig:execution-contact-rules}(a), \stepmark{2}). In simulation, the adapter attaches these observations to every model call; in real-world execution, the model requests fresh images and states through tool calls. The interaction history retains earlier observations, actions, and feedback within the episode. Without depth images or contact-force measurements, grasp, alignment, and contact outcomes must be inferred from visual changes, proprioceptive states, and execution feedback, and force regulation (Section~\ref{sec:contact-hierarchy}) is exercised only through commanded motion and gripper opening.

\input{figures/figure_execution_contact_rules}

\noindent\textbf{Execution and verification.} In simulation, the model returns one structured action per call (Figure~\ref{fig:execution-contact-rules}(a), \stepmark{3}--\stepmark{4}), giving each commanded arm a target end-effector pose and a gripper opening that sets finger width, not grip force; the cuRobo planner~\citep{sundaralingam2023curobo} computes a collision-checked joint trajectory to the target, Isaac Sim joint drives track it, and planning failures or collisions of non-finger parts stop the action and are reported as feedback. In real-world execution, tool calls drive a control service that runs the robot SDK with transmission limits and protection against lost heartbeat or feedback (Appendix~\ref{app:real-world-setup}), and the Skills direct the model to move in bounded segments. In both environments, execution feedback shows whether a command was completed, but only new observations show whether the intended interaction occurred. The model therefore verifies each outcome (\stepmark{5}) before continuing, correcting, or requesting termination; a termination request does not establish success, which is evaluated independently (Section~\ref{sec:evaluation-metrics}). In Marker Cap Removal (Figure~\ref{fig:execution-contact-rules}(c)), both grasps must remain stable and the cap must separate before the next action.

\Needspace{4\baselineskip}
\subsection{\revision{Evaluation Protocol and Metrics}}
\label{sec:evaluation-metrics}
\label{sec:trial-protocol}

\noindent\textbf{Trial protocol.}
Each trial is one episode that starts from reset object and arm states, and each task is evaluated in three trials per environment. Continuation segments resume unfinished episodes with their interaction history and remain part of the same trial; simulation budgets are not replenished. Trials end differently in the two environments. In simulation, a trial ends when the independent evaluator detects success or a task-defined failure state, or when the model-call or time budget is exhausted; a termination request ends the trial only if the evaluator confirms success. In the real world, no preset task-level wall-clock or model-call limit is imposed. The model may end a trial by reporting completion or judging that it cannot complete the task. Trials also end when an operator stops or takes over execution, or when a commanded motion exceeds the arm's kinematic limits and triggers a protective stop (Appendix~\ref{app:real-world-setup}). Budgets, time limits, initial configurations, and reset procedures are given in Appendix~\ref{app:trial-statistics}.

\noindent\textbf{Success rates.}
A trial is an autonomous success only if it achieves the task goal and satisfies all process constraints (Section~\ref{sec:task-success-criteria}), including release or final-state stability where specified. Both environments use the same success criteria: the evaluator checks them automatically in simulation, whereas real-world outcomes are judged against them after each trial from execution videos (Appendix~\ref{app:result-qualifications}). All other trials count as failures, including budget exhaustion in simulation, real-world termination for inability to complete, protective stops, and operator takeover. For task $i$ in environment $e\in\{\mathrm{sim},\mathrm{real}\}$, the success rate is $\mathrm{SR}_{i,e}=k_{i,e}/n_{i,e}$, where $k_{i,e}$ of $n_{i,e}=3$ trials succeed. For a task set $\mathcal T$ (one interaction level or the full suite), the average success rate $\mathrm{SR}_e(\mathcal T)=|\mathcal T|^{-1}\sum_{i\in\mathcal T}\mathrm{SR}_{i,e}$ weights tasks equally.

\noindent\textbf{Sim--Real and behavioral analysis.}
Simulation and real-world results are compared at two levels: task and level success rates, and the interaction processes that produce them. Execution videos, action records, and robot feedback are used to examine corresponding cases at the same interaction stages: which contacts are established, whether they produce the required object motion, support, or material response, where progress stops, and whether corrective actions restore it. This distinguishes shared interaction difficulties from merely similar final outcomes, and local progress from complete task success; differences in initial configuration or assessment are noted where they affect interpretation. Section~\ref{sec:sim-real-consistency} reports the comparison, Section~\ref{sec:initial-condition-sensitivity} the effect of initial conditions, and Section~\ref{sec:failure-analysis} four interaction bottlenecks, with criteria in Appendix~\ref{app:failure-annotation}.

%% file: figures/figure_execution_contact_rules.tex
\begingroup
\input{figures/comparison_figure_style}
\definecolor{FsAmber}{HTML}{E39B0B}
\tikzset{
  fsbox/.style={draw=PVLine, fill=white, rounded corners=4pt, line width=0.6pt,
    blur shadow={shadow xshift=0pt, shadow yshift=-0.8pt, shadow blur radius=2pt, shadow opacity=16}},
  fstitle/.style={font=\PVSans\fontseries{b}\fontsize{8.5}{10}\selectfont, text=PVInk, inner sep=0pt, anchor=west},
  fstext/.style={font=\PVSans\fontsize{7}{8.6}\selectfont, text=PVInk, inner sep=0pt, align=left},
  fsmuted/.style={font=\PVSans\fontsize{7}{8.6}\selectfont, text=PVMuted, inner sep=0pt, align=left},
  fsarrow/.style={-{Stealth[length=4.6pt,width=4.2pt,round]}, line width=1.1pt, line cap=round, draw=PVBlue},
  fsstep/.style={circle, fill=PVBlue, inner sep=0pt, minimum size=10pt,
    font=\PVMono\fontseries{b}\fontsize{7}{7}\selectfont, text=white},
  fsicon/.style={inner sep=0pt, text=PVBlue!75, font=\fontsize{8.5}{8.5}\selectfont},
  fslanechip/.style={pv chip, minimum width=22pt},
  fspill/.style={rounded corners=5pt, inner xsep=4pt, inner ysep=2pt, line width=0.6pt,
    font=\PVSans\fontseries{sb}\fontsize{7}{7}\selectfont, anchor=north west},
  fsskill/.style={draw=PVBlue!45, fill=white, rounded corners=3pt, line width=0.6pt, inner xsep=3.4pt,
    inner ysep=2.6pt, font=\PVSans\fontsize{7}{8}\selectfont, text=PVInk, anchor=north west},
}
\newcommand{\FigSixThumb}[4]{%
  \PVPhoto{#1}{#2}{27}{20}{#3}
  \node[pv glass, inner xsep=1.8pt, inner ysep=1.2pt, font=\PVMono\fontseries{b}\fontsize{7}{7}\selectfont,
    anchor=north west] at (#1+1.5,#2+18.5) {#4};}
\begin{figure*}[!t]
\centering
\begin{tikzpicture}[x=1pt,y=1pt]
\PVPanelTitle{0}{189.5}{a}{One Decision Cycle}
\node[pv chip, fill=PVReal, anchor=base east] at (300,189.5) {REAL WORLD};
\node[pv chip, fill=PVSim, anchor=base east] at (250,189.5) {SIMULATION};
\path[fsbox] (0,100) rectangle (96,182);
\path[fsbox] (108,100) rectangle (200,182);
\path[rounded corners=4pt, top color=PVBlue!90!PVDeep, bottom color=PVNavy,
  blur shadow={shadow xshift=0pt, shadow yshift=-1pt, shadow blur radius=2.6pt, shadow opacity=32}]
  (212,100) rectangle (300,182);
\path[fsbox] (0,0) rectangle (136,78);
\path[fsbox] (148,0) rectangle (300,78);
\node[fsstep] at (10,172) {1}; \node[fstitle] at (18.5,172) {Inputs};
\node[rounded corners=2pt, fill=PVSky2, text=PVDeep, inner xsep=2.4pt, inner ysep=1.4pt,
  font=\PVSans\fontseries{sb}\fontsize{7}{7}\selectfont, anchor=east] at (92,172) {\faLock\,fixed};
\node[fstext, anchor=north west, text width=88pt] at (5,162) {Task instruction\\Interface description};
\fill[PVSim!10, rounded corners=2pt] (4,126) rectangle (92,138);
\node[fslanechip, fill=PVSim, anchor=west] at (5.5,132) {SIM};
\node[fstext, anchor=west] at (30,132) {model instructions};
\fill[PVReal!10, rounded corners=2pt] (4,112.5) rectangle (92,124.5);
\node[fslanechip, fill=PVReal, anchor=west] at (5.5,118.5) {REAL};
\node[fstext, anchor=west] at (30,118.5) {basic Skills (b)};
\node[fsmuted, anchor=west] at (5,105.5) {No further hints};
\node[fsstep] at (118,172) {2}; \node[fstitle] at (126.5,172) {Observation};
\node[fsicon, anchor=east] at (195,172) {\faCamera};
\FigSixThumb{112}{136}{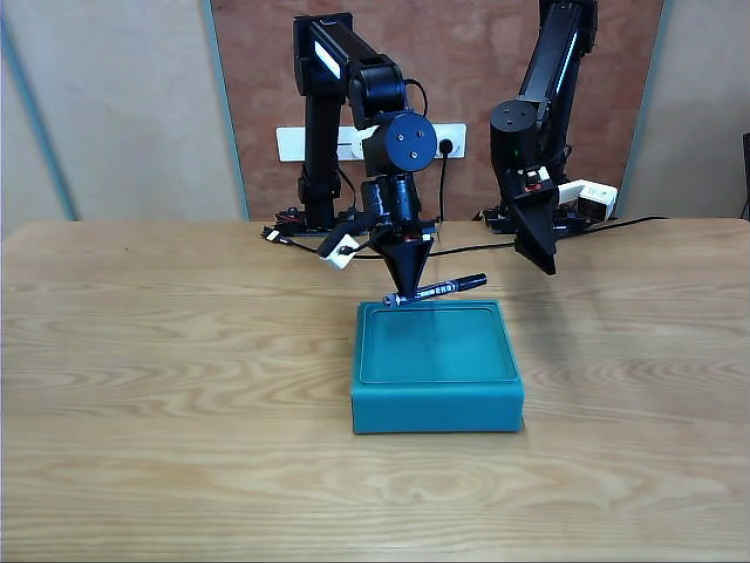}{E}
\FigSixThumb{141}{136}{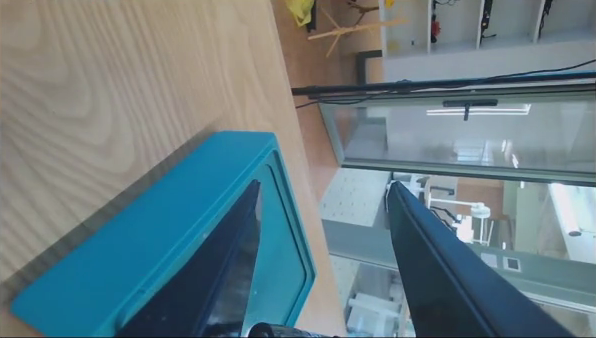}{L}
\FigSixThumb{170}{136}{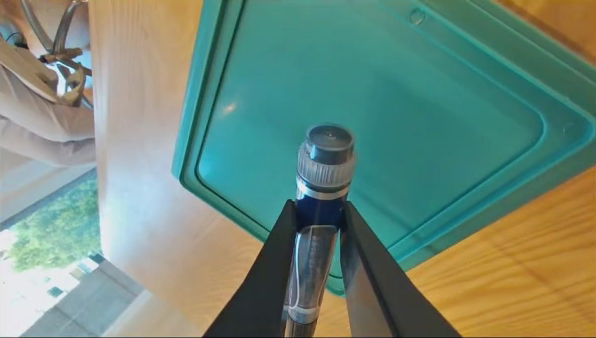}{R}
\node[fsmuted, anchor=west] at (112,126) {Camera views (E, L, R)};
\node[fstext, anchor=west] at (112,115.5) {State \textperiodcentered{} feedback \textperiodcentered{} history};
\node[fsstep, fill=white, text=PVDeep] at (222,172) {3}; \node[fstitle, text=white] at (230.5,172) {Decision};
\node[fsicon, text=white!85!PVBlue, anchor=east] at (295,172) {\faBrain};
\node[inner sep=0pt, font=\PVMono\fontseries{b}\fontsize{11.5}{13.5}\selectfont, text=white, align=center] at (256,141) {GPT-6\\Astra};
\node[inner sep=0pt, font=\PVSans\fontsize{7}{8.6}\selectfont, text=white!85!PVSky, align=center] at (256,111) {selects the next action};
\node[fsstep] at (158,68) {4}; \node[fstitle] at (166.5,68) {Execution};
\node[fsicon, anchor=east] at (295,68) {\faCogs};
\fill[PVSim!10, rounded corners=2pt] (152,31.5) rectangle (296,56.5);
\node[fslanechip, fill=PVSim, anchor=west] at (156,44) {SIM};
\node[fstext, anchor=west, text width=110pt] at (183,44) {Structured action \textrightarrow\\cuRobo planner \textrightarrow{} joint drives};
\fill[PVReal!10, rounded corners=2pt] (152,4) rectangle (296,29);
\node[fslanechip, fill=PVReal, anchor=west] at (156,16.5) {REAL};
\node[fstext, anchor=west, text width=110pt] at (183,16.5) {Tool calls \textrightarrow{} control service\\\textrightarrow{} ARX R5 arms};
\node[fsstep] at (10,68) {5}; \node[fstitle] at (18.5,68) {Verification};
\node[fsicon, anchor=east] at (131,68) {\faClipboardCheck};
\node[fstext, anchor=north west, text width=126pt] at (5,57.5) {Check the outcome:\\\textcolor{PVMuted}{grasp, alignment, separation}};
\node[fspill, draw=PVGood, fill=PVGood!10, text=PVGood!80!black] at (5,26) {Continue};
\node[fspill, draw=FsAmber, fill=FsAmber!12, text=FsAmber!75!black] at (47,26) {Correct};
\node[fspill, draw=PVBad, fill=PVBad!10, text=PVBad!85!black] at (85,26) {Terminate};
\draw[fsarrow] (97,141) -- (107,141);
\draw[fsarrow] (201,141) -- (211,141);
\draw[fsarrow] (256,99) -- (256,79);
\draw[fsarrow] (147,39) -- (137,39);
\draw[fsarrow] (120,79) -- (120,99);
\node[fsmuted, text=PVBlue, anchor=east, font=\PVSans\fontseries{sb}\fontsize{7}{8.6}\selectfont] at (115,89) {\faRedo\,\,Next decision};
\PVPanelTitle{316}{189.5}{b}{Basic Skills}
\node[pv chip, fill=PVReal, anchor=base east] at (470,189.5) {REAL WORLD};
\node[fsmuted, anchor=west] at (316,177) {\textcolor{PVBlue}{\faRobot}\,\,Embodiment and interface};
\node[fsskill] (fs1) at (316,171.5) {Arm\textendash camera mapping};
\node[fsskill] at ([xshift=3pt]fs1.north east) {Gripper semantics};
\node[fsmuted, anchor=west] at (316,150) {\textcolor{PVBlue}{\faCompass}\,\,General guidance};
\node[fsskill] (fs2) at (316,144.5) {Observe and verify};
\node[fsskill] at ([xshift=3pt]fs2.north east) {Dual-arm coordination};
\node[fsskill] at (316,126.5) {Local recovery};
\node[fsmuted, anchor=west] at (316,106) {\textcolor{PVBlue}{\faLock}\,\,\texttt{SKILL.md} files fixed across trials};
\PVPanelTitle{316}{84.5}{c}{Verifying the Outcome}
\node[anchor=base west, inner sep=0pt] at (316,70.5) {\PLTaskChip{L2-01}};
\node[anchor=base west, inner sep=0pt, font=\PVSans\fontseries{sb}\fontsize{8}{9.5}\selectfont, text=PVInk] at (345,70.5) {Marker Cap Removal};
\PVPhoto{316}{4}{74}{58}{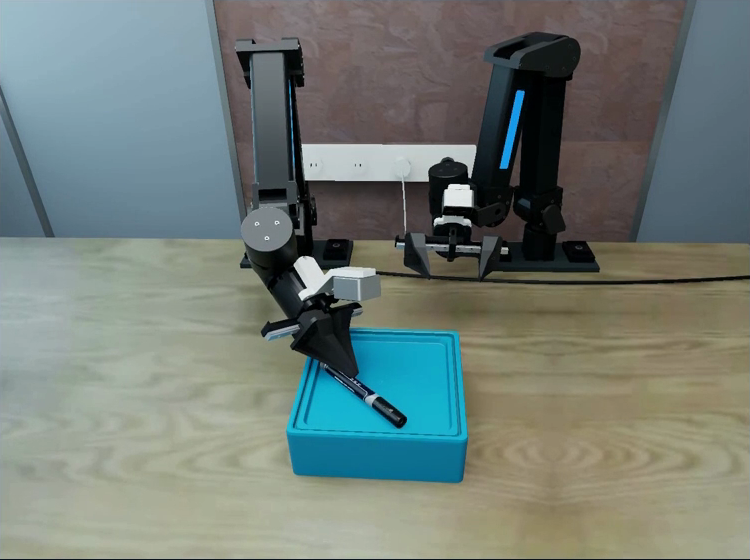}
\PVPhoto{396}{4}{74}{58}{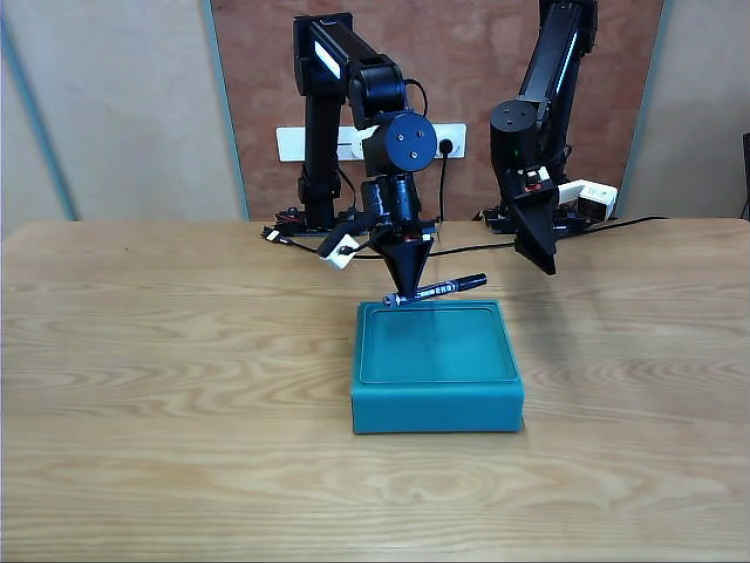}
\node[pv chip, fill=PVSim, anchor=north west] at (319,59) {SIM};
\node[pv chip, fill=PVReal, anchor=north west] at (399,59) {REAL};
\node[anchor=west, inner sep=0pt, font=\PVSans\fontsize{7}{8.6}\selectfont, text=PVInk] at (316,-8)
  {{\color{PVBlue}\faSquare[regular]}\,\,Grasp stability\hspace{1.2em}{\color{PVBlue}\faSquare[regular]}\,\,Cap separation};
\end{tikzpicture}
\caption{\textbf{Model inputs and closed-loop execution.}
(a) One decision cycle. \stepmark{1}~The task instruction and the interface description are fixed for the episode; the interface description is given as model instructions in simulation and by the basic Skills in the real world, and no further hints are provided. \stepmark{2}~The model observes the external and wrist camera views, proprioceptive states, execution feedback, and the interaction history. \stepmark{3}~GPT-6 Astra selects the next action. \stepmark{4}~In simulation, the cuRobo planner and joint drives execute a structured action; in the real world, tool calls drive the ARX R5 arms through a control service. \stepmark{5}~New observations verify the outcome before the model continues, corrects, or requests termination.
(b) The real-world basic Skills describe the embodiment and interface and give general guidance on observation and verification, dual-arm coordination, and local recovery; their \texttt{SKILL.md} files remain fixed across trials (Appendix~\ref{app:basic-skills}).
(c) In matched simulated and real-world Marker Cap Removal scenes, both grasps must remain stable and the cap must separate before the next action.}
\label{fig:execution-contact-rules}
\end{figure*}
\arrayrulecolor{black}
\endgroup

%% file: sec/6_results_and_analysis.tex
\section{Results and Analysis: Contact-Rich Capabilities and Limits}
\label{sec:results}
We examine Astra's zero-shot performance across the task hierarchy, compare real-world and simulated outcomes, and analyze how initial configurations and execution failures shape task completion. The evaluation covers 16 tasks with three trials per task in each environment, using the protocol defined in Section~\ref{sec:evaluation-setup}. Our analysis connects complete-task outcomes with observable intermediate behavior to identify both demonstrated capabilities and limits of sustained physical interaction. Per-task results are reported in Table~\ref{tab:level-performance}.
\input{tables/table02_level_performance}

\subsection{Zero-Shot Performance Across Interaction Levels}
\label{sec:performance-levels}
\label{sec:demonstrated-capabilities}
Across the 16-task suite, Astra achieves average success rates of 45.8\% in the real world and 41.7\% in simulation. The results in Table~\ref{tab:level-performance} show how this performance varies across tasks and interaction levels.

\textcolor{black}{\textbf{Astra demonstrates strong zero-shot performance in basic pick-and-place tasks requiring semantic understanding and object localization, but struggles with tasks requiring more complex spatial reasoning.}} Marker Placement, Charger Alignment, Cup Nesting, and Flower Transfer each succeed in all three real-world trials. Simulation also records successful executions across these tasks, with Marker Placement and Flower Transfer each reported as successful in all three trials and Charger Alignment and Cup Nesting in two of three. These results show that Astra can coordinate object localization, grasping, orientation adjustment, and placement across varied household objects without task-specific training. \textit{Hat Hanging illustrates the limitation in spatial reasoning: the robot must reason about the relative orientation of the hat opening and the coat-rack branch, then rotate the hat into alignment. Although Astra can grasp and lift the hat, it fails to leave it independently supported by the branch in either environment. Consistent with Section~\ref{sec:contact-hierarchy}, spatial reasoning and geometric precision contribute to task difficulty alongside the interaction demands of L1--L3.}

\textcolor{black}{\textbf{Performance declines as physical interaction demands increase.}} Real-world success falls from 80.0\% at L1 to 47.6\% at L2 and 0.0\% at L3; simulation follows the same progression, from 66.7\% to 47.6\% and 0.0\%. None of the four L3 tasks (Bottle Cap Unscrewing, Nut Unscrewing, Stacked-Book Pushing \& Rotation, and Block Sweeping) succeeds in any trial in either environment. Their specifications require coordinated contact reconfiguration and force regulation over successive actions. The absence of complete successes, despite local progress in several executions, identifies sustained physical interaction as a major limitation of the evaluated system and a priority for extending general-purpose models to practical manipulation.

\textcolor{black}{\textbf{The L1--L3 hierarchy captures the increasing challenges of physical interaction and provides a structured benchmark for characterizing the capability boundaries of general-purpose models.}} Grounded in contact configuration and force regulation requirements, it connects task demands to observed execution limits. The decline in completion across both environments reveals challenges in sustained physical interaction that evaluations centered on basic grasping and placement may overlook. This progression provides concrete evaluation targets for advancing reliable physical interaction in general-purpose models.
\FloatBarrier

\subsection{Correspondence Between Real and Simulated Performance}
\label{sec:sim-real-consistency}

\looseness=-1 \textcolor{black}{\textbf{For GPT-6 Astra, real-world and simulated evaluations reveal consistent performance trends across levels and individual tasks.}} As shown in Table~\ref{tab:level-performance}, 10 of the 16 tasks have identical observed success rates (3/3 in three tasks and 0/3 in seven), while the remaining six differ by one successful trial out of three. Both environments exhibit declining completion from L1 to L3, indicating consistent task-level performance profiles.

\textcolor{black}{\textbf{Real and simulated executions exhibit similar intermediate capabilities and recurring interaction bottlenecks.}} Figure~\ref{fig:sim-real-shared-difficulties} illustrates this correspondence through selected interaction stages. In Hat Hanging (L1-03; panel~(a)), the robot can grasp and lift the hat in both environments, but its opening remains misaligned with the rack branch. Completing the task requires complex spatial reasoning about relative orientation beyond locating and lifting the object. Marker Cap Removal (L2-01; panel~(b)) exposes a different obstacle: inadequate cap grasps fail to establish the contact configuration needed for separation, leaving the cap attached. In Jelly Handling (L2-06; panel~(c)), the gripper closes on the jelly, yet the object remains on its support. Grasp location and applied force must work together to sustain a lifting grasp, exposing coupled difficulties in contact configuration and force regulation. Bottle Cap Unscrewing (L3-01; panel~(d)) progresses to cap contact, but subsequent slippage or dropping interrupts the operation, showing difficulty maintaining an effective cap grasp during manipulation. The three paired examples in Block Sweeping (L3-04; panel~(e)) bring together difficulties in directing dispersed blocks toward the dustpan, establishing a useful brush grasp, and regulating brush contact. Successful collection requires coordinating these capabilities as successive sweeps change the block distribution, while avoiding actions that scatter blocks away from the target. These cases link shared local progress to interaction difficulties that prevent complete task success.

\input{figures/figure_sim_real_shared_difficulties}
\input{figures/figure_sim_real_outcome_differences}

\textcolor{black}{\textbf{Differences in task success can coexist with shared capabilities and interaction bottlenecks.}} Figure~\ref{fig:sim-real-outcome-differences} shows why outcome scores need to be interpreted alongside execution behavior. In Charger Alignment (L1-02; panel~(a)), both environments show grasping, transport, and positioning near the socket. Both environments apply the same alignment criterion: real-world trials were judged visually and stopped once alignment appeared adequate, whereas simulation checks it with a fixed geometric tolerance; the unsuccessful simulated trial repeatedly refined alignment until exhausting its model-call budget. The outcome difference therefore reflects assessment granularity and execution limits alongside the difficulty of precise alignment. Cup Nesting (L1-04; panel~(b)) distinguishes placement progress from final-state stability: the cups remain nested after release in the real-world example but topple after release in one simulated trial. This failure was not observed in the three real-world trials. In Book Pulling (L2-02; panel~(c)), the selected configurations illustrate the shared difficulty of choosing effective locations for left-arm pulling and right-arm support. The supporting arm must restrain the upper book while allowing the target book to be extracted; unsuccessful executions reveal inadequate coordination between support and pulling in both environments. The task thus couples contact configuration with bimanual force regulation. Real-world cables and wrist-mounted cameras further restrict access and gripper orientation, making suitable contacts harder to establish. These constraints point to setup details that could be matched more closely across environments. Drawer Pull \& Return (L2-05; panel~(d)) provides complementary evidence of shared partial capability: both environments show local pulling and pushing, including opening the drawer and moving it inward. These actions do not reliably complete the pull-and-return sequence, which requires effective contact and force regulation throughout.

\textcolor{black}{\textbf{For GPT-6 Astra, physically and visually matched simulation captures key capabilities and limitations revealed by real-world evaluation.}} In our visually reconstructed environments with calibrated basic physical properties, the agreement in outcomes and execution behavior indicates that simulated tasks probe challenges encountered on the real platform. This supports assessing competence and persistent interaction bottlenecks in simulation.

\looseness=-1 \textcolor{black}{\textbf{Our simulation benchmark provides a repeatable and scalable testbed whose outcomes correspond to real-world evaluation for GPT-6 Astra.}} Explicit task specifications and controlled initialization support systematic comparisons across models, tasks, and initial configurations. Parallel execution reduces dependence on robot availability and manual resets, enabling broader evaluation.

\subsection{Sensitivity to Initial Conditions}
\label{sec:initial-condition-sensitivity}
\input{figures/figure07_initial_conditions}

\textcolor{black}{\textbf{Initial conditions shape a task's physical interaction demands and its level in the hierarchy.}} Figure~\ref{fig:initial-conditions} illustrates this dependence through two real-world variants of Block Sweeping. With two blocks placed near the dustpan entrance (Figure~\ref{fig:initial-conditions}(a)), collection requires only short, localized sweeps. The main requirement is to regulate the sweeping action within a relatively stable contact mode, consistent with the L2 definition. In the recorded execution, local corrections were sufficient to guide both blocks into the dustpan.

When the initial configuration changes to three dispersed blocks (Figure~\ref{fig:initial-conditions}(b)), a single localized sweep can no longer collect them all. The robot must reposition the brush, select new contact locations and sweep directions, and regulate each sweep as the block distribution changes. This coordination of contact reconfiguration and force regulation places the three-block variant (L3-04) at L3; the recorded execution remains incomplete.

The same principle applies to the initial state of mechanical connections. A nearly unscrewed bottle cap requires only a short turn with torque regulation under a maintained grasp, consistent with L2. The L3-01 configuration instead requires coordinated regrasping and continued turning. As defined in Section~\ref{sec:contact-hierarchy}, levels characterize specific task configurations, whose difficulty can vary continuously with initial conditions.

\subsection{Failure Modes and Interaction Bottlenecks}
\label{sec:failure-analysis}
Drawing on execution records and the cases in Figures~\ref{fig:sim-real-shared-difficulties}, \ref{fig:sim-real-outcome-differences}, and~\ref{fig:additional-failure-cases}, we analyze four recurring interaction bottlenecks across tasks. These bottlenecks may overlap within an execution and do not represent mutually exclusive failure causes. Execution limits end trials but are not interaction bottlenecks.

\input{figures/figure10_additional_failure_cases}

\textcolor{black}{\textbf{Spatial reasoning and execution precision remain bottlenecks beyond object localization.}} Hat Hanging (L1-03; Figure~\ref{fig:sim-real-shared-difficulties}(a)) exposes the difficulty of reasoning about the orientation of the hat opening relative to the rack branch after the object has been located and lifted. Charger Alignment (L1-02; Figure~\ref{fig:sim-real-outcome-differences}(a)) illustrates the complementary demand for execution precision: reaching the socket vicinity does not ensure that subsequent position and orientation adjustments achieve the required alignment. Together, these cases distinguish determining the required spatial relationships from realizing them accurately through robot motion. Block Sweeping (L3-04; Figure~\ref{fig:sim-real-shared-difficulties}(e)) extends this challenge to sweep planning as object positions change.

\textcolor{black}{\textbf{Selecting and establishing appropriate contact configurations remains a recurring challenge.}} A useful contact must combine an accessible location with an orientation that supports the intended action. Marker Cap Removal (L2-01; Figure~\ref{fig:sim-real-shared-difficulties}(b)) exposes difficulties in establishing a precise pulling grasp on the cap. Orange Handling (L2-04; Figure~\ref{fig:additional-failure-cases}(a)) adds material constraints: an upright segment remains unlifted when a suitable grasp cannot be established under the no-damage requirement. Nut Unscrewing (L3-02; Figure~\ref{fig:additional-failure-cases}(c)) further requires contacts that permit subsequent relative rotation; the execution record reports infeasible turning poses despite bimanual contact. Related contact-selection difficulties arise in book extraction, drawer pulling and pushing, and brush pickup (L2-02, L2-05, and L3-04; Figures~\ref{fig:sim-real-outcome-differences}(c,d) and~\ref{fig:sim-real-shared-difficulties}(e)). Contact selection must therefore account for geometry, access, material properties, and force direction.

\textcolor{black}{\textbf{Force regulation remains a key limitation in task execution.}} After contact is established, actions must produce the required physical effect while preserving grasp and support. Book Pulling (L2-02; Figure~\ref{fig:sim-real-outcome-differences}(c)) requires coordinated support and extraction: documented attempts either disturb the upper book or leave the target book blocked, highlighting the need to balance right-arm support with left-arm pulling. Drawer Pull \& Return (L2-05; Figure~\ref{fig:sim-real-outcome-differences}(d)) similarly exposes difficulties in regulating both the outward pull and the inward push through the full sequence. Bottle Cap Unscrewing (L3-01; Figure~\ref{fig:sim-real-shared-difficulties}(d)) illustrates loss of grasp during manipulation. Whiteboard Wiping (L2-07; Figure~\ref{fig:additional-failure-cases}(b)) provides a complementary example in which repeated wiping leaves visible writing. Together, these cases show difficulty producing the required contact effects but do not isolate force regulation from contact configuration or motion execution.

\textcolor{black}{\textbf{Astra struggles to coordinate contact configuration and force regulation within and across interactions.}} Jelly Handling (L2-06; Figure~\ref{fig:sim-real-shared-difficulties}(c)) exposes coupling within a single grasp: contact location and applied force jointly determine whether the deformable object can be supported. The unsuccessful attempts illustrate this coupling without isolating either component as the sole cause. Across successive actions, this coordination must also adapt to the changing physical state. Stacked-Book Pushing \& Rotation (L3-03; Figure~\ref{fig:additional-failure-cases}(d)) achieves partial translation and rotation without completing the required transfer. In Block Sweeping (L3-04), the examples in Figure~\ref{fig:sim-real-shared-difficulties}(e) expose difficulties in sweep planning, brush contact selection, and force regulation; successive sweeps require these capabilities to work together as the block distribution changes. Effective local actions do not reliably accumulate into sustained progress and task completion.

%% file: tables/table02_level_performance.tex
\input{tables/evaluation_result_cells}
\begingroup
\renewcommand{\AstraCallPending}{}
\newcommand{\AstraTableOneRemoveFlag}[1]{%
  \expandafter\patchcmd\csname AstraResult-#1-sim-zs\endcsname
    {\(^{\dagger}\)}{}{}{}}
\forcsvlist{\AstraTableOneRemoveFlag}{L1-05,L2-03,L2-04,L2-07,Main-L1,Main-L2,Main-All}
\begin{table*}[!t]
\centering
\caption{\textbf{Zero-shot performance across tasks and interaction levels.} Real and simulated results show broadly similar performance profiles, with success rates declining from L1 to L3 and no observed L3 successes. Each task has three trials per environment; \(k/n\) denotes successes/trials. All means weight tasks equally.}
\label{tab:level-performance}
\small
\setlength{\tabcolsep}{3pt}
\renewcommand{\arraystretch}{1.12}
\begin{tabularx}{\linewidth}{@{}l>{\raggedright\arraybackslash}X*{2}{>{\centering\arraybackslash}p{0.20\linewidth}}@{}}
\toprule
\textbf{Task ID} & \textbf{Task} & \textbf{Simulation zero-shot} & \textbf{Real-world zero-shot} \\
& & \textbf{SR (\(k/n\))} & \textbf{SR (\(k/n\))} \\
\midrule
\multicolumn{2}{@{}l}{\textbf{L1 mean}} & \AstraResultCell{Main-L1}{sim-zs} & \AstraResultCell{Main-L1}{real-main} \\
L1-01 & \AstraResultCell{L1-01}{name} & \AstraResultCell{L1-01}{sim-zs} & \AstraResultCell{L1-01}{real-main} \\
L1-02 & \AstraResultCell{L1-02}{name} & \AstraResultCell{L1-02}{sim-zs} & \AstraResultCell{L1-02}{real-main} \\
L1-03 & \AstraResultCell{L1-03}{name} & \AstraResultCell{L1-03}{sim-zs} & \AstraResultCell{L1-03}{real-main} \\
L1-04 & \AstraResultCell{L1-04}{name} & \AstraResultCell{L1-04}{sim-zs} & \AstraResultCell{L1-04}{real-main} \\
L1-05 & \AstraResultCell{L1-05}{name} & \AstraResultCell{L1-05}{sim-zs} & \AstraResultCell{L1-05}{real-main} \\
\midrule
\multicolumn{2}{@{}l}{\textbf{L2 mean}} & \AstraResultCell{Main-L2}{sim-zs} & \AstraResultCell{Main-L2}{real-main} \\
L2-01 & \AstraResultCell{L2-01}{name} & \AstraResultCell{L2-01}{sim-zs} & \AstraResultCell{L2-01}{real-main} \\
L2-02 & \AstraResultCell{L2-02}{name} & \AstraResultCell{L2-02}{sim-zs} & \AstraResultCell{L2-02}{real-main} \\
L2-03 & \AstraResultCell{L2-03}{name} & \AstraResultCell{L2-03}{sim-zs} & \AstraResultCell{L2-03}{real-main} \\
L2-04 & \AstraResultCell{L2-04}{name} & \AstraResultCell{L2-04}{sim-zs} & \AstraResultCell{L2-04}{real-main} \\
L2-05 & \AstraResultCell{L2-05}{name} & \AstraResultCell{L2-05}{sim-zs} & \AstraResultCell{L2-05}{real-main} \\
L2-06 & \AstraResultCell{L2-06}{name} & \AstraResultCell{L2-06}{sim-zs} & \AstraResultCell{L2-06}{real-main} \\
L2-07 & \AstraResultCell{L2-07}{name} & \AstraResultCell{L2-07}{sim-zs} & \AstraResultCell{L2-07}{real-main} \\
\midrule
\multicolumn{2}{@{}l}{\textbf{L3 mean}} & \AstraResultCell{Main-L3}{sim-zs} & \AstraResultCell{Main-L3}{real-main} \\
L3-01 & \AstraResultCell{L3-01}{name} & \AstraResultCell{L3-01}{sim-zs} & \AstraResultCell{L3-01}{real-main} \\
L3-02 & \AstraResultCell{L3-02}{name} & \AstraResultCell{L3-02}{sim-zs} & \AstraResultCell{L3-02}{real-main} \\
L3-03 & \AstraResultCell{L3-03}{name} & \AstraResultCell{L3-03}{sim-zs} & \AstraResultCell{L3-03}{real-main} \\
L3-04 & \AstraResultCell{L3-04}{name} & \AstraResultCell{L3-04}{sim-zs} & \AstraResultCell{L3-04}{real-main} \\
\midrule
\multicolumn{2}{@{}l}{\textbf{Overall mean}} & \AstraResultCell{Main-All}{sim-zs} & \AstraResultCell{Main-All}{real-main} \\
\bottomrule
\end{tabularx}
\end{table*}
\endgroup

%% file: tables/evaluation_result_cells.tex
\ifdefined\AstraEvaluationDataLoaded\else
\def\AstraEvaluationDataLoaded{1}
\newcommand{\AstraSetResultCell}[3]{%
  \expandafter\gdef\csname AstraResult-#1-#2\endcsname{#3}}
\newcommand{\AstraResultCell}[2]{\csname AstraResult-#1-#2\endcsname}
\newcommand{\AstraCallPending}{\textcolor{red}{\textit{TBD}}}
\newcommand{\AstraCallCell}[2]{%
  \ifcsname AstraResult-#1-#2-calls\endcsname
    \csname AstraResult-#1-#2-calls\endcsname
  \else\AstraCallPending\fi}

\AstraSetResultCell{L1-01}{name}{Marker Placement}
\AstraSetResultCell{L1-01}{sim-zs}{100\% (3/3)}
\AstraSetResultCell{L1-01}{real-skills}{100\% (3/3)}

\AstraSetResultCell{L1-02}{name}{Charger Alignment}
\AstraSetResultCell{L1-02}{sim-zs}{66.7\% (2/3)}
\AstraSetResultCell{L1-02}{real-skills}{100\% (3/3)}

\AstraSetResultCell{L1-03}{name}{Hat Hanging}
\AstraSetResultCell{L1-03}{sim-zs}{0\% (0/3)}
\AstraSetResultCell{L1-03}{real-skills}{0\% (0/3)}

\AstraSetResultCell{L1-04}{name}{Cup Nesting}
\AstraSetResultCell{L1-04}{sim-zs}{66.7\% (2/3)}
\AstraSetResultCell{L1-04}{real-skills}{100\% (3/3)}

\AstraSetResultCell{L1-05}{name}{Flower Transfer}
\AstraSetResultCell{L1-05}{sim-zs}{100\% (3/3)\(^{\dagger}\)}
\AstraSetResultCell{L1-05}{real-skills}{100\% (3/3)}

\AstraSetResultCell{L2-01}{name}{Marker Cap Removal}
\AstraSetResultCell{L2-01}{sim-zs}{0\% (0/3)}
\AstraSetResultCell{L2-01}{real-skills}{0\% (0/3)}

\AstraSetResultCell{L2-02}{name}{Book Pulling}
\AstraSetResultCell{L2-02}{sim-zs}{33.3\% (1/3)}
\AstraSetResultCell{L2-02}{real-skills}{0\% (0/3)}

\AstraSetResultCell{L2-03}{name}{Marble Pouring}
\AstraSetResultCell{L2-03}{sim-zs}{100\% (3/3)\(^{\dagger}\)}
\AstraSetResultCell{L2-03}{real-skills}{100\% (3/3)}

\AstraSetResultCell{L2-04}{name}{Orange Handling}
\AstraSetResultCell{L2-04}{sim-zs}{66.7\% (2/3)\(^{\dagger}\)}
\AstraSetResultCell{L2-04}{real-skills}{100\% (3/3)}

\AstraSetResultCell{L2-05}{name}{Drawer Pull \& Return}
\AstraSetResultCell{L2-05}{sim-zs}{33.3\% (1/3)}
\AstraSetResultCell{L2-05}{real-skills}{66.7\% (2/3)}

\AstraSetResultCell{L2-06}{name}{Jelly Handling}
\AstraSetResultCell{L2-06}{sim-zs}{0\% (0/3)}
\AstraSetResultCell{L2-06}{real-skills}{0\% (0/3)}

\AstraSetResultCell{L2-07}{name}{Whiteboard Wiping}
\AstraSetResultCell{L2-07}{sim-zs}{100\% (3/3)\(^{\dagger}\)}
\AstraSetResultCell{L2-07}{real-skills}{66.7\% (2/3)}

\AstraSetResultCell{L3-01}{name}{Bottle Cap Unscrewing}
\AstraSetResultCell{L3-01}{sim-zs}{0\% (0/3)}
\AstraSetResultCell{L3-01}{real-skills}{0\% (0/3)}

\AstraSetResultCell{L3-02}{name}{Nut Unscrewing}
\AstraSetResultCell{L3-02}{sim-zs}{0\% (0/3)}
\AstraSetResultCell{L3-02}{real-skills}{0\% (0/3)}

\AstraSetResultCell{L3-03}{name}{Stacked-Book Pushing \& Rotation}
\AstraSetResultCell{L3-03}{sim-zs}{0\% (0/3)}
\AstraSetResultCell{L3-03}{real-skills}{0\% (0/3)}

\AstraSetResultCell{L3-04}{name}{Block Sweeping}
\AstraSetResultCell{L3-04}{sim-zs}{0\% (0/3)}
\AstraSetResultCell{L3-04}{real-skills}{0\% (0/3)}

\AstraSetResultCell{L1-01}{real-main}{\AstraResultCell{L1-01}{real-skills}}
\AstraSetResultCell{L1-01}{real-main-calls}{\AstraCallCell{L1-01}{real-skills}}
\AstraSetResultCell{L1-02}{real-main}{\AstraResultCell{L1-02}{real-skills}}
\AstraSetResultCell{L1-02}{real-main-calls}{\AstraCallCell{L1-02}{real-skills}}
\AstraSetResultCell{L1-03}{real-main}{\AstraResultCell{L1-03}{real-skills}}
\AstraSetResultCell{L1-03}{real-main-calls}{\AstraCallCell{L1-03}{real-skills}}
\AstraSetResultCell{L1-04}{real-main}{\AstraResultCell{L1-04}{real-skills}}
\AstraSetResultCell{L1-04}{real-main-calls}{\AstraCallCell{L1-04}{real-skills}}
\AstraSetResultCell{L1-05}{real-main}{\AstraResultCell{L1-05}{real-skills}}
\AstraSetResultCell{L1-05}{real-main-calls}{\AstraCallCell{L1-05}{real-skills}}
\AstraSetResultCell{L2-01}{real-main}{\AstraResultCell{L2-01}{real-skills}}
\AstraSetResultCell{L2-01}{real-main-calls}{\AstraCallCell{L2-01}{real-skills}}
\AstraSetResultCell{L2-02}{real-main}{\AstraResultCell{L2-02}{real-skills}}
\AstraSetResultCell{L2-02}{real-main-calls}{\AstraCallCell{L2-02}{real-skills}}
\AstraSetResultCell{L2-03}{real-main}{\AstraResultCell{L2-03}{real-skills}}
\AstraSetResultCell{L2-03}{real-main-calls}{\AstraCallCell{L2-03}{real-skills}}
\AstraSetResultCell{L2-04}{real-main}{\AstraResultCell{L2-04}{real-skills}}
\AstraSetResultCell{L2-04}{real-main-calls}{\AstraCallCell{L2-04}{real-skills}}
\AstraSetResultCell{L2-05}{real-main}{\AstraResultCell{L2-05}{real-skills}}
\AstraSetResultCell{L2-05}{real-main-calls}{\AstraCallCell{L2-05}{real-skills}}
\AstraSetResultCell{L2-06}{real-main}{\AstraResultCell{L2-06}{real-skills}}
\AstraSetResultCell{L2-06}{real-main-calls}{\AstraCallCell{L2-06}{real-skills}}
\AstraSetResultCell{L2-07}{real-main}{\AstraResultCell{L2-07}{real-skills}}
\AstraSetResultCell{L2-07}{real-main-calls}{\AstraCallCell{L2-07}{real-skills}}
\AstraSetResultCell{L3-01}{real-main}{\AstraResultCell{L3-01}{real-skills}}
\AstraSetResultCell{L3-01}{real-main-calls}{\AstraCallCell{L3-01}{real-skills}}
\AstraSetResultCell{L3-02}{real-main}{\AstraResultCell{L3-02}{real-skills}}
\AstraSetResultCell{L3-02}{real-main-calls}{\AstraCallCell{L3-02}{real-skills}}
\AstraSetResultCell{L3-03}{real-main}{\AstraResultCell{L3-03}{real-skills}}
\AstraSetResultCell{L3-03}{real-main-calls}{\AstraCallCell{L3-03}{real-skills}}
\AstraSetResultCell{L3-04}{real-main}{\AstraResultCell{L3-04}{real-skills}}
\AstraSetResultCell{L3-04}{real-main-calls}{\AstraCallCell{L3-04}{real-skills}}

\AstraSetResultCell{Main-L1}{real-main}{80.0\% (12/15)}
\AstraSetResultCell{Main-L2}{real-main}{47.6\% (10/21)}
\AstraSetResultCell{Main-L3}{real-main}{0.0\% (0/12)}
\AstraSetResultCell{Main-All}{real-main}{45.8\% (22/48)}
\AstraSetResultCell{Main-L1}{sim-zs}{66.7\% (10/15)\(^{\dagger}\)}
\AstraSetResultCell{Main-L2}{sim-zs}{47.6\% (10/21)\(^{\dagger}\)}
\AstraSetResultCell{Main-L3}{sim-zs}{0.0\% (0/12)}
\AstraSetResultCell{Main-All}{sim-zs}{41.7\% (20/48)\(^{\dagger}\)}
\AstraSetResultCell{L1-01}{real-skills-calls}{124}
\AstraSetResultCell{L1-02}{real-skills-calls}{188}
\AstraSetResultCell{L1-03}{real-skills-calls}{311}
\AstraSetResultCell{L1-04}{real-skills-calls}{236}
\AstraSetResultCell{L1-05}{real-skills-calls}{156}
\AstraSetResultCell{L2-01}{real-skills-calls}{392}
\AstraSetResultCell{L2-02}{real-skills-calls}{137}
\AstraSetResultCell{L2-03}{real-skills-calls}{149}
\AstraSetResultCell{L2-04}{real-skills-calls}{243}
\AstraSetResultCell{L2-05}{real-skills-calls}{168}
\AstraSetResultCell{L2-06}{real-skills-calls}{179}
\AstraSetResultCell{L2-07}{real-skills-calls}{154}
\AstraSetResultCell{L3-01}{real-skills-calls}{146}
\AstraSetResultCell{L3-02}{real-skills-calls}{284}
\AstraSetResultCell{L3-03}{real-skills-calls}{309}
\AstraSetResultCell{L3-04}{real-skills-calls}{242}
\AstraSetResultCell{Main-L1}{real-main-calls}{203}
\AstraSetResultCell{Main-L2}{real-main-calls}{203}
\AstraSetResultCell{Main-L3}{real-main-calls}{245}
\AstraSetResultCell{Main-All}{real-main-calls}{213}
\fi

%% file: figures/figure_sim_real_shared_difficulties.tex
\begingroup
\input{figures/comparison_figure_style}
\newcommand{\FigSevenPairBand}[1]{\parbox{\dimexpr0.5\linewidth-3mm\relax}{\PLEnvLabel{#1}}}
\newcommand{\FigSevenMechanism}[1]{{\ComparisonFont\fontsize{8}{9.5}\selectfont\color{PLNavy}#1}}
\begin{figure*}[!p]
\centering
\fontsize{8}{9.5}\selectfont
\begin{minipage}[t]{0.484\linewidth}
\PLPanelHead{a}{L1-03}{Hat Hanging}{\PLBottleneckSpatial}{\PLTally{L1-03}}
\vspace{3pt}
\begin{tabularx}{\linewidth}{@{}X@{\hspace{3mm}}X@{}}
\PLEnvLabel{sim} & \PLEnvLabel{real}\\[2pt]
\ComparisonImage{33mm}{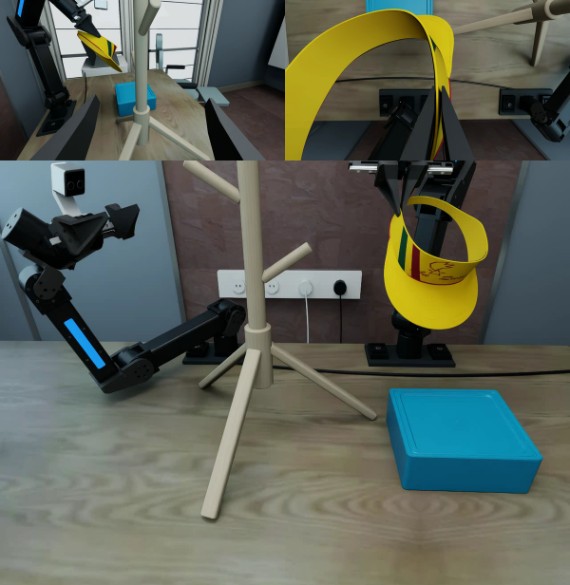}{Opening not aligned with branch} &
\ComparisonImage{33mm}{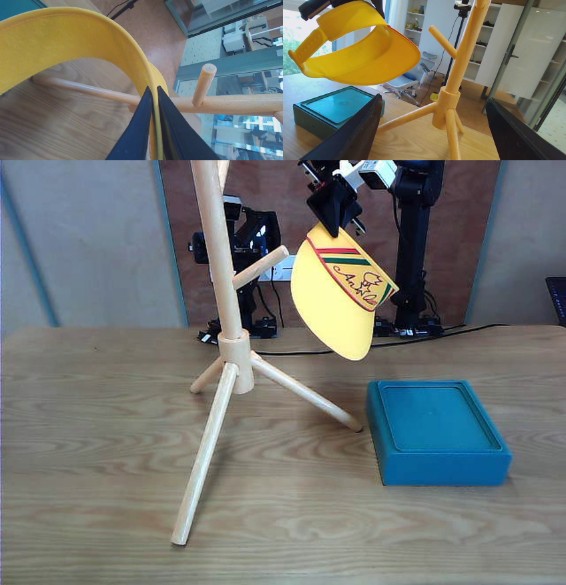}{Opening not aligned with branch}\\
\end{tabularx}
\end{minipage}\hfill
\begin{minipage}[t]{0.484\linewidth}
\PLPanelHead{b}{L2-01}{Marker Cap Removal}{\PLBottleneckContact}{\PLTally{L2-01}}
\vspace{3pt}
\begin{tabularx}{\linewidth}{@{}X@{\hspace{3mm}}X@{}}
\PLEnvLabel{sim} & \PLEnvLabel{real}\\[2pt]
\ComparisonImage{33mm}{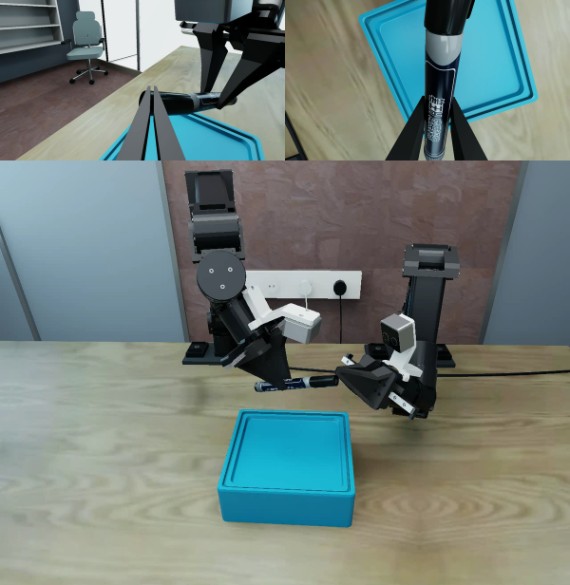}{Cap remains attached} &
\ComparisonImage{33mm}{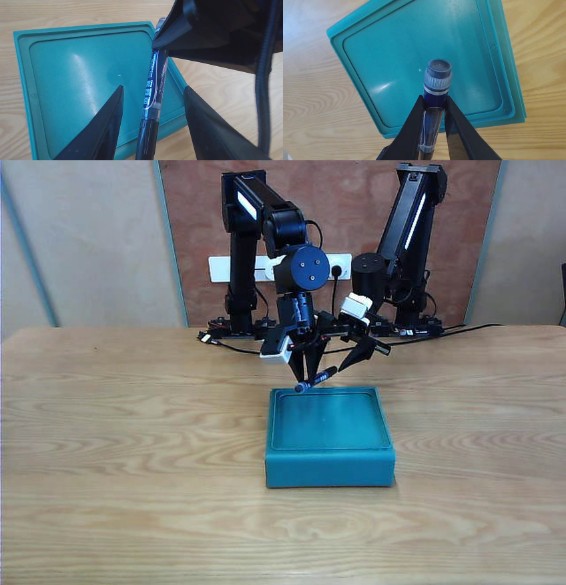}{Cap remains attached}\\
\end{tabularx}
\end{minipage}
\par\vspace{8.5pt}
\PLPanelHead{c}{L2-06}{Jelly Handling}{\PLBottleneckCoordination}{\PLTally{L2-06}}
\vspace{3pt}
\begin{tabularx}{\linewidth}{@{}X@{\hspace{1.5mm}\PLArrow{PLSim}{14mm}\hspace{1.5mm}}X@{\PLGutter}X@{\hspace{1.5mm}\PLArrow{PLReal}{14mm}\hspace{1.5mm}}X@{}}
\multicolumn{2}{@{}l@{}}{\FigSevenPairBand{sim}} & \multicolumn{2}{@{}r@{}}{\FigSevenPairBand{real}}\\[2pt]
\ComparisonImage{33mm}{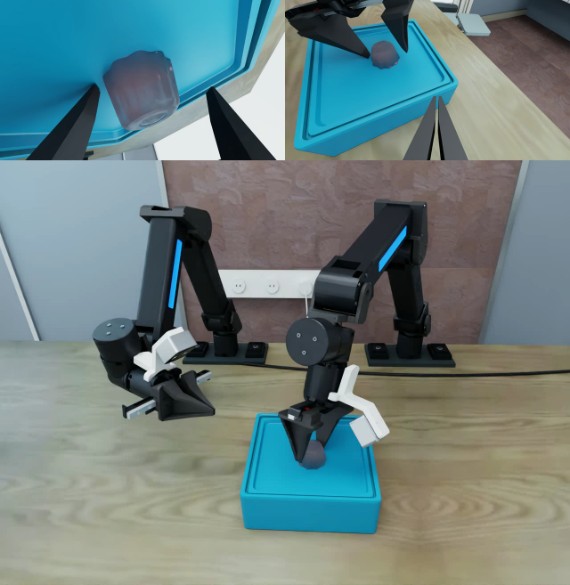}{Gripper closes on jelly} &
\ComparisonImage{33mm}{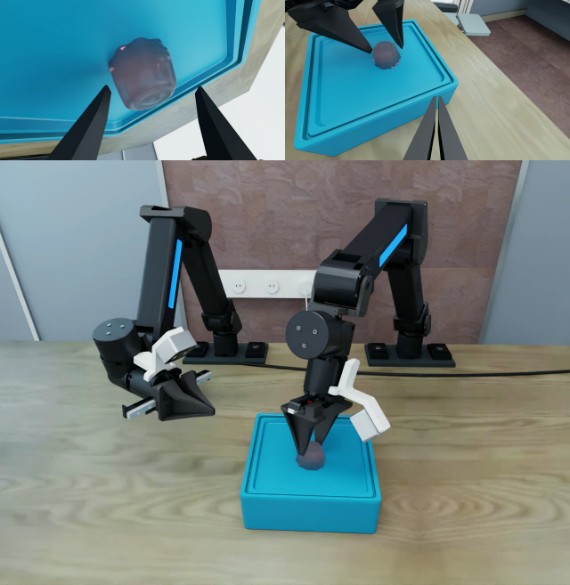}{Jelly remains on support} &
\ComparisonImage{33mm}{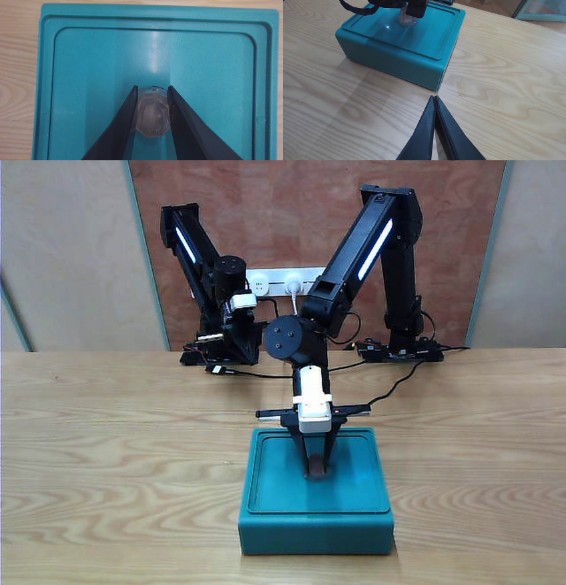}{Gripper closes on jelly} &
\ComparisonImage{33mm}{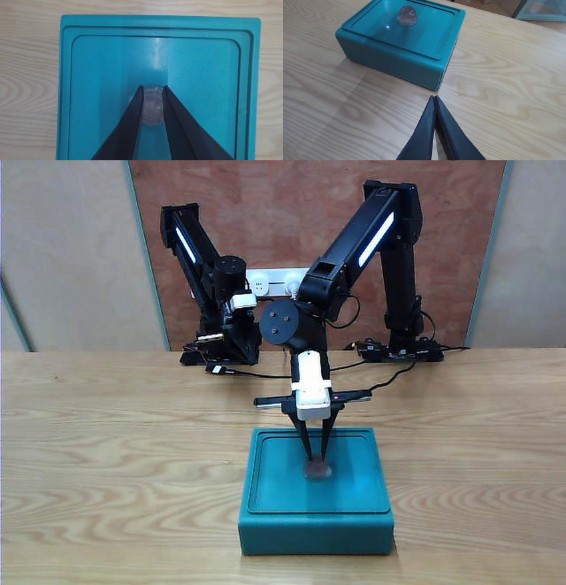}{Jelly remains on support}\\
\end{tabularx}
\par\vspace{8.5pt}
\PLPanelHead{d}{L3-01}{Bottle Cap Unscrewing}{\PLBottleneckForce}{\PLTally{L3-01}}
\vspace{3pt}
\begin{tabularx}{\linewidth}{@{}X@{\hspace{1.5mm}\PLArrow{PLSim}{13mm}\hspace{1.5mm}}X@{\PLGutter}X@{\hspace{1.5mm}\PLArrow{PLReal}{13mm}\hspace{1.5mm}}X@{}}
\multicolumn{2}{@{}l@{}}{\FigSevenPairBand{sim}} & \multicolumn{2}{@{}r@{}}{\FigSevenPairBand{real}}\\[2pt]
\ComparisonImage{30mm}{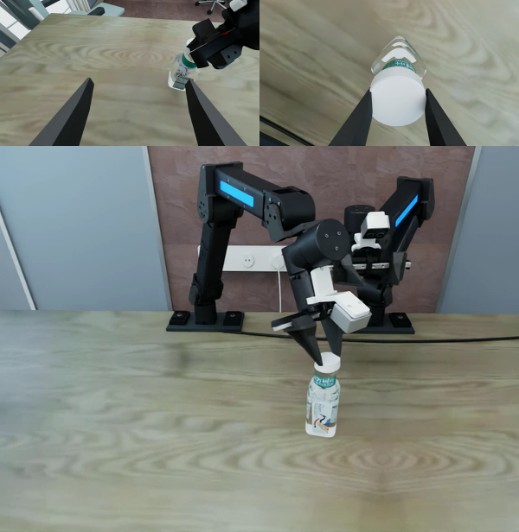}{Gripper at bottle cap} &
\ComparisonImage{30mm}{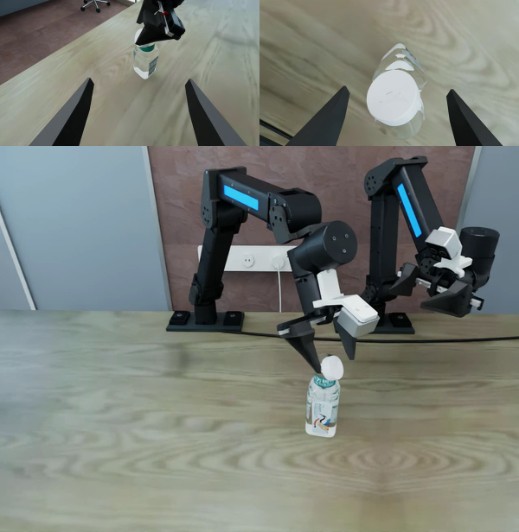}{Cap slipping from grasp} &
\ComparisonImage{30mm}{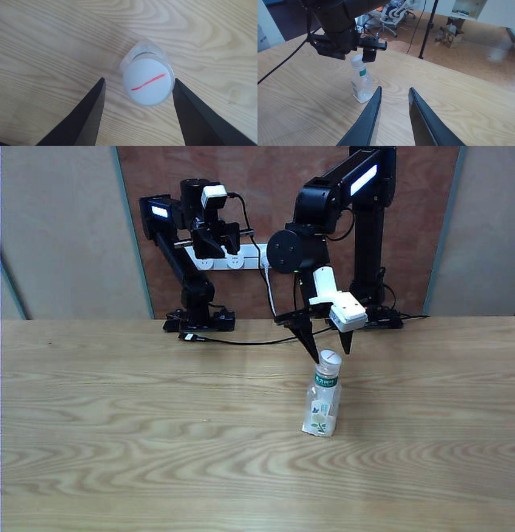}{Gripper contacts cap} &
\ComparisonImage{30mm}{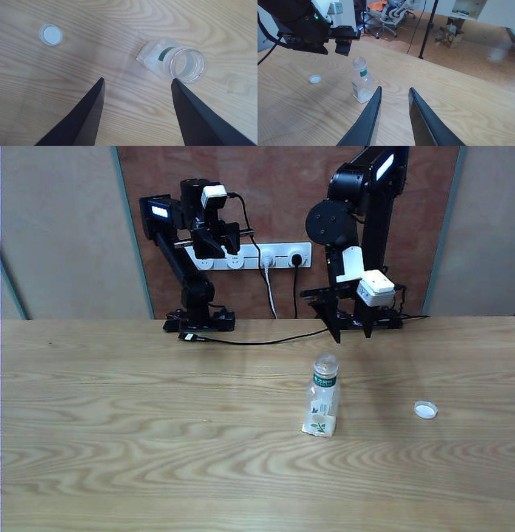}{Cap dropped onto table}\\
\end{tabularx}
\par\vspace{8.5pt}
\PLPanelHead{e}{L3-04}{Block Sweeping}{\PLBottleneckCoordination}{\PLTally{L3-04}}
\vspace{3pt}
\begin{tabularx}{\linewidth}{@{}X@{\hspace{1.5mm}}X@{\hspace{4mm}}X@{\hspace{1.5mm}}X@{\hspace{4mm}}X@{\hspace{1.5mm}}X@{}}
\multicolumn{2}{c}{\FigSevenMechanism{\PLIcon{crosshairs}Spatial reasoning}} &
\multicolumn{2}{c}{\FigSevenMechanism{\PLIcon{hand-paper}Contact selection}} &
\multicolumn{2}{c}{\FigSevenMechanism{\PLIcon{compress-arrows-alt}Force regulation}}\\[2pt]
\PLEnvLabel{sim} & \PLEnvLabel{real} & \PLEnvLabel{sim} & \PLEnvLabel{real} & \PLEnvLabel{sim} & \PLEnvLabel{real}\\[2pt]
\ComparisonImage{28mm}{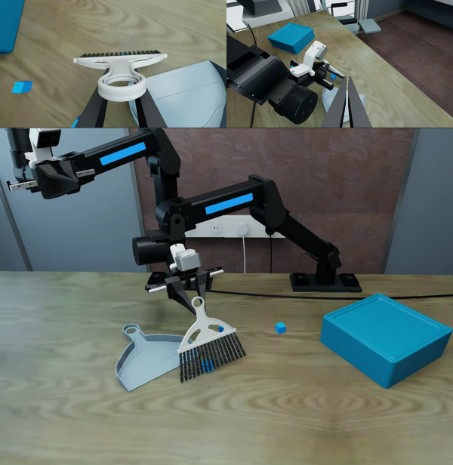}{Blocks remain outside dustpan} &
\ComparisonImage{28mm}{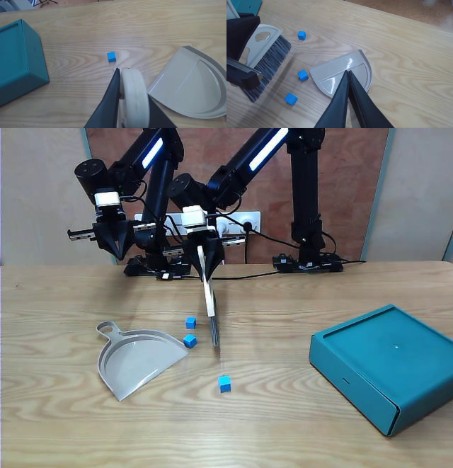}{Blocks remain outside dustpan} &
\ComparisonImage{28mm}{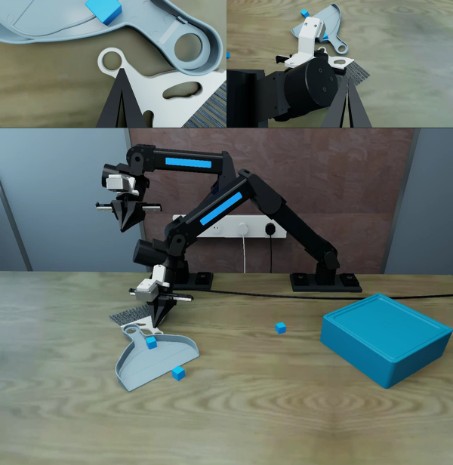}{Gripper at handle loop} &
\ComparisonImage{28mm}{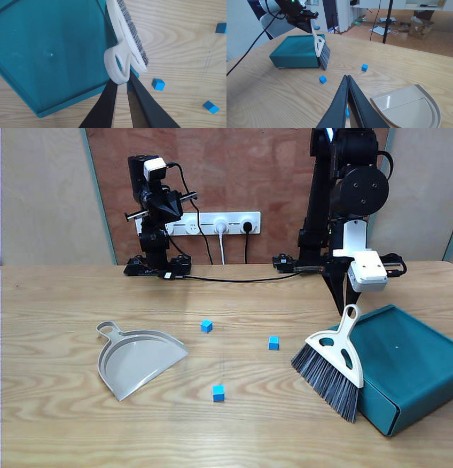}{Gripper at handle loop} &
\ComparisonImage{28mm}{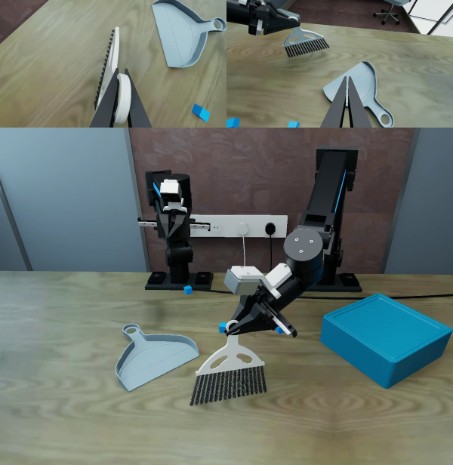}{Block displaced away from dustpan} &
\ComparisonImage{28mm}{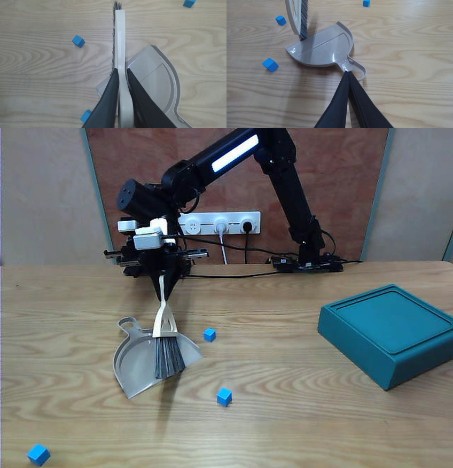}{Brush contact at the dustpan lip}\\
\end{tabularx}
\caption{\textbf{Failure-pattern comparison across real and simulated executions.} (a) Hat--rack alignment. (b) Cap contact during uncapping. (c) Unsuccessful jelly lifting under coupled contact and force demands. (d) Cap slippage. (e) Coordination of spatial reasoning, contact selection, and force regulation during sweeping. Headers give the task ID, interaction bottleneck (Section~\ref{sec:failure-analysis}), and success counts in simulation and the real world (Table~\ref{tab:level-performance}). Multiview images place the wrist views above the external view. Labels describe observable states.}
\label{fig:sim-real-shared-difficulties}
\end{figure*}
\arrayrulecolor{black}
\endgroup

%% file: figures/figure_sim_real_outcome_differences.tex
\begingroup
\input{figures/comparison_figure_style}
\newcommand{\FigEightHalf}[1]{\begin{minipage}{\dimexpr.5\linewidth-3mm\relax}\centering#1\end{minipage}}
\newcommand{\FigEightConfigurations}[1]{{\ComparisonFont\color{#1}\fontsize{7.5}{9}\selectfont Selected configurations}}
\begin{figure*}[!t]
\centering
\FigEightHalf{\PLEnvLabel{sim}}\hfill\FigEightHalf{\PLEnvLabel{real}}
\par\vspace{4pt}
\PLPanelHead{a}{L1-02}{Charger Alignment}{\PLBottleneckSpatial}{\PLTally{L1-02}}
\vspace{4pt}
\begin{tabularx}{\linewidth}{@{}X@{\hspace{1.5mm}\PLArrow{PLSim}{12.5mm}\hspace{1.5mm}}X@{\PLGutter}X@{\hspace{1.5mm}\PLArrow{PLReal}{12.5mm}\hspace{1.5mm}}X@{}}
\ComparisonImage{26mm}{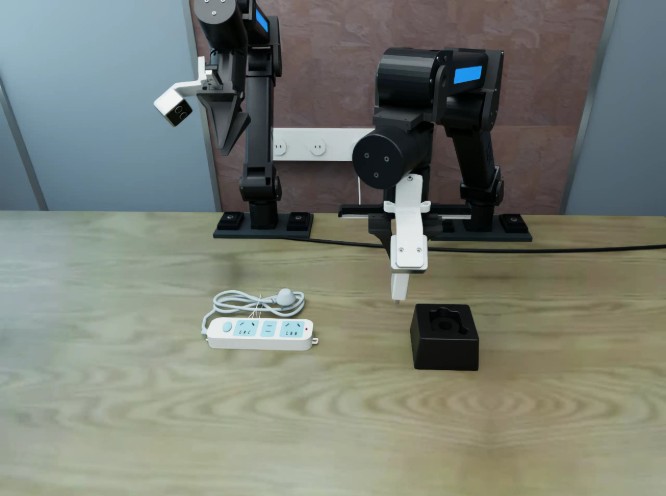}{Grasped and transported} &
\ComparisonImage{26mm}{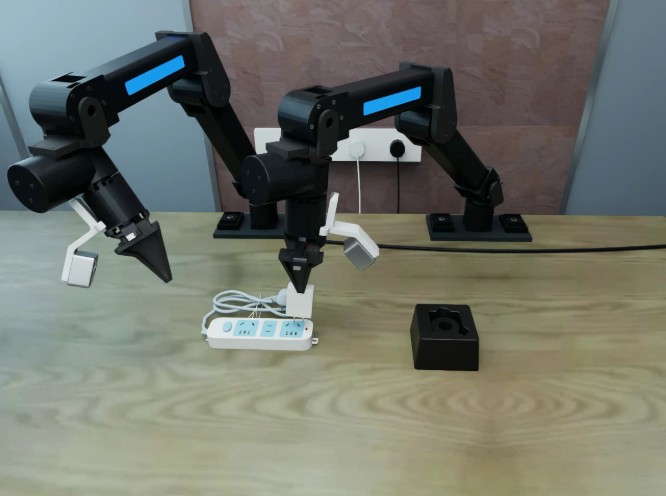}{Held near the target socket} &
\ComparisonImage{26mm}{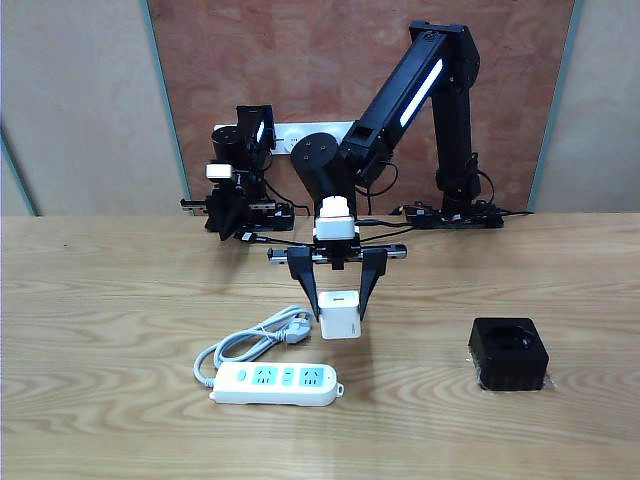}{Transport toward the socket} &
\ComparisonImage{26mm}{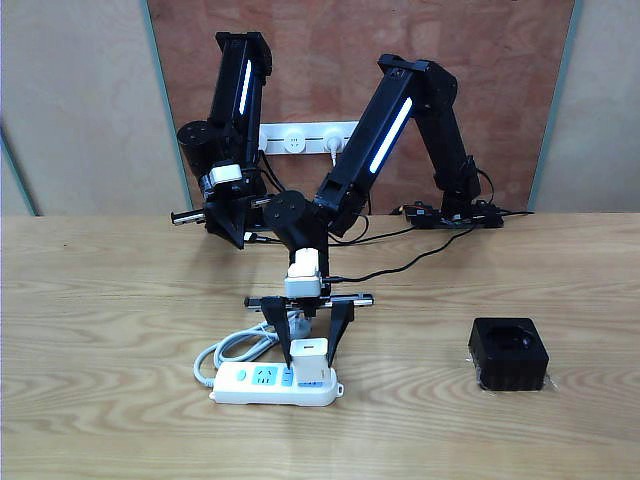}{Held near the target socket}\\
\end{tabularx}
\par\vspace{8pt}
\PLPanelHead{b}{L1-04}{Cup Nesting}{Observed outcome: cups toppled after release}{\PLTally{L1-04}}
\vspace{4pt}
\begin{tabularx}{\linewidth}{@{}X@{\hspace{1.5mm}\PLArrow{PLSim}{12.5mm}\hspace{1.5mm}}X@{\PLGutter}X@{\hspace{1.5mm}\PLArrow{PLReal}{12.5mm}\hspace{1.5mm}}X@{}}
\ComparisonImage{26mm}{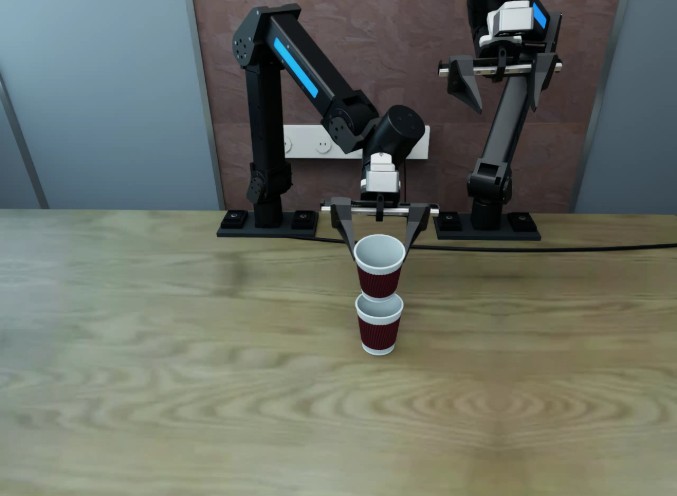}{Nesting attempt} &
\ComparisonImage{26mm}{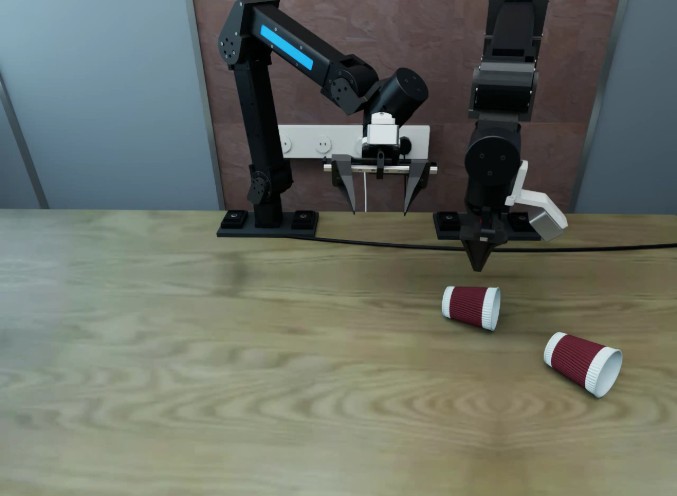}{Cups toppled} &
\ComparisonImage{26mm}{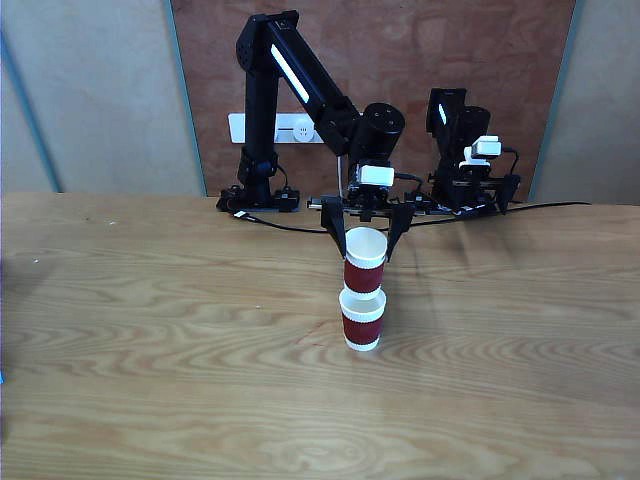}{Cup above the target cup} &
\ComparisonImage{26mm}{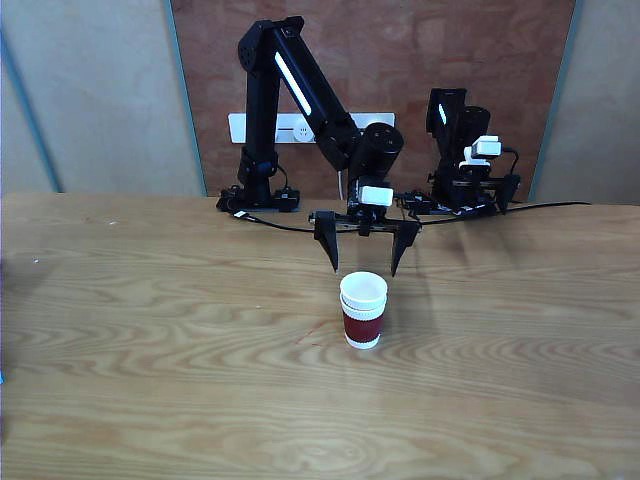}{Cups nested after release}\\
\end{tabularx}
\par\vspace{8pt}
\PLPanelHead{c}{L2-02}{Book Pulling}{\PLBottleneckForce}{\PLTally{L2-02}}
\vspace{3pt}
\FigEightHalf{\FigEightConfigurations{PLSim}}\hfill\FigEightHalf{\FigEightConfigurations{PLReal}}
\par\vspace{2pt}
\begin{tabularx}{\linewidth}{@{}X@{\hspace{3mm}}X@{\PLGutter}X@{\hspace{3mm}}X@{}}
\ComparisonImage{26mm}{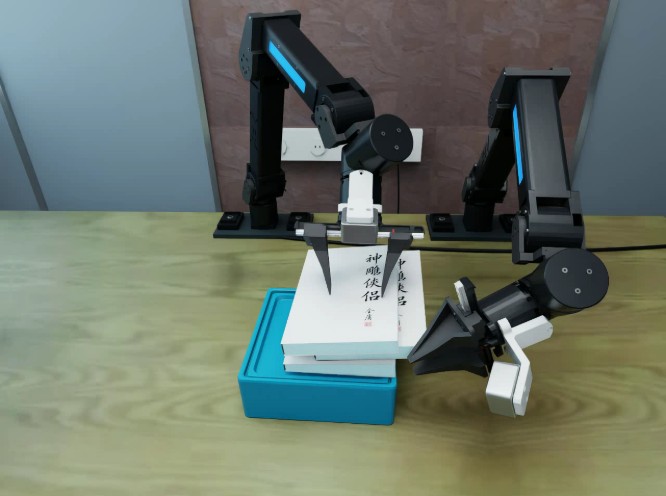}{Side grasp and top support} &
\ComparisonImage{26mm}{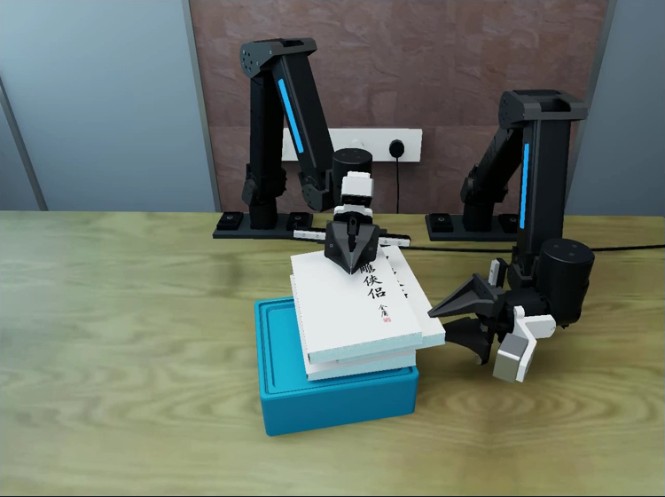}{Side grasp and top support} &
\ComparisonImage{26mm}{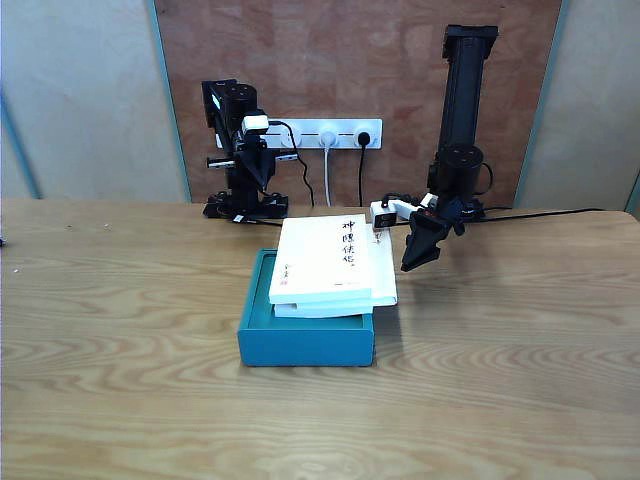}{Repositioned grasp attempt} &
\ComparisonImage{26mm}{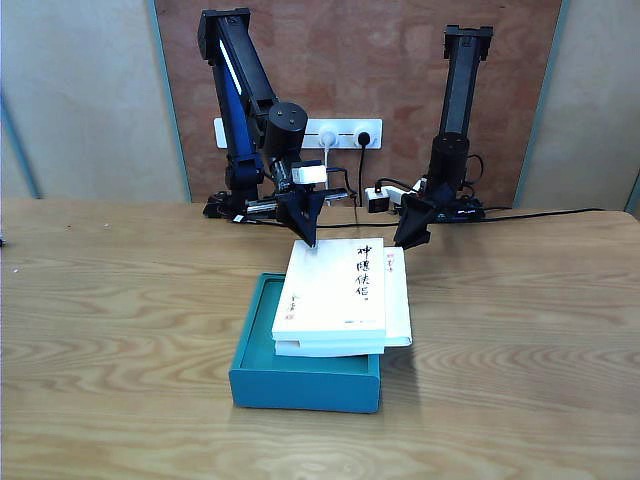}{Initial grasp approach}\\
\end{tabularx}
\par\vspace{8pt}
\PLPanelHead{d}{L2-05}{Drawer Pull \& Return}{\PLBottleneckForce}{\PLTally{L2-05}}
\vspace{4pt}
\begin{tabularx}{\linewidth}{@{}X@{\hspace{1.5mm}\PLArrow{PLSim}{12.5mm}\hspace{1.5mm}}X@{\PLGutter}X@{\hspace{1.5mm}\PLArrow{PLReal}{12.5mm}\hspace{1.5mm}}X@{}}
\ComparisonImage{26mm}{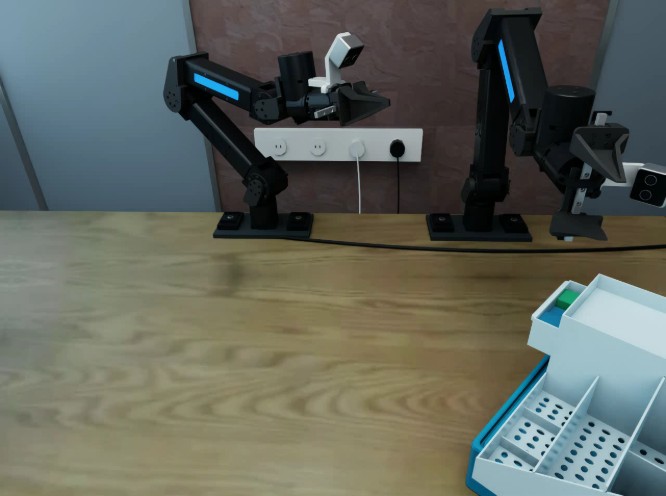}{Drawer partly open} &
\ComparisonImage{26mm}{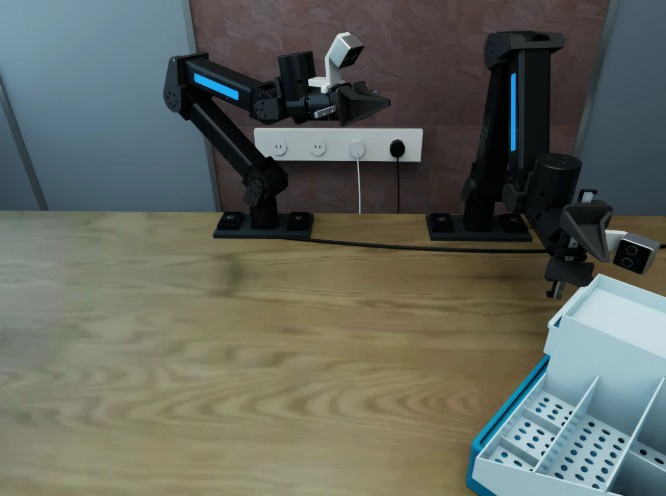}{Drawer moved inward} &
\ComparisonImage{26mm}{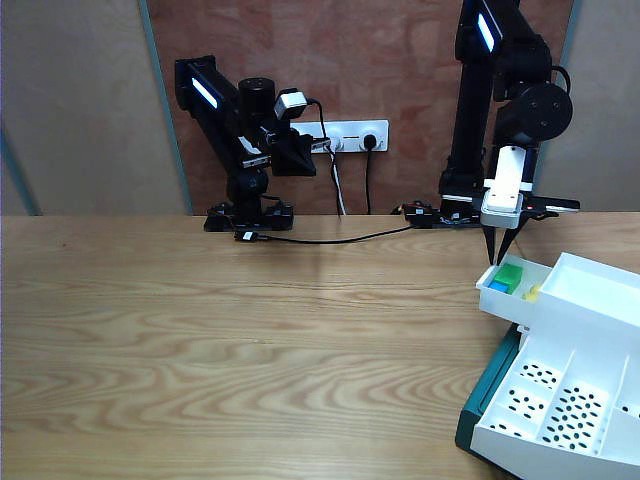}{Drawer extended} &
\ComparisonImage{26mm}{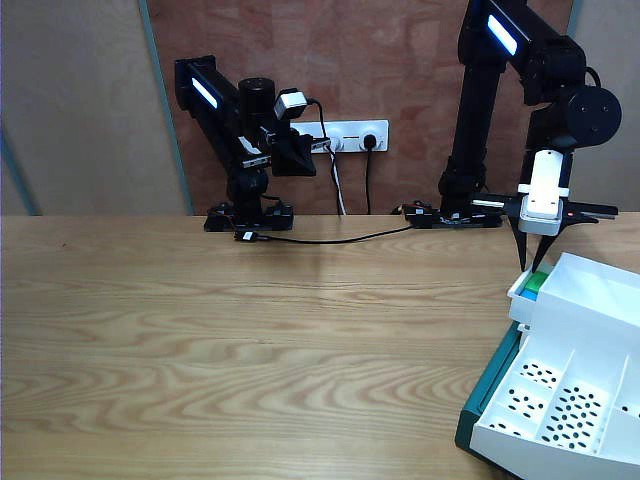}{Drawer moved inward}\\
\end{tabularx}
\caption{\textbf{Comparing local interaction progress and task outcomes across real and simulated environments.} Panel (a) concerns assessment granularity: the same alignment criterion was judged visually in the real world and checked with a fixed geometric tolerance in simulation. The unsuccessful simulated trial exhausted its model-call budget during repeated corrections. Panels (b--d) concern final-state stability after release, grasp selection with bimanual force regulation, and local drawer motion and force regulation. Headers follow Figure~\ref{fig:sim-real-shared-difficulties}, except that panel (b) reports an observed outcome without assigning a specific interaction bottleneck.}
\label{fig:sim-real-outcome-differences}
\end{figure*}
\arrayrulecolor{black}
\endgroup

%% file: figures/figure07_initial_conditions.tex
\begingroup
\input{figures/comparison_figure_style}
\begin{figure*}[!t]
\centering
\PLPanelHead{a}{L2}{Two blocks near the dustpan entrance}{Local sweeping under a relatively stable contact mode}{\PLEnvNote{real}}
\vspace{4pt}
\begin{tabularx}{\linewidth}{@{}X@{\hspace{1.5mm}\PLArrow{PLReal}{15mm}\hspace{1.5mm}}X@{\hspace{1.5mm}\PLArrow{PLReal}{15mm}\hspace{1.5mm}}X@{}}
\ComparisonImage{32mm}{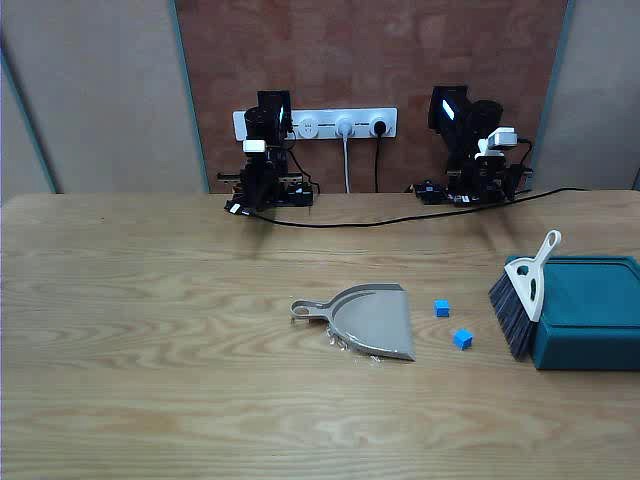}{Initial state} &
\ComparisonImage{32mm}{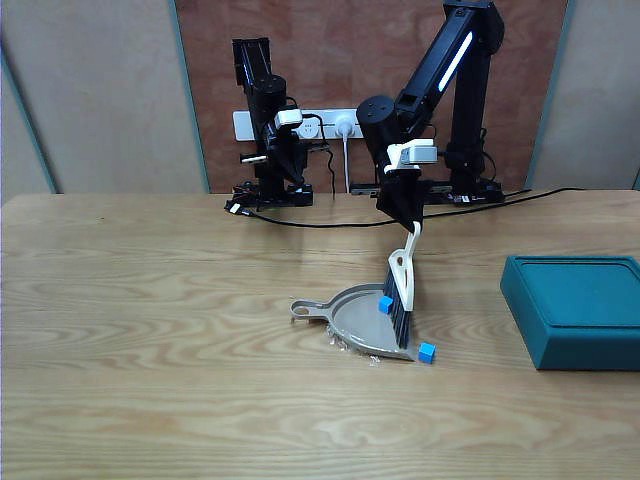}{One block inside; one outside} &
\ComparisonImage{32mm}{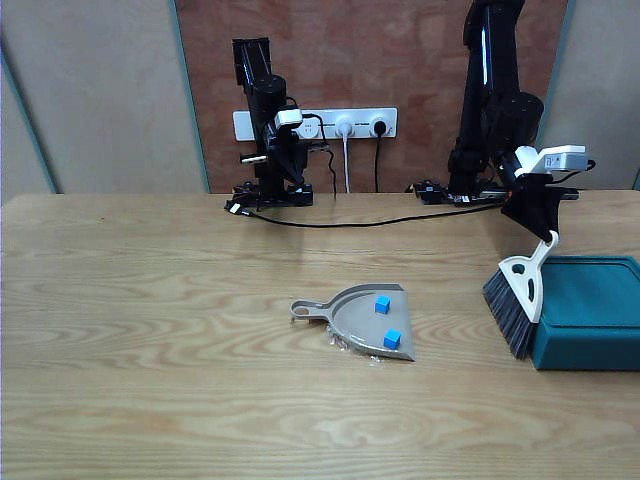}{Both inside; brush withdrawn}\\
\end{tabularx}
\ComparisonRule
\PLPanelHead{b}{L3}{Three dispersed blocks}{Successive contact reconfiguration and force regulation}{\PLEnvNote{real}}
\vspace{4pt}
\begin{tabularx}{\linewidth}{@{}X@{\hspace{1.5mm}\PLArrow{PLReal}{15mm}\hspace{1.5mm}}X@{\hspace{1.5mm}\PLArrow{PLReal}{15mm}\hspace{1.5mm}}X@{}}
\ComparisonImage{32mm}{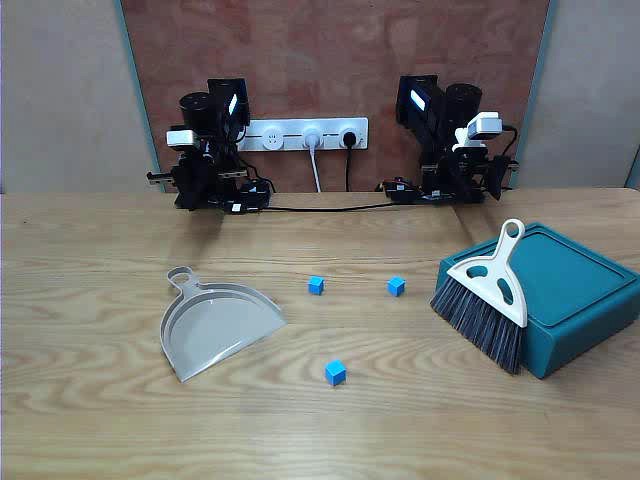}{Initial state} &
\ComparisonImage{32mm}{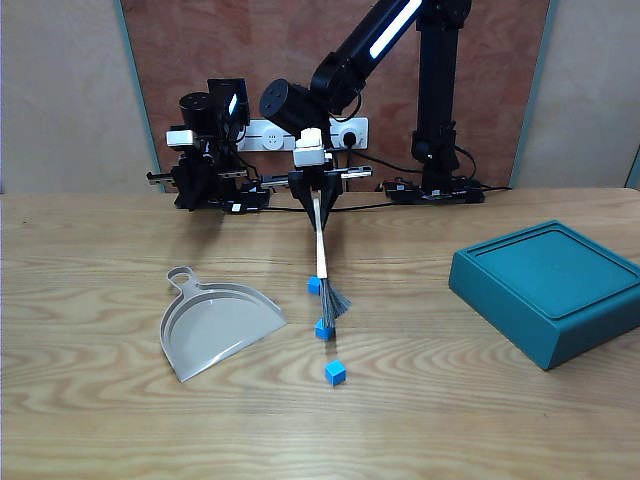}{Brush repositioned near blocks} &
\ComparisonImage{32mm}{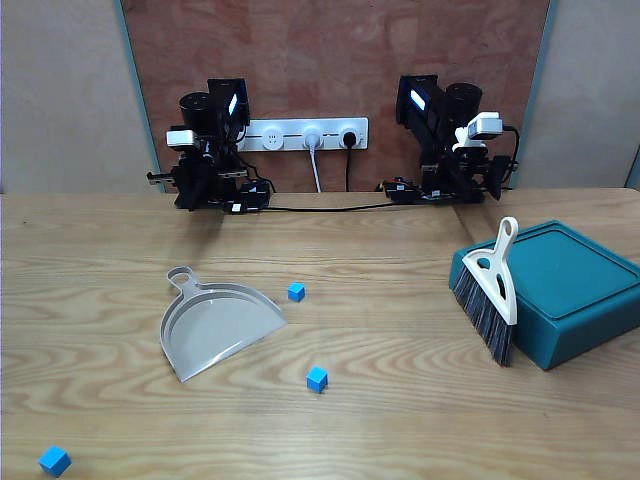}{Blocks remain outside}\\
\end{tabularx}
\caption{\textbf{Initial configurations change the interaction demands of Block Sweeping.} (a) Two blocks near the dustpan entrance are collected through local sweeping under a relatively stable contact mode, consistent with L2. (b) Three dispersed blocks require successive contact reconfiguration and force regulation, consistent with L3; the recorded execution remains incomplete. Each row shows initial, intermediate, and final states. These variants differ in block count, layout, and dustpan orientation; they are not a controlled layout ablation or extra trials in Table~\ref{tab:level-performance}.}
\label{fig:initial-conditions}
\end{figure*}
\arrayrulecolor{black}
\endgroup

%% file: figures/figure10_additional_failure_cases.tex
\begingroup
\input{figures/comparison_figure_style}
\makeatletter
\def\PL@dagger#1\PL@enddagger{}
\makeatother
\newcommand{\FigTenCase}[5]{%
  \ifstrequal{#1}{sim}{\colorlet{FigTenEnv}{PLSim}}{\colorlet{FigTenEnv}{PLReal}}%
  \begin{tabularx}{\linewidth}{@{}X@{\hspace{1.5mm}\PLArrow{FigTenEnv}{12mm}\hspace{1.5mm}}X@{}}
  \ComparisonImage{29mm}{#2}{#3} & \ComparisonImage{29mm}{#4}{#5}\\
  \end{tabularx}}
\begin{figure*}[!t]
\centering
\begin{minipage}[t]{0.485\linewidth}
\PLPanelHead{a}{L2-04}{Orange Handling}{\PLBottleneckContact\hfill\PLEnvNote{sim}}{\PLTally{L2-04}}
\vspace{3pt}
\FigTenCase{sim}{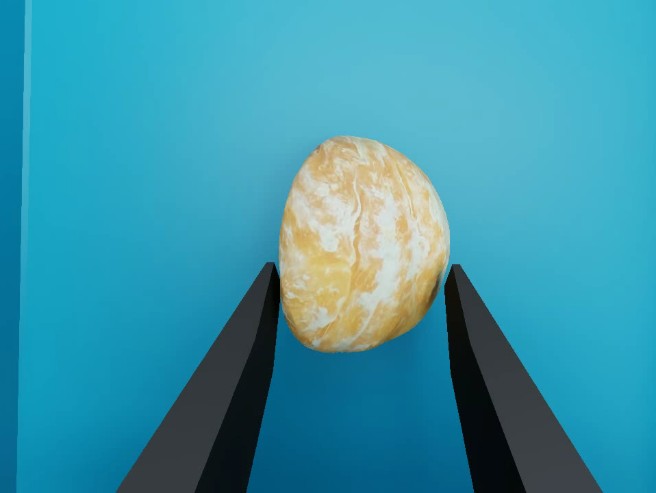}{Approaching the upright segment}{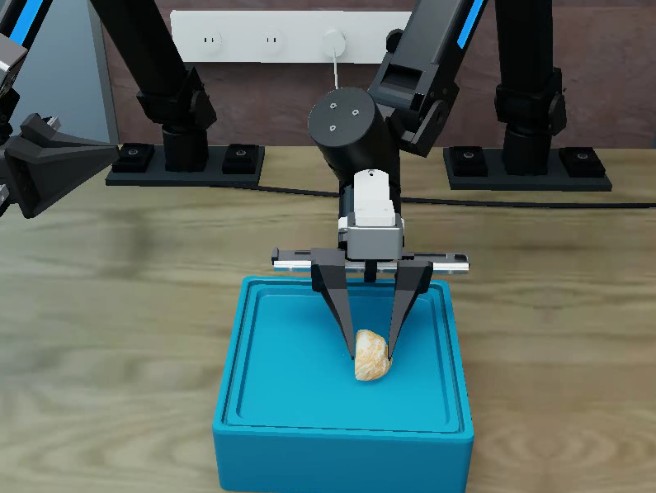}{Segment remains on support}
\end{minipage}\hfill
\begin{minipage}[t]{0.485\linewidth}
\PLPanelHead{b}{L2-07}{\parbox[t]{100pt}{\raggedright\strut Whiteboard Wiping\\\strut}}{\PLBottleneckForce\hfill\PLEnvNote{real}}{\PLTally{L2-07}}
\vspace{3pt}
\FigTenCase{real}{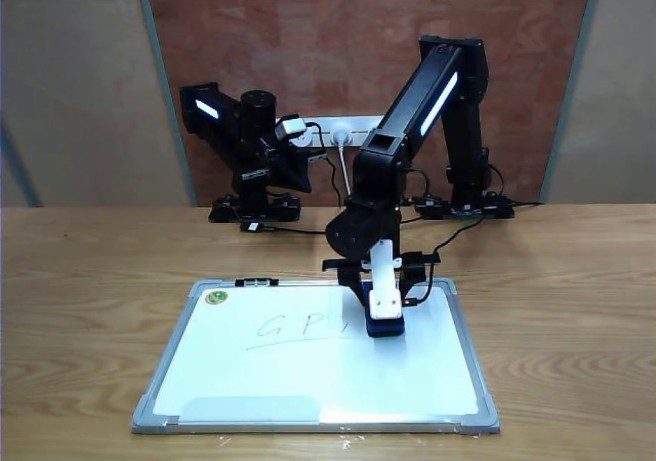}{Wiping attempt}{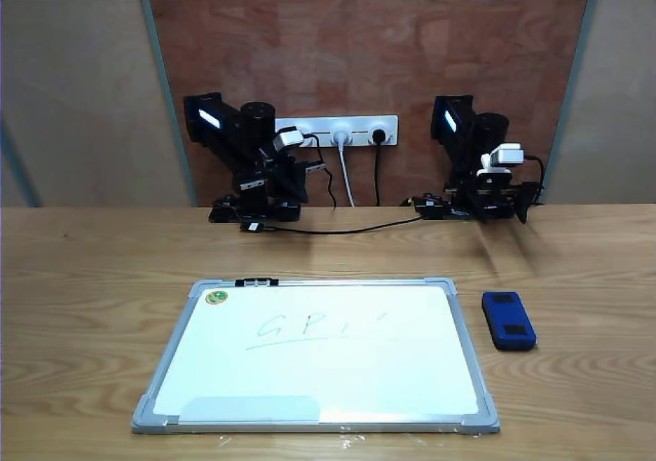}{Residual writing}
\end{minipage}
\par\vspace{10pt}
\begin{minipage}[t]{0.485\linewidth}
\PLPanelHead{c}{L3-02}{Nut Unscrewing}{\PLBottleneckContact\hfill\PLEnvNote{sim}}{\PLTally{L3-02}}
\vspace{3pt}
\FigTenCase{sim}{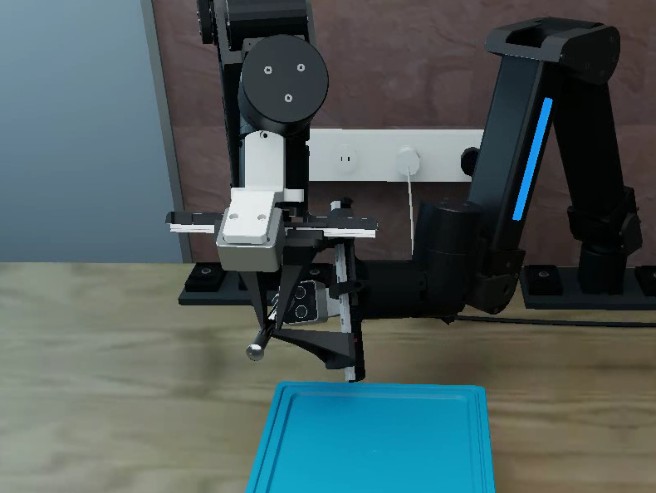}{Bimanual contact with the assembly}{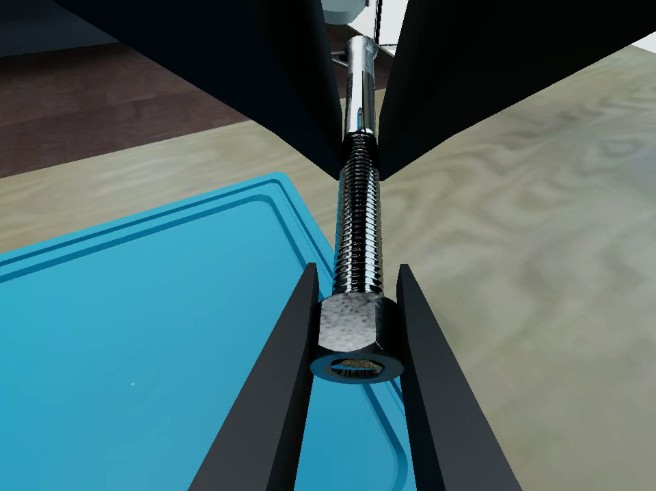}{Nut remains on the threaded rod}
\end{minipage}\hfill
\begin{minipage}[t]{0.485\linewidth}
\PLPanelHead{d}{L3-03}{\parbox[t]{100pt}{\raggedright\strut Stacked-Book\\\strut Pushing \& Rotation}}{\PLBottleneckCoordination\hfill\PLEnvNote{sim}}{\PLTally{L3-03}}
\vspace{3pt}
\FigTenCase{sim}{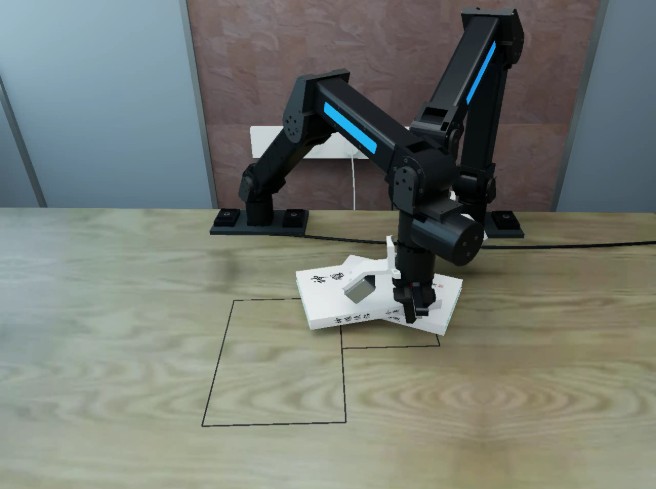}{Bimanual pushing and partial rotation}{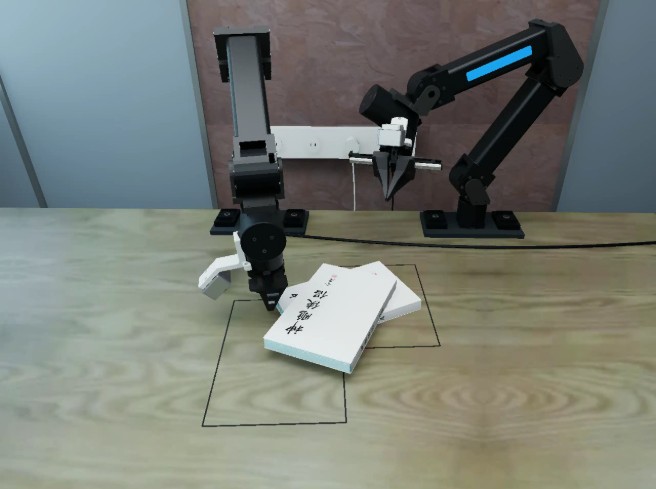}{Stack remains partly outside target}
\end{minipage}
\caption{\textbf{Additional failure cases.} (a) Grasping attempts leave an upright orange segment on its support. (b) Wiping leaves visible writing on the board. (c) The nut remains on the threaded rod despite bimanual contact; the execution record reports infeasible subsequent turning poses. (d) Supported pushing produces partial translation and rotation without completing the full-turn transfer. These cases complement Figures~\ref{fig:sim-real-shared-difficulties} and~\ref{fig:sim-real-outcome-differences}. SIM/REAL labels report successes out of three trials per task (Table~\ref{tab:level-performance}).}
\label{fig:additional-failure-cases}
\end{figure*}
\arrayrulecolor{black}
\endgroup

%% file: sec/7_discussion_and_limitations.tex
\par
\Needspace{4\baselineskip}
\subsection{\revision{Limitations and Scope}}
\label{sec:limitations-scope}
\label{sec:discussion}
The evaluation covers 16 household tasks with three trials per task in each environment. Limited real-robot resources and low parallelization constrain trial counts and statistical precision. The Sim--Real correspondence is therefore descriptive and specific to GPT-6 Astra; whether simulation predicts real-world performance for other models remains to be tested. Coverage is limited to the tested task versions, initial configurations, and robot platform. Our analysis characterizes interaction failures in this system; contributions of model decisions, sensing, and robot execution remain to be isolated.
\par

%% file: sec/8_conclusion.tex
\par
\Needspace{4\baselineskip}
\section{Conclusion}
\label{sec:conclusion}
We evaluate GPT-6 Astra's zero-shot contact-rich manipulation capabilities on the 16 household tasks of LLM4Robot-Bench in matched real and simulated environments. Astra demonstrates strong performance on basic pick-and-place tasks requiring semantic understanding and object localization, but performance declines as physical interaction demands increase, with no complete successes on the tested L3 tasks. Difficulties in complex spatial reasoning, contact configuration, force regulation, and their coordination limit individual actions and sustained progress across successive interactions. These findings reveal a gap between effective local actions and reliable task completion, identifying sustained physical interaction as a priority for general-purpose embodied models.
\par
For GPT-6 Astra, the observed correspondence in task outcomes and interaction bottlenecks indicates that physically and visually matched simulation can reflect key capabilities and limitations revealed by real-world evaluation. LLM4Robot-Bench's L1--L3 hierarchy characterizes capability boundaries across increasing demands on contact configuration, force regulation, and their coordination. Explicit task specifications and controlled initialization support repeatable comparisons, while parallel execution reduces dependence on real-robot resources. This provides a standardized, scalable benchmark for evaluating progress toward reliable physical interaction, a prerequisite for general physical intelligence.
\par

%% file: sec/supplement_format.tex
\setcounter{table}{0}
\renewcommand{\thetable}{S\arabic{table}}
\providecommand{\theHtable}{}
\renewcommand{\theHtable}{supplement.table.\arabic{table}}
\setcounter{figure}{0}
\renewcommand{\thefigure}{S\arabic{figure}}
\providecommand{\theHfigure}{}
\renewcommand{\theHfigure}{supplement.figure.\arabic{figure}}
\definecolor{InteractionLevelL1}{HTML}{087F8C}
\definecolor{InteractionLevelL2}{HTML}{BD5234}
\definecolor{InteractionLevelL3}{HTML}{20466E}
\newcommand{\TaskSpecField}[1]{\par\noindent\textbf{#1}\enspace}

%% file: sec/A_task_specifications.tex
\section{Detailed Task Specifications}
\label{app:task-specifications}
Each task card lists the initial conditions, goal and constraints, task instruction, criterion for complete success, and the contact requirements that motivate the task's interaction level. Table~\ref{tab:level-assignment} summarizes how these requirements determine each level under the criteria in Section~\ref{sec:contact-hierarchy}. The quoted task instruction is retained unchanged; the goal description and success criteria operationalize that instruction without adding unrequested completion steps. Instructions remain fixed within each episode.

\begin{table}[!htbp]
\centering
\caption{\textbf{Interaction-level assignment of the 16 tasks.} Requirements follow the definitions in Section~\ref{sec:contact-hierarchy} (Table~\ref{tab:level-criteria}). \emph{No additional requirement} means no force-regulation requirement beyond ordinary support, transport, and geometric placement. Natural rolling, sliding, and content release do not by themselves constitute contact reconfiguration.}
\label{tab:level-assignment}
\footnotesize
\setlength{\tabcolsep}{3pt}
\renewcommand{\arraystretch}{1.1}
\begin{tabularx}{\linewidth}{@{}>{\raggedright\arraybackslash}p{0.235\linewidth}>{\raggedright\arraybackslash}p{0.19\linewidth}>{\raggedright\arraybackslash}X>{\raggedright\arraybackslash}p{0.205\linewidth}@{}}
\toprule
\textbf{Task} & \textbf{Contact configuration} & \textbf{Force-regulation requirements} & \textbf{Contact reconfiguration} \\
\midrule
\multicolumn{4}{@{}l}{\textit{L1: contact selection and geometric alignment}} \\
L1-01 Marker Placement & Grasp leaves one end free for placement & No additional requirement & Not required \\
L1-02 Charger Alignment & Gripper--charger; prongs aligned, not inserted & No additional requirement & Not required \\
L1-03 Hat Hanging & Grasp leaves the head opening accessible & No additional requirement & Not required \\
L1-04 Cup Nesting & Sequential grasps of the two cups & No additional requirement & Placement-only sequential grasps \\
L1-05 Flower Transfer & Grasp leaves sufficient stem length free & No additional requirement & Not required \\
\midrule
\multicolumn{4}{@{}l}{\textit{L2: force regulation without required contact reconfiguration}} \\
L2-01 Marker Cap Removal & Grippers--body and cap & Pull exceeds cap retention without slip, along the marker axis & Not required \\
L2-02 Book Pulling & Grasp middle book; stabilize upper book & Pull overcomes friction between books without displacing the stack & Not required \\
L2-03 Marble Pouring & Gripper--source cup; source cup--marbles & Regulate the marbles' motion and release into the receiver through cup orientation and motion & Not required during pouring \\
L2-04 Orange Handling & Gripper--segment & Maintain support while limiting compression to avoid damage throughout handling & Not required \\
L2-05 Drawer Pull \& Return & Gripper--handle & Pull and push overcome guide friction, along the guide, without moving the cabinet & Not required; one grasp pulls and pushes \\
L2-06 Jelly Handling & Gripper--jelly & Maintain support while limiting compression to avoid damage throughout handling & Not required \\
L2-07 Whiteboard Wiping & Gripper--eraser; eraser--board & Pressing force sufficient to erase while wiping & Not required; the contact slides \\
\midrule
\multicolumn{4}{@{}l}{\textit{L3: force regulation with contact reconfiguration}} \\
L3-01 Bottle Cap Unscrewing & Gripper--cap; cap--bottle thread; bottle--fixture & Apply unscrewing torque while the bottle remains fixed & Required; the rotation exceeds one grasp's turning range \\
L3-02 Nut Unscrewing & Grippers--nut and bolt body; nut--bolt thread & Apply unscrewing torque while the other arm resists the reaction & Required; the rotation exceeds one grasp's turning range \\
L3-03 Stacked-Book Pushing \& Rotation & Grippers--stack; stack--table & Push force and turning torque that keep the stack together, without lifting & Required; pushing contacts must change during the full turn \\
L3-04 Block Sweeping & Gripper--brush; brush--blocks & Control block motion toward the dustpan through brush contact & Required; separated blocks need successive brush contacts \\
\bottomrule
\end{tabularx}
\end{table}

\begingroup
\definecolor{TaskLevelOne}{HTML}{087F8C}
\definecolor{TaskLevelTwo}{HTML}{BD5234}
\definecolor{TaskLevelThree}{HTML}{20466E}
\colorlet{TaskLevelColor}{TaskLevelOne}
\titleformat*{\subsection}{\large\bfseries\color{TaskLevelColor}}
\titleformat{\paragraph}[block]
  {\normalfont\normalsize\bfseries\color{TaskLevelColor}}{}{0pt}{\underline}
\titlespacing*{\paragraph}{0pt}{0pt}{0.35\baselineskip}

\Needspace{7\baselineskip}
\subsection{L1 Task Specifications}

\Needspace{5\baselineskip}

\paragraph{L1-01: Marker Placement.}
\label{app:task-L1-01}
\TaskSpecField{Initial conditions.} A capped marker, approximately 150 mm long, stands upright to the right of a fixed, upright ceramic cup. The cup is approximately 85 mm high with a 70.8 mm inner opening; the object centers are approximately 300 mm apart. Both grippers start empty.
\TaskSpecField{Goal and constraints.} The left arm grasps the marker and places either end inside the cup without removing its cap, then releases it. The right arm observes.
\TaskSpecField{Task instruction.} \textit{``Pick up the thick marker with the left arm and place it in the wide-mouth cup, then release it.''}
\TaskSpecField{Complete success.} \looseness=-1 After release and gripper withdrawal, the marker remains inside the cup and is supported by it. This wide-mouth placement task does not require a close-fitting insertion or a specified insertion depth.
\TaskSpecField{Contact requirements.} A stable grasp and geometric alignment of the marker with the cup opening suffice; no additional force regulation is required, which motivates the L1 assignment.
\par\medskip

\Needspace{5\baselineskip}

\paragraph{L1-02: Charger Alignment.}
\label{app:task-L1-02}
\TaskSpecField{Initial conditions.} The charger stands upright in a freely resting grooved support, with its prongs pointing downward and its upper housing exposed. A power strip is fixed separately to its left, with the socket openings facing upward; their centers are approximately 330 mm apart. Both grippers start empty.
\TaskSpecField{Goal and constraints.} The left arm lifts the charger clear of its support and aligns its two prongs with the corresponding holes of the designated socket nearest the left arm. The right arm observes.
\TaskSpecField{Task instruction.} \textit{``Pick up the power adapter with the left arm and align its prongs with the socket holes closest to the left arm.''}
\TaskSpecField{Complete success.} The charger is held near the designated socket with both prongs aligned with their respective holes. Insertion, release, and electrical operation are not required.
\TaskSpecField{Contact requirements.} Precise alignment of both prongs with the socket holes is the main demand; insertion and additional force regulation are not required, which motivates the L1 assignment.
\par\medskip

\Needspace{5\baselineskip}

\paragraph{L1-03: Hat Hanging.}
\label{app:task-L1-03}
\TaskSpecField{Initial conditions.} A yellow open-top visor rests freely on a raised box beside a fixed, approximately 740 mm-high coat rack. The lowest slanted branch faces the left arm, and the visor rim is accessible for grasping. Both grippers start empty.
\TaskSpecField{Goal and constraints.} The left arm picks up the visor, passes its head opening over the rounded end of the lowest branch, and releases it so that the band is supported by that branch. The right arm observes.
\TaskSpecField{Task instruction.} \textit{``Use the left arm to hang the hat on the lowest rod through its head opening.''}
\TaskSpecField{Complete success.} The released visor is supported through its head opening by the designated branch, independently of the gripper. Perching on the branch tip or using another branch is insufficient. Natural oscillation is permitted; no additional stability dwell is required.
\TaskSpecField{Contact requirements.} The grasp must hold the hat securely while leaving its head opening accessible for alignment with the branch; no additional force regulation is required, which motivates the L1 assignment.
\par\medskip

\Needspace{5\baselineskip}

\paragraph{L1-04: Cup Nesting.}
\label{app:task-L1-04}
\TaskSpecField{Initial conditions.} Two identical empty paper cups stand upright and unfixed, one at the left rear and the other at the right front of the workspace. Each is approximately 80 mm high with a 70 mm outer rim diameter. Both grippers start empty.
\TaskSpecField{Goal and constraints.} The left arm places the left cup at the central location and releases it. The right arm then places the right cup inside it and releases it; the arms operate sequentially.
\TaskSpecField{Task instruction.} \textit{``Use the left arm to place the left cup at the center and release it, then use the right arm to nest the right cup inside it and release it.''}
\TaskSpecField{Complete success.} The upper cup is actually nested in and independently supported by the lower cup after release and withdrawal, with stable nesting. No specified yaw orientation or strict centering is required. An open gripper that still traps the cup does not constitute release.
\TaskSpecField{Contact requirements.} Sequential stable grasps and alignment of the upper cup with the lower cup suffice; no additional force regulation is required, which motivates the L1 assignment.
\par\medskip

\Needspace{5\baselineskip}

\paragraph{L1-05: Flower Transfer.}
\label{app:task-L1-05}
\TaskSpecField{Initial conditions.} A white artificial flower rests in the left paper cup, which also contains a charger. The right paper cup, containing glass balls and pearls, is the receiving container. Both grippers start empty.
\TaskSpecField{Goal and constraints.} The left arm retrieves the flower from the left cup, inserts its stem into the right cup, and releases it. The grasp must leave sufficient stem length free for insertion. The right arm observes.
\TaskSpecField{Task instruction.} \textit{``Pick up the flower from the left cup, insert its stem into the right cup, and release it.''}
\TaskSpecField{Complete success.} The flower and receiving cup remain independently stable after release and gripper withdrawal, with the stem inside the receiving cup.
\TaskSpecField{Contact requirements.} A grasp that leaves enough stem free and alignment of the stem with the receiving cup suffice; no additional force regulation is required, which motivates the L1 assignment.
\par\medskip

\colorlet{TaskLevelColor}{TaskLevelTwo}
\Needspace{7\baselineskip}
\subsection{L2 Task Specifications}

\Needspace{5\baselineskip}

\paragraph{L2-01: Marker Cap Removal.}
\label{app:task-L2-01}
\TaskSpecField{Initial conditions.} An approximately 150 mm-long capped marker lies horizontally on a raised box between the arms, with the body to the right and cap to the left. It is not fixed, and both grippers start empty.
\TaskSpecField{Goal and constraints.} The right arm picks up and stabilizes the marker body. The left arm grasps the cap and pulls it away along the marker axis, maintaining both grasps until separation. The task ends at complete separation; placing down or releasing either component is not required.
\TaskSpecField{Task instruction.} \textit{``Pick up the black pen with the right arm, then grasp the cap with the left arm and pull it off along the pen's axis.''}
\TaskSpecField{Complete success.} The cap fully separates from the marker body after the instructed bimanual pull. Continued holding is sufficient; placement, release, and arm return are not required.
\TaskSpecField{Contact requirements.} Both grasps must be maintained while the left arm overcomes the cap's resistance to separation along the marker axis, which motivates the L2 assignment.
\par\medskip

\Needspace{5\baselineskip}

\paragraph{L2-02: Book Pulling.}
\label{app:task-L2-02}
\TaskSpecField{Initial conditions.} Three approximately 220 by 140 by 20 mm books are stacked on a support box. In the reference camera view, the middle book protrudes 40 mm to the right, with 5 mm clearance above it. The exposed side is accessible to the left gripper, and both grippers start empty.
\TaskSpecField{Goal and constraints.} The right arm stabilizes the upper book while the left arm grasps the middle book and pulls it 5 cm outward from its initial position. Complete removal from the stack is not required.
\TaskSpecField{Task instruction.} \textit{``Hold the top book steady with the right arm and pull the middle book out 5 cm with the left arm.''}
\TaskSpecField{Complete success.} The middle book moves the required 5 cm while the upper book remains stabilized. Motion of the entire stack or accumulated end-effector commands does not establish this displacement.
\TaskSpecField{Contact requirements.} The left grasp must pull the middle book out of the stack while the right arm keeps the upper book stable; regulating the pulling force within sustained contact, without displacing the stack, motivates the L2 assignment.
\par\medskip

\Needspace{5\baselineskip}

\paragraph{L2-03: Marble Pouring.}
\label{app:task-L2-03}
\TaskSpecField{Initial conditions.} Two upright paper cups, approximately 80 mm high with a 70 mm outer rim diameter, are placed approximately 200 mm apart. The left source cup contains twelve freely rolling glass marbles; it is not fixed to the table. The right receiving cup is fixed at its initial position. Both grippers start empty.
\TaskSpecField{Goal and constraints.} The left arm grasps the source cup, pours into the fixed receiving cup until the source is empty, then returns the source cup to its original position and releases it. The right arm only assists observation.
\TaskSpecField{Task instruction.} \textit{``Pick up the left cup with the left arm, pour its contents into the right cup until empty, then place it back and release it.''}
\TaskSpecField{Complete success.} The source cup is empty, at least six of the twelve marbles remain in the receiving cup, and the receiving cup remains at its initial fixed position. The source cup is then returned to its original position and released. Cup tilt alone is insufficient.
\TaskSpecField{Contact requirements.} \looseness=-1 Cup orientation and motion change how the cup supports and guides the marbles, controlling their motion relative to the cup and their velocity at release. Regulating this motion and release falls within the task-level definition of force regulation (Section~\ref{sec:contact-hierarchy}). The robot can maintain its grasp without re-establishing contacts to continue pouring, motivating the L2 assignment.
\par\medskip

\Needspace{5\baselineskip}

\paragraph{L2-04: Orange Handling.}
\label{app:task-L2-04}
\TaskSpecField{Initial conditions.} A peeled orange segment lies freely on a fixed support box between the arms, with its broad flesh side upward and its long axis along the camera-to-robot direction. Both grippers start empty.
\TaskSpecField{Goal and constraints.} \looseness=-1 The left arm lifts the segment 5--20 cm above its initial position, returns it to the same support, and releases it without damage throughout the operation. The right arm observes.
\TaskSpecField{Task instruction.} \textit{``Use the left arm to lift the orange segment 5--20 cm without damaging it,  then place it back on the box and release it.''}
\TaskSpecField{Complete success.} The specified lift and return are completed, and the segment remains independently supported after release, with the no-damage constraint satisfied: no crushing is visible in the real world, and the simulated segment shows no structural damage.
\TaskSpecField{Contact requirements.} A sustained grasp must hold the deformable segment firmly enough to lift it without damaging it, which motivates the L2 assignment.
\par\medskip

\Needspace{5\baselineskip}

\paragraph{L2-05: Drawer Pull \& Return.}
\label{app:task-L2-05}
\TaskSpecField{Initial conditions.} The upper drawer of the storage cabinet is partly open, with its initial opening recorded relative to the cabinet. Both grippers start empty.
\TaskSpecField{Goal and constraints.} The left arm establishes a grasp, pulls the drawer approximately 5 cm farther out along its guide, and pushes it back to the recorded initial opening. The right arm observes. Moving the cabinet as a whole does not satisfy the task.
\TaskSpecField{Task instruction.} \textit{``Use the left arm to pull the partially open top drawer out another 5 cm, then push it back to its original position.''}
\TaskSpecField{Complete success.} The drawer first moves outward by the prescribed amount and then returns to its initial opening. Merely ending at the initial opening without the required outward movement, or remaining open after a pull, is insufficient.
\TaskSpecField{Contact requirements.} A sustained grasp must transmit pulling and pushing force along the drawer guide without moving the cabinet, which motivates the L2 assignment.
\par\medskip

\Needspace{5\baselineskip}

\paragraph{L2-06: Jelly Handling.}
\label{app:task-L2-06}
\TaskSpecField{Initial conditions.} An unpackaged jelly rests freely on the green box. Both grippers start empty.
\TaskSpecField{Goal and constraints.} The left arm lifts the jelly 5--20 cm above its initial position, returns it to the same box, and releases it without crushing it throughout the operation. The right arm observes.
\TaskSpecField{Task instruction.} \looseness=-1 \textit{``Pick up the jelly from the green box with the left arm, lift it 5--20 cm, then place it back and release it.''}
\TaskSpecField{Complete success.} The specified lift and return are completed, and the jelly remains independently supported on its original box after release, with the no-damage constraint satisfied throughout: no crushing is visible in the real world, and the simulated jelly shows no structural damage.
\TaskSpecField{Contact requirements.} A sustained grasp must hold the deformable jelly firmly enough to lift and return it without damaging it, which motivates the L2 assignment.
\par\medskip

\Needspace{5\baselineskip}

\paragraph{L2-07: Whiteboard Wiping.}
\label{app:task-L2-07}
\TaskSpecField{Initial conditions.} A marked whiteboard and a blue rigid eraser are presented separately. The eraser rests on the table, and both grippers start empty.
\TaskSpecField{Goal and constraints.} The left arm acquires the eraser and removes all writing with its working surface. The right arm observes. The goal concerns the board's visible state, not traversal of a nominal path.
\TaskSpecField{Task instruction.} \textit{``Pick up the board eraser with the left arm and erase the writing on the whiteboard.''}
\TaskSpecField{Complete success.} All writing on the board has been erased. Performing a contact wiping motion without removing the writing is insufficient.
\TaskSpecField{Contact requirements.} Sustained contact between the eraser surface and the board suffices for the operation. Coordinating pressing force with wiping motion to remove the writing motivates the L2 assignment.
\par\medskip

\colorlet{TaskLevelColor}{TaskLevelThree}
\Needspace{7\baselineskip}
\subsection{L3 Task Specifications}

\Needspace{5\baselineskip}

\paragraph{L3-01: Bottle Cap Unscrewing.}
\label{app:task-L3-01}
\TaskSpecField{Initial conditions.} \looseness=-1 The bottle body is fixed to the tabletop with its cap engaged. Both grippers start empty.
\TaskSpecField{Goal and constraints.} The right arm grasps and unscrews the cap, then lifts it 3--5 cm away from the bottle opening. The task instruction assigns cap manipulation to the right arm. In simulation, both arms are enabled for manipulation, so this arm-role constraint is specified by the instruction and is not enforced by the action interface. The fixture holds the bottle body throughout the operation.
\TaskSpecField{Task instruction.} \textit{``The bottle body is fixed to the table. Unscrew the cap with the right arm, then lift the cap 3--5 cm away from the bottle opening.''}
\TaskSpecField{Complete success.} The cap fully disengages from the thread and is separated by 3--5 cm from the bottle opening while the bottle remains fixed. Grasping the cap, issuing a wrist rotation, or partially loosening the thread is not complete success.
\TaskSpecField{Contact requirements.} Regrasping is required when the remaining screw travel exceeds the turning range of one grasp. Coordinating these grasp changes with effective torque transmission against the fixed bottle motivates the L3 assignment.
\par\medskip

\Needspace{5\baselineskip}

\paragraph{L3-02: Nut Unscrewing.}
\label{app:task-L3-02}
\TaskSpecField{Initial conditions.} An expansion-bolt assembly, with an approximately 120 mm-long rod, lies horizontally on a support box, with the body to the right and nut/threaded end to the left. The assembly is unfixed, the nut is pre-loosened by three turns, and both grippers start empty.
\TaskSpecField{Goal and constraints.} The right arm picks up and stabilizes the body; the left arm grasps the hexagonal nut and unscrews it counterclockwise until it completely separates from the threaded rod.
\TaskSpecField{Task instruction.} \textit{``Pick up and hold the threaded rod steady with the right arm, then unscrew the hex nut counterclockwise with the left arm until it comes off.''}
\TaskSpecField{Complete success.} The nut completely separates from the threaded rod, and both components remain supported. Partial loosening is not complete success. Continued holding of the separated parts is sufficient; placing them on the table is not required.
\TaskSpecField{Contact requirements.} \looseness=-1 Regrasping is required when the remaining thread travel exceeds the turning range of one grasp, and the turning gripper must restore effective contact after each repositioning while the right arm keeps the body stable. Coordinating these grasp changes with torque transmission motivates the L3 assignment.
\par\medskip

\Needspace{5\baselineskip}

\paragraph{L3-03: Stacked-Book Pushing \& Rotation.}
\label{app:task-L3-03}
\TaskSpecField{Initial conditions.} Two books are stacked on the table inside a black rectangular start region. A second black rectangle defines the target region. Both grippers start empty.
\TaskSpecField{Goal and constraints.} Both grippers push the stack from the start region into the target region, rotating it one full in-plane turn ($360^{\circ}$) during the transfer. The books must remain stacked and supported by the table throughout; grasping or lifting is prohibited. No centering or prescribed final orientation is required.
\TaskSpecField{Task instruction.} \textit{``Use both grippers to push the two stacked books from one black rectangle fully into the other, rotating the stack one full turn along the way. Keep the books stacked on the table without grasping or lifting them.''}
\TaskSpecField{Complete success.} Both books are wholly within the target rectangle, the full turn was completed in transit, and the stacking, tabletop-support, and no-grasp/no-lift constraints were maintained throughout.
\TaskSpecField{Contact requirements.} The full-turn transfer requires coordinating changes in pushing contact with force direction and turning torque while preserving the stack and tabletop support. These coupled requirements, rather than the use of two arms alone, motivate the L3 assignment.
\par\medskip

\Needspace{5\baselineskip}

\paragraph{L3-04: Block Sweeping.}
\label{app:task-L3-04}
\TaskSpecField{Initial conditions.} In the reference camera view, a fixed dustpan sits to the left with its thin entrance lip on the table, its opening toward the lower right, and its handle toward the upper left. A brush leans against a green box to the right, with its handle toward the upper right and its bristles toward the lower left. Three blue blocks lie between the brush and dustpan in a triangular arrangement: two nearer the arms and one nearer the camera. Both grippers start empty. Section~\ref{sec:initial-condition-sensitivity} analyzes a separate two-block configuration.
\TaskSpecField{Goal and constraints.} The left arm autonomously picks up the brush and sweeps all three blocks into the fixed dustpan. The right arm observes. The dustpan must remain fixed, and the grippers must not manipulate the blocks directly. Setting down the brush is not required.
\TaskSpecField{Task instruction.} \textit{``Pick up the brush next to the green box with the left arm and use it to sweep all three blue blocks into the fixed dustpan. The right arm observes. All three blocks must be inside the pan and supported by it. Keep the dustpan fixed and do not manipulate blocks directly with the grippers. Setting down the brush is not required.''}
\TaskSpecField{Complete success.} All three blocks are inside the dustpan and supported by it, with the dustpan remaining fixed and no direct gripper manipulation of the blocks. The final brush position does not matter.
\TaskSpecField{Contact requirements.} Collecting the three spatially separated blocks requires successive changes in brush--block contact coordinated with force direction and sweeping action, which motivates the L3 assignment. The brush must direct each block toward the dustpan while adjusting to the changing collection configuration. Collection is scored by the physical outcome, without prescribing a particular sequence of strokes.
\par\medskip

\endgroup

%% file: sec/B_real_world_setup.tex
\FloatBarrier
\Needspace{8\baselineskip}
\section{Sim--Real Implementation Details}
\label{app:implementation-details}
\subsection{Real-World Configuration}
\label{app:real-world-setup}
\paragraph{Hardware and sensing.}
Each ARX R5 arm has six revolute joints and an R5-series coupled two-finger gripper with a single opening degree of freedom. The wrist cameras (Orbbec Gemini 305) acquire $848 \times 480$ images at 15\,fps, and the external camera (Orbbec Gemini Pro) acquires $640 \times 480$ images at 30\,fps. Although these cameras support depth sensing, the model receives only their RGB images; depth images are neither provided nor available through tool calls. Robot states are read from the arms' CAN feedback.

\paragraph{Image observations.}
The camera interface returns JPEG snapshots at the above resolutions without resizing or cropping. The model requests observations through tools rather than receiving every camera frame. Frame identifiers and host receipt times are used to check freshness; the three cameras are not hardware synchronized. The stated frame rates describe camera acquisition, not model observation or control rates.

\paragraph{Control and protection.}
Gripper commands specify a normalized opening, with 0 denoting closed and 1 open, which is mapped through the per-arm profiles to SDK position values; these commands do not specify contact force. A control service runs the robot SDK, limits command transmission, and requests protection when heartbeat or feedback from the frontend is lost. When a commanded motion exceeds the arm's kinematic limits, the arm controller triggers a protective stop and halts the arm; the trial then ends and counts as a failure (Section~\ref{sec:trial-protocol}). Command frames, units, and gripper semantics available to the model are described with the basic Skills (Appendix~\ref{app:basic-skills}).

%% file: sec/C_simulation_and_assets.tex
\subsection{Simulation Implementation}
\label{app:simulation-assets}
\begingroup
\setlength{\parskip}{0pt}
\titlespacing*{\paragraph}{0pt}{6pt plus 1pt minus 1pt}{1em}
\paragraph{Simulator and scene setup.}
The simulated environments run directly on Isaac Sim~5.1.0. For each task, a configuration file loads the dual-arm robot, task assets, supports, and the three-camera rig, and each reset initializes object poses, fixation states, joint displacements, and thread engagement (Appendix~\ref{app:task-specifications}; Figure~\ref{fig:simulation-assets}). Camera parameters are estimated from videos of the real scene, and the rendered views approximately preserve the aspect ratios of the corresponding real cameras.

\paragraph{Physics and control settings.}
Table~\ref{tab:sim-settings} lists the simulation settings. Drive gains and rendering settings are shared across tasks, whereas cap and thread contacts and the finite-element jelly use smaller physics steps.
\input{tables/table_s06_alignment_validation}

\paragraph{Contact and material models.}
Rigid objects retain their collision geometry; fixed objects and supports are anchored in the scene, and passive joints represent constrained motions such as drawer sliding. The orange and jelly use GPU finite-element models: the orange applies explicit contact and rest offsets, and the jelly uses the elastic material of its asset. Whiteboard Wiping approximates erasure by accumulating force-weighted coverage over ink cells: contact force is projected onto the board normal, force and motion deadbands are applied, and a cell disappears once its coverage reaches a threshold, so the rendered board shows the remaining ink. This is a simulation proxy rather than a measured material-removal law. Dimensions and masses follow physical measurements (Section~\ref{sec:physical-visual-alignment}); parameters that could not be measured, such as the jelly's elastic coefficients, are estimates.

\paragraph{Outcome measurement.}
The automatic evaluator uses simulator-side quantities that are not part of the model's observations: filtered contact-force matrices for gripper--object and support contacts, soft-object displacement and gripper proximity, and the ink-cell states of Whiteboard Wiping. Proximity alone is not treated as a contact-force measurement.

\par
\endgroup

%% file: tables/table_s06_alignment_validation.tex
\begin{table}[!t]
\centering
\small
\setlength{\tabcolsep}{4pt}
\renewcommand{\arraystretch}{1.12}
\caption{\textbf{Simulation settings used in the evaluation.} Material and wiping values are simulation parameters rather than measured physical properties.}
\label{tab:sim-settings}
\begin{tabular}{@{}>{\raggedright\arraybackslash}p{0.22\linewidth}>{\raggedright\arraybackslash}p{\dimexpr0.78\linewidth-2\tabcolsep\relax}@{}}
\toprule
\textbf{Component} & \textbf{Setting} \\
\midrule
Physics stepping & 240 Hz by default; 1920 Hz for Marker Cap Removal, Bottle Cap Unscrewing, and Nut Unscrewing; 360 Hz for Jelly Handling. \\
\addlinespace
Constraint solver & TGS; 32 articulation position iterations; 4 velocity iterations for L1 tasks and 0 for all other tasks. \\
\addlinespace
Joint drives & Stiffness/damping in Isaac Sim joint-drive units: arm joints 1500/80; finger joints 2000/40. \\
\addlinespace
Gripper opening & Range 0--70 mm; each finger joint moves by half of the commanded opening. \\
\addlinespace
Camera images & Rendered at $1280\times954$ pixels (external) and $1280\times720$ pixels (each wrist); downscaled to a 768-pixel long edge before being sent to the model. \\
\addlinespace
Orange contact margins & Contact offset 3 mm; rest offset 1.5 mm. \\
\addlinespace
Jelly material & Young's modulus $E=3$ kPa; Poisson ratio $\nu=0.45$; density $\rho=1030$ kg/m$^3$; dynamic friction coefficient $\mu_d=0.6$. \\
\addlinespace
Wiping accumulation & Force weight $1.5/[1+(2\,\mathrm{N}/(F-0.1\,\mathrm{N}))^2]$ for normal force $F>0.1$ N and zero otherwise; motion deadband 0.5 mm; maximum accepted sliding sample 20 mm; coverage threshold 2.5 (dimensionless). \\
\bottomrule
\end{tabular}
\end{table}

%% file: sec/E_evaluation_specifications.tex
\FloatBarrier
\Needspace{8\baselineskip}
\section{Evaluation Protocol Details}
\label{app:evaluation-specifications}

\subsection{Basic Skills Description}
\label{app:basic-skills}
\label{app:experience-records}
Basic Skills provide the interface description and general operating guidance for real-world execution (Section~\ref{sec:observation-action-interface}). They are made available to Codex before each session and remain unchanged across trials. Figure~\ref{fig:basic-skills} summarizes their content in condensed form. The guidance is task-agnostic: it specifies how to read the embodiment and interface, observe and verify physical outcomes, coordinate the two arms, and recover locally, but not how to complete any evaluation task.

\begin{figure}[!p]
\centering
\captionsetup{position=bottom,skip=10pt}
\input{figures/box_c1_basic_skills}
\caption{\textbf{Basic Skills for real-world execution.} The guidance covers embodiment checks, active observation, gripper control, contact verification, dual-arm coordination, recovery, and completion assessment.}
\label{fig:basic-skills}
\end{figure}

\subsection{Trial Execution and Model Calls}
\label{app:trial-statistics}
\noindent\textbf{Initial states and resets.}
Simulated initial states are defined by saved scenes and layout configurations. Some layouts are generated from random seeds, checked for geometry and reachability, and then saved; others apply predefined pose transformations or use separate scene files. Each episode loads its configuration, resets and settles the scene, and then obtains the first observation; run records preserve the loaded configuration. In the real world, an operator manually restores object placement, supports or fixtures, and mechanical engagement to the initial conditions in Appendix~\ref{app:task-specifications}, using a fixed set of object instances with repeated copies where needed. Real and simulated initial states thus follow the same task specifications, but exact object poses are not matched one-to-one. 

\noindent\textbf{Budgets and continuations.}
By default, a simulated episode is limited to 150 model calls and 5 hours of wall-clock time. Real-world episodes have no preset time or model-call limit and end when the model reports completion or judges the task infeasible; the latter counts as failure. Trials also terminate on an operator stop or takeover, or a protective stop (Section~\ref{sec:trial-protocol}). Continuations resume unfinished episodes with their interaction history and remain within the same trial; simulation budgets are not replenished. Independently reset executions are new trials. An operator may stop the robot after confirming autonomous completion; completion that requires operator assistance does not count as autonomous success.

\noindent\textbf{Model context in simulation.}
Each simulated decision is a separate Codex CLI call with tool use and Skill discovery disabled. The fixed model instructions contain the interface description, general operating guidance, arm roles, and a brief initial-state note. Each call carries the full text history of the episode, the current images, and images from the two most recent decisions; history is cleared between episodes.

\noindent\textbf{Model calls.}
A model call is one inference request actually issued to the model; tool invocations, robot actions, controller updates, and calls made during Skills preparation are not counted, and calls in continuation segments belong to the original episode. Table~\ref{tab:supp-model-calls} reports mean model calls per task for reference only: the environments differ in execution granularity (one structured action per simulated call versus several tool calls per real-world action), termination limits, and counting methods, so the counts are not directly comparable.

\begin{table}[!htbp]
\centering\small
\caption{\textbf{Mean model calls per task, reported for reference.} Successful and unsuccessful trials are included. Values are rounded to the nearest integer; level and overall rows average unrounded task means. Counts are not directly comparable between environments: each simulated call returns one structured action executed by the planner, whereas a real-world action takes several tool calls, since the model requests fresh observations and moves the arms in bounded segments to refine their poses.}
\label{tab:supp-model-calls}
\setlength{\tabcolsep}{3pt}
\begin{tabularx}{\linewidth}{@{}l>{\raggedright\arraybackslash}X>{\centering\arraybackslash}p{0.16\linewidth}>{\centering\arraybackslash}p{0.16\linewidth}@{}}
\toprule
\textbf{Task ID} & \textbf{Task} & \textbf{Simulation} & \textbf{Real world} \\
\midrule
\multicolumn{2}{@{}l}{\textbf{L1 mean}} & 46 & \AstraCallCell{Main-L1}{real-main} \\
L1-01 & \AstraResultCell{L1-01}{name} & 18 & \AstraCallCell{L1-01}{real-main} \\
L1-02 & \AstraResultCell{L1-02}{name} & 74 & \AstraCallCell{L1-02}{real-main} \\
L1-03 & \AstraResultCell{L1-03}{name} & 84 & \AstraCallCell{L1-03}{real-main} \\
L1-04 & \AstraResultCell{L1-04}{name} & 35 & \AstraCallCell{L1-04}{real-main} \\
L1-05 & \AstraResultCell{L1-05}{name} & 16 & \AstraCallCell{L1-05}{real-main} \\
\midrule
\multicolumn{2}{@{}l}{\textbf{L2 mean}} & 93 & \AstraCallCell{Main-L2}{real-main} \\
L2-01 & \AstraResultCell{L2-01}{name} & 102 & \AstraCallCell{L2-01}{real-main} \\
L2-02 & \AstraResultCell{L2-02}{name} & 86 & \AstraCallCell{L2-02}{real-main} \\
L2-03 & \AstraResultCell{L2-03}{name} & 80 & \AstraCallCell{L2-03}{real-main} \\
L2-04 & \AstraResultCell{L2-04}{name} & 108 & \AstraCallCell{L2-04}{real-main} \\
L2-05 & \AstraResultCell{L2-05}{name} & 92 & \AstraCallCell{L2-05}{real-main} \\
L2-06 & \AstraResultCell{L2-06}{name} & 72 & \AstraCallCell{L2-06}{real-main} \\
L2-07 & \AstraResultCell{L2-07}{name} & 114 & \AstraCallCell{L2-07}{real-main} \\
\midrule
\multicolumn{2}{@{}l}{\textbf{L3 mean}} & 113 & \AstraCallCell{Main-L3}{real-main} \\
L3-01 & \AstraResultCell{L3-01}{name} & 94 & \AstraCallCell{L3-01}{real-main} \\
L3-02 & \AstraResultCell{L3-02}{name} & 80 & \AstraCallCell{L3-02}{real-main} \\
L3-03 & \AstraResultCell{L3-03}{name} & 136 & \AstraCallCell{L3-03}{real-main} \\
L3-04 & \AstraResultCell{L3-04}{name} & 140 & \AstraCallCell{L3-04}{real-main} \\
\midrule
\multicolumn{2}{@{}l}{\textbf{Overall mean}} & 83 & \AstraCallCell{Main-All}{real-main} \\
\bottomrule
\end{tabularx}
\end{table}

\enlargethispage{\baselineskip}
\subsection{Bottleneck Criteria and Outcome Qualifications}
\label{app:failure-annotation}
Table~\ref{tab:supp-protocol-failures} lists the observable evidence, drawn from execution videos, action records, and robot feedback, used to assign the interaction bottlenecks in Section~\ref{sec:failure-analysis}. Assignment is qualitative: a trial may show several bottlenecks, and budget exhaustion or protective stops are termination conditions, not bottlenecks.

\paragraph{Reported outcome qualifications.}\phantomsection\label{app:result-qualifications}
\looseness=-1 Both environments apply the success criteria in Appendix~\ref{app:task-specifications}. Simulation checks them automatically with fixed, executable conditions (Appendix~\ref{app:simulation-assets}), whereas real-world outcomes are judged visually by a human observer and therefore involve subjective judgment. This mainly affects Charger Alignment (Section~\ref{sec:sim-real-consistency}), where trials were accepted and manually terminated once alignment appeared adequate, and Whiteboard Wiping, where trials were accepted once the writing appeared erased overall even if faint marks remained. Real-world outcomes of these tasks may thus vary slightly with the observer's judgment.

\noindent\begin{minipage}{\linewidth}
\begingroup
\centering\small
\captionof{table}{\textbf{Observable evidence for the four interaction bottlenecks.}}
\label{tab:supp-protocol-failures}
\begin{tabularx}{\linewidth}{@{}>{\raggedright\arraybackslash}p{0.29\linewidth}X@{}}
\toprule
\textbf{Bottleneck} & \textbf{Observable evidence} \\
\midrule
Spatial reasoning and execution precision & Object localization is achieved, but relative orientation, placement, alignment, or motion planning does not realize the spatial relationship required for the task. \\
Contact configuration selection and establishment & The chosen contact location or orientation does not provide an accessible, stable, and effective interaction under the relevant geometry and material constraints. \\
Force regulation & Action records and physical responses indicate difficulty controlling the task-required effects of contact forces or torques on motion or deformation, beyond ordinary support, transport, and geometric placement. A missed grasp or final-state instability alone is insufficient. \\
Coordination within and across interactions & Contact configuration and force regulation fail jointly within an interaction, or successive actions do not adapt to changing physical states and accumulate into task completion. \\
\bottomrule
\end{tabularx}
\par
\endgroup
\normalsize

\par
\end{minipage}

%% file: figures/box_c1_basic_skills.tex
\begingroup
\definecolor{skilltitle}{HTML}{2B394B}
\definecolor{skillbody}{HTML}{F5F7FA}
\definecolor{skillborder}{HTML}{BBCBDD}
\begin{tcolorbox}[
  enhanced, sharp corners, boxrule=0.4pt,
  colback=skillbody, colframe=skillborder,
  colbacktitle=skilltitle, coltitle=white,
  title={Basic Skills for Real-World Execution},
  fonttitle=\bfseries\fontsize{10.5}{12.5}\selectfont,
  fontupper=\fontsize{9.5}{11.1}\selectfont,
  left=8pt, right=8pt, top=5pt, bottom=4pt,
  toptitle=4pt, bottomtitle=4pt,
  before skip=0pt, after skip=0pt
]
\setlength{\parindent}{0pt}
\setlength{\parskip}{3pt}
\setlist[itemize]{label=\textendash,leftmargin=10pt,labelsep=3pt,
  topsep=1pt,itemsep=0.5pt,parsep=0pt,partopsep=0pt}

\textbf{Task, interface, and embodiment}
\begin{itemize}
  \item Identify the objects, goal, arm roles, permitted contacts, prohibited actions, and completion criteria \mbox{from the task instruction}.
  \item Read enable and hold/motion states, faults, feedback freshness, target and measured joint values, tracking error, and gripper opening. Confirm control ownership and live images.
  \item Verify arm identities, command frames and units, tool length, gripper geometry, and tabletop position. Image left--right does not determine robot-coordinate directions. Use the deployed limits, not \mbox{historical case-specific values}.
\end{itemize}

\textbf{Active visual observation}
\begin{itemize}
  \item Verify the overview and physical left- and right-wrist views from live images. Use the overview for scene and arm clearance, and wrist views for grasping and local contact.
  \item Resolve occlusion with complementary views. If needed, hold the working arm and move the empty observation arm along a checked path, accounting for its links, gripper, and cables.
  \item Inspect fresh images before selecting the next action. If observations become stale, refresh them and reassess the action. Do not infer physical insertion or contact from image overlap alone.
\end{itemize}

\textbf{Gripper control and grasp verification}
\begin{itemize}
  \item Treat gripper opening as a position target, not a force command. Align height and depth before gradual closure; further closure cannot compensate for fingers that do not surround the object.
  \item Verify grasping from the object's response to subsequent permitted motion. A closing command, completed arm motion, or reached joint target alone does not establish a stable grasp.
\end{itemize}

\textbf{Segmented motion and contact verification}
\begin{itemize}
  \item Raise and reorient in clear space, then approach in bounded segments. Reduce increments near the tabletop, objects, or the other gripper, while preserving the deployed motion limits.
  \item Check feedback before each segment; monitor faults, tracking error, and feedback freshness during execution. Afterwards, hold and inspect the views before continuing.
  \item Compare the object, its support, and the surrounding scene before and after motion. Check whether the intended displacement or contact occurred and whether support, alignment, and stability were maintained.
\end{itemize}

\textbf{Dual-arm coordination}
\begin{itemize}
  \item Plan from a consistent current state and check clearance throughout the motion, including links, tools, and cables. Single-arm reachability does not establish collision-free dual-arm motion.
  \item Concurrent requests do not imply synchronized hardware execution.
\end{itemize}

\textbf{Stopping and local recovery}
\begin{itemize}
  \item Stop advancing if motion is blocked, objects slip or tilt, arms approach each other, or contact becomes unclear. Use the verified stop/hold mechanism for abnormal conditions.
  \item Reassess depth, contact location, object support, and viewpoint before retrying. Regrasp or retreat only when support and the path are clear; do not repeatedly close or lift at the same failed grasp.
  \item Resume only with new evidence supporting the correction. Replan from the current state rather than replaying a trajectory or blindly returning home. Preserve holding when disabling could cause an arm to sag.
\end{itemize}

\textbf{Completion verification}
\begin{itemize}
  \item Verify the final goal, required constraints, and object stability. Inspect actual boundaries and support from complementary views; an upper-surface projection alone may not establish the object's footprint.
  \item Report completion as unconfirmed when evidence is insufficient. State whether the grippers have withdrawn and whether either arm remains enabled or has an active trajectory.
\end{itemize}

\begin{tcolorbox}[
  enhanced, sharp corners, boxrule=0pt,
  colback=skilltitle, coltext=white,
  left=6pt, right=6pt, top=5pt, bottom=5pt,
  before skip=4pt, after skip=0pt,
  fontupper=\fontsize{9.5}{11.1}\selectfont
]
\textbf{Execution report}\par
Report the outcome, supporting observations, corrections, remaining uncertainty, and final arm/hold state. Distinguish measured feedback from visual judgments.
\end{tcolorbox}
\end{tcolorbox}
\endgroup